\documentclass{article}
\usepackage{ijcai26}

\usepackage{times}
\usepackage{newtxtext}
\usepackage{soul}
\usepackage{url}
\usepackage[hidelinks]{hyperref}
\usepackage[utf8]{inputenc}
\usepackage[small]{caption}
\usepackage{graphicx}
\usepackage{amsmath}
\usepackage{amsthm}
\usepackage{xspace}
\usepackage{booktabs}
\usepackage[switch]{lineno}
\usepackage{amsfonts}  

\usepackage{latexsym}
\usepackage{graphicx}
\usepackage{subcaption} 
\usepackage{color}
\usepackage{paralist}
\usepackage{comment}
\usepackage{todonotes}
\usepackage{placeins}
\usepackage{float}
\usepackage{placeins}

\usepackage[ruled,vlined]{algorithm2e}
\SetKwInput{Input}{Input}
\SetKwInput{Output}{Output}
\SetAlgoCaptionSeparator{.}

\newtheorem{theorem}{Theorem}
\newtheorem{lemma}[theorem]{Lemma}

\newtheorem{assumption}[theorem]{Assumption} 
\newtheorem{remark}[theorem]{Remark}

\begingroup
\catcode`\~=11
\gdef\urltilde{\lower 0.6ex\hbox{~}}
\endgroup

\newcommand{\A}{\mathcal{A}}

\newcommand{\M}{\mathcal{M}}
\newcommand{\N}{\mathcal{N}}

\newcommand{\ra}{\rightarrow}

\newcommand{\tup}[1]{(#1)}            

\newcommand{\Nat}{{\rm I\kern-.23em N}}

\newcommand{\LTL}{{\sc ltl}\xspace}

\newcommand{\LTLf}{{\sc ltl}$_f$\xspace}

\newcommand{\LDLf}{{\sc ldl}$_f$\xspace}

\newcommand{\argmax}{\mathop{\mathrm{argmax}}}

\newcommand{\rmax}{\textsc{R-Max}\xspace}
\newcommand{\qlearn}{\textsc{Q-learning}\xspace}
\newcommand{\qrmax}{\textsc{QR-Max}\xspace}
\newcommand{\rmaxrm}{\textsc{R-MaxRM}\xspace}
\newcommand{\qrmaxrm}{\textsc{QR-MaxRM}\xspace}

\newcommand{\bqrmax}{\textsc{Bucket-QR-Max}\xspace}

\usepackage{pifont}
\newcommand{\xmark}{\ding{55}}  

\title{Model-Based Reinforcement Learning in Discrete-Action\\ Non-Markovian Reward Decision Processes}

\author{
Alessandro Trapasso$^{1, 2}$
\and
Luca Iocchi$^2$\and
Fabio Patrizi$^2$\\
\affiliations
$^1$Fondazione Bruno Kessler, Trento, Italy\\
$^2$Sapienza University of Rome, Rome, Italy\\
\emails
atrapasso@fbk.eu,
\{trapasso, iocchi, patrizi\}@diag.uniroma1.it
}

\begin{document}

\maketitle

\begin{abstract}
Many practical decision-making problems involve tasks whose success depends on the entire system history, rather than on achieving a state with desired properties. Markovian Reinforcement Learning (RL) can be inadequate for such tasks, while modeling them as non-Markovian reward decision processes (NMRDPs) enables agents to handle temporal dependencies.
Existing approaches offer limited formal guarantees on both (near-)optimality and sample efficiency. We address both issues with \qrmax, a novel model-based algorithm for NMRDPs with discrete actions that factorizes Markovian transition learning from non-Markovian reward handling via reward machines. To our knowledge, this is the first model-based RL algorithm for discrete-action NMRDPs that leverages this factorization to obtain PAC convergence to $\varepsilon$-optimal policies with polynomial sample complexity.
We then extend \qrmax\ to continuous state spaces with \bqrmax, a SimHash-based discretizer that preserves the same factorized structure and achieves fast and stable learning without manual gridding or function approximation. We experimentally compare our method with state-of-the-art model-based RL approaches on environments of increasing complexity, showing substantially improved sample efficiency and greater robustness in finding optimal policies.
\end{abstract}


\section{Introduction}

Reinforcement Learning (RL) has achieved striking success in domains that range from game playing to robotics, thanks to its ability to shape behaviour from sparse feedback.  
Designing such feedback, however, is difficult when the objective spans multiple temporally ordered sub‑goals: specifying such tasks with a Markovian reward often yields brittle or misleading signals.  Non‑Markovian Reward Decision Processes (NMRDPs) address this difficulty by allowing the reward to depend on the environment history; compact logical or automata formalisms, including Reward Machines and \LTLf/\LDLf specifications, allow for encoding these rewards succinctly~\cite{BacchusBG96,TGSPK06,BDGP2018,pmlr-v80-icarte18a}.  Any NMRDP can be reduced to an ordinary Markov Decision Process (MDP) whose states are pairs $(s,q)$ of environment state and automaton state~\cite{BDGP2018}.  Model‑free agents operate in $S\times Q$ without storing a transition model \cite{de2020restraining,pmlr-v80-icarte18a}, while model‑based algorithms such as \rmax\ \cite{Brafman03} can in principle learn a PAC‑optimal policy on the same space \cite{GaonB20}.  
Despite this theoretical effectiveness, treating $(s,q)$ as an opaque state discards the fact that the transition kernel is still Markovian and that only the reward component is history‑dependent, leading to redundant exploration.

We show that preserving this structural distinction unlocks substantial gains.  By factorizing the dynamics into the Markovian environment transition $P(s'|s,a)$ and the deterministic automaton update $q'=\eta(q,s')$, we devise  \qrmax, a model‑based algorithm that reuses each learned environment transition for every automaton state and retains separate optimistic bonuses.  We show that \textsc{QR-Max} is PAC--MDP: with high probability, all but a polynomial number of interaction steps are \(\varepsilon\)-optimal. Its sample bound has the factorized unknown-component term
\(U=|S||A|m_E+|S||Q|m_Q\), where \(m_E,m_Q\) are the environment and automaton knownness thresholds, instead of the product-space count
\(|S||Q||A|m\) incurred by applying \rmax\ directly on \(S\times Q\). Thus factorization avoids relearning the environment transition model separately for every automaton state.
To the best of our knowledge, this is the first model-based RL algorithm for discrete-action NMRDPs that explicitly leverages the decoupling between environment and automaton dynamics to obtain PAC guarantees with improved sample-complexity bounds.
Building on the same idea, we also introduce \textsc{Bucket-QR-Max}, a variant of \qrmax combining SimHash discretisation \cite{Charikar2002} with the factorised counts, which extends the approach to continuous state spaces without manual gridding and, under a standard bounded-discretisation assumption, admits a conditional PAC--MDP bound.

We evaluate \qrmax\ on five maps, each combined with seven distinct Reward Machines of increasing spatial and temporal complexity, and we also include a continuous‑state variant. 
On the benchmarks we consider, our approach is one to two orders of magnitude more sample-efficient than standard \rmax, the reward-machine-aware QRM, and optimistic explorers such as UCBVI, PSRL and OPSRL, while retaining formal PAC guarantees.
These results suggest that \qrmax and its continuous-state extension are practical model-based algorithms with formal sample-complexity bounds for discrete-action NMRDPs. Moreover, optimal discrete-action solutions can drive effective and efficient training in continuous-actions MDPs using hierarchical RL approaches (e.g., \cite{cipollone_ras_2025}).

\section{Related Work}

RL with non-Markovian rewards has been studied via NMRDPs, Reward Machines (RMs), and temporal-logic specifications. Two strands are most relevant here: model-based RL for NMRDPs and structure-aware methods for non-Markovian tasks.
Prior work applies \rmax\ to NMRDPs by incrementally extending an MDP with reward-automaton states, i.e., learning in a representation akin to the \rmax\ baseline we use~\cite{GaonB20}. In contrast, our approach improves sample efficiency by factorizing environment dynamics and reusing learned environment transitions across automaton states.

We also benchmark against state-of-the-art model-based exploration methods for finite MDPs. \textsc{UCBVI}~\cite{azar17a} matches minimax-optimal regret using empirical counts and confidence bonuses, with memory $\mathcal O(S^{2}A)$. \textsc{PSRL}~\cite{osband2013} and its optimistic variant \textsc{OPSRL}~\cite{tiapkin2022optimistic} maintain Dirichlet posteriors for transitions, yielding memory $\mathcal O(H,S^{2}A)$ (see Strens~\shortcite{Strens2000}, §5.1); this quadratic dependence on $S$ makes them impractical on our largest grids, hence we omit \textsc{PSRL}/\textsc{OPSRL} when the state space is very large.
On the 15$\times$15 grid with $H{=}250$, \textsc{PSRL/OPSRL} need $\approx50.6$M entries vs.\ $0.204$M for \qrmax, exceeding our budget (see Appendix).

A recent line of work develops structure-aware algorithms for non-Markovian tasks.
Exploration in \emph{deterministic} RMs has been studied with regret bounds exploiting automaton structure~\cite{Bourel2023}, and extensions to Probabilistic RMs (PRMs) have also been proposed~\cite{DBLP:journals/corr/abs-2412-19194,Lin_Zhang_2025}.
These methods typically optimize regret (often average-reward or finite-horizon) under a known RM/PRM, and are therefore not directly comparable to our discounted PAC-style evaluation; in our setting we provide a tabular model-based baseline and improve over \rmax\ on $S\times Q$ by reusing environment transitions across automaton states. Our factorization is orthogonal to factored MDPs, where the state is represented by variables and the transition model factorises as a dynamic Bayesian network~\cite{10.5555/1624312.1624325,Guestrin_2003,Strehl2007FactoredRMax,osband2014nearoptimalreinforcementlearningfactored}. Here, the underlying environment need not be factored: the gain comes from separating Markovian environment dynamics from the non-Markovian reward automaton. Other NMRDP and temporal-logic approaches either estimate alternative dynamics or focus mostly on model-free learning~\cite{Gupta2021NonMarkovianRL,pmlr-v80-icarte18a,de2020restraining}, whereas we focus on model-based PAC learning. Among model-free methods, one approach maximizes the probability of satisfying an \LTL formula~\cite{ShaoK23}; it differs from our setting in focusing on infinite traces, while we use finite-trace automata (regular languages). Nonetheless, on finite traces such approaches reduce to Q-learning and QRM, which we compare against, so our experiments indirectly account for~\cite{ShaoK23}.

Finally, a substantial body of work aims to learn the NMR model itself, e.g.,~\cite{GaonB20}; this is outside our scope since we assume a given NMR specification. To the best of our knowledge, \qrmax\ is the first tabular model-based RL algorithm for discounted NMRDPs that factorizes environment and reward dynamics to obtain PAC-MDP guarantees with improved sample-complexity bounds over \rmax\ on $S\times Q$.

\section{Background}
A \emph{Markov Decision Process} (MDP) is a tuple $\M=\tup{S,A,P,R_E}$ with finite states $S$, actions $A$, transition kernel $P(s'|s,a)$, and reward function $R_E:S\times A\times S\ra \mathbb{R}$.
A \emph{policy} $\rho:S\ra A$ induces the \emph{return}
$v^\rho(s)=E_\rho[r_0+\gamma r_1+\gamma^2 r_2+\cdots]$, where
$r_i=R_E(s_i,\rho(s_i),s_{i+1})$, $s_0=s$, and $\gamma\in[0,1]$ is the \emph{discount factor}.
The goal is to find an \emph{optimal policy} $\rho^*$ maximizing $v^*(s)$ for all $s\in S$.
In RL, the functions $P$ and $R_E$ are unknown and must be learned through interaction.

RL methods are typically \emph{model-free} or \emph{model-based}:
the former seek $\rho^*$ without explicitly estimating $P$ and $R_E$, while the latter learn both.
Two representative algorithms are \qlearn~\cite{watkins1992} and \rmax~\cite{Brafman03}.
A central notion is the optimal state-action value function
\(Q^\star:S\times A\to\mathbb{R}\), where \(Q^\star(s,a)\)
is the expected discounted return obtained by executing \(a\) in
\(s\) and then acting optimally; an optimal policy is
\(\rho^\star(s)\in\arg\max_a Q^\star(s,a)\). While \(Q^\star\)
can be computed from \(P\) and \(R_E\) (e.g., via \emph{Value Iteration}~\cite{sutton}), this is not directly possible in RL since $P$ and $R_E$ are unknown.

{\qlearn} maintains an estimate \(\widehat Q\), initialized arbitrarily and updated from experience using an update rule based on \emph{Bellman's optimality condition}; with sufficient exploration, \(\widehat Q\) approximates \(Q^\star\) and induces a near-optimal policy.
In contrast, {\rmax} estimates $P$ and $R_E$ by collecting samples for state-action pairs $(s,a)$ until they are deemed \emph{known}; it then plans in the learned model (typically via value iteration), using optimism to drive exploration.
In standard RL, $R_E$ is \emph{Markovian}, depending only on the current (and possibly next) state; here we consider \emph{non-Markovian} rewards.

A \emph{Non-Markovian Reward Decision Process} (NMRDP) is a tuple $\N=\tup{\M,R_{\A}}$, where $R_{\A}:S^+\ra \mathbb R$ is a \emph{Non-Markovian Reward Function} (NMR) depending on state histories.
NMRs can be specified compactly, e.g., by Reward Machines~\cite{pmlr-v80-icarte18a} or Restraining Bolts~\cite{de2020restraining}.
We unify them using a \textit{Deterministic Finite-state Automaton} (DFA)
$\A=\tup{S,Q,q_0,\eta,F}$, where $S$ is the input alphabet, $Q$ the automaton states, $q_0$ the initial state, $\eta:Q\times S\ra Q$ the transition function, and $F$ the final states.
We assume full observability of both the environment state and the automaton state: after executing an action, the agent observes $\tup{s', q'}$ (equivalently, it can update $q'$ deterministically when $\A$ is known).
The reward is given by $R_{\A}(q,s',q')$ for the last automaton transition $\tup{q,s',q'}$.
In this notation, a reward machine is a deterministic automaton endowed with transition rewards, while a restraining bolt is the DFA induced by the temporal specification; acceptance can be encoded through $R_{\A}$, e.g., by assigning positive reward to transitions entering $F$. Thus $F$ has no separate algorithmic role unless acceptance itself is part of the reward specification.

$\A$ compactly captures temporal properties of the history; in practice $|Q|\ll|S|$.
In an NMRDP $\N=\tup{\M,R_{\A}}$ with $\M=\tup{S,A,P,R_E}$, after a trajectory $s_0a_0\cdots s a s'$, the reward is $R_E(s,a,s')+R_{\A}(q,s',q')$, where $q$ (resp.\ $q'$) is the DFA state after consuming $s_0\cdots s$ (resp.\ $s_0\cdots s s'$).
Although this makes the reward \emph{stateful}, NMRDPs can be reduced to Markovian RL by augmenting the state with the automaton component.
As shown in~\cite{BDGP2018}, $\N$ is equivalent to an MDP $\M'=\tup{S',A,P',R'}$ over $S'=S\times Q$, with $P'$ capturing synchronous transitions and $R'$ Markovian.
Standard RL methods can then be applied (see, e.g.,~\cite{de2020restraining,pmlr-v80-icarte18a}).

However, this reduction treats $(s,q)$ as an atomic state, losing the distinction between the MDP dynamics ($\M$) and the DFA dynamics ($\A$).
For instance, learning transitions from $(s,q)$ on action $a$ does not help with $(s,q')$, even though the environment dynamics from $s$ are identical.
Since $\A$ is deterministic (and typically easier to learn), this coupling can yield redundant exploration.
We investigate how to retain this opportunity by factorizing $\N$'s dynamics into $\M$'s and $\A$'s.


\section{Decoupling Environment and Reward Dynamics}
\label{sec:factorization}

We model an NMRDP $\N=\tup{\M,R_{\A}}$, by combining the dynamics of the environment $\M$ with that of the NMR $R_{\A}$ and the DFA $\A$ it is based on, as follows: 
\begin{inparaenum}
\item[(1)] when the system is reset, the environment is set to an initial state $s_0$, $\A$ is set to its initial state $q_0$, and a dummy DFA transition $q_0 = \eta(q_0, s_0)$ is performed, s.t.~$\A$ consumes $s_0$;
\item[(2)] at each step, given the current state $(s,q)$ and the chosen action $a$, the environment progresses to a state $s'$ (with probability $P(s'|s,a)$), $\A$ progresses to $q' = \eta(q,s')$ (with probability 1), 
and the returned total reward is 
$R_E(s,a,s')+R_{\A}(q,s',q')$.
\end{inparaenum}

The resulting transition function is captured as follows:
\begin{eqnarray*} 
P(s', q' | s, q, a) =  P(q' | s',q,s,a) \, P(s' | s,q,a), 
\end{eqnarray*}
where $P(s', q' | s, q, a)$, $P(q' | s',q,s,a)$, 
$P(s' | s,q,a)$ are the transition probability distributions of $\N$, $R_{\A}$'s automaton $\A$, and the environment $\M$, respectively.
However, since $\M$ is Markovian:
$P(s' | s,q,a) = P(s' | s,a)$, i.e., given $s$ and $a$, $s'$ is independent of the history leading $\A$ to $q$.
Also, since $q'$ depends only on $q$ and $s'$:
$P(q' | s',q,s,a) = P(q' | s',q)$.
Thus, $\N$'s transition model is:
\begin{equation}\label{eq:P}
P(s', q' | s, q, a)  =   P(s' | s,a) P(q' | s',q).
\end{equation}

As for the reward function, we simply have:
\begin{equation} 
R(s,q,a,s',q') = R_E(s,a,s') + R_{\A}(q,s',q').
\label{eq:R}
\end{equation}
Note that $R_{\A}(q,s',q')$ does not depend on $s$ or $a$, i.e., on how state $(s',q')$ has been reached from any state $(s,q)$.

Equations \ref{eq:P} and \ref{eq:R} describe a factorization of the system dynamics that decouples the Markovian environment dynamics ($P(s' | s,a)$, $R_E(s,a,s')$) from the non-Markovian reward expressed by the DFA ($ P(q' | s',q)$, $ R_{\A}(q,s',q')$).

\paragraph{Observability and RM Knowledge.}
We assume the agent observes the current automaton state $q$ (as standard in RM settings via a monitor over observable propositions)
and receives the RM reward component $r_{\A}$ from the RM monitor/wrapper. We consider two settings: \qrmax\ does \emph{not} assume knowledge of the automaton transition function $\eta$ (it only receives the $q\!\to\!q'$ signal), while \qrmaxrm\ assumes the RM is given (including $\eta$), enabling counterfactual updates across automaton states. If $q$ is not observable, the problem becomes partially observable and is outside our scope.



\paragraph{Learning Algorithms.}
Discrete model-based RL can be applied in the joint state space $S \times Q$ to estimate the transition and reward functions
$P(s', q' | s, q, a)$ and $R(s,q,a,s',q')$. 
In this paper, we compare our method based on the above factorization with different model-based RL approaches that use the joint space $S \times Q$, such as that in~\cite{BDGP2018,GaonB20}.
%
%
As shown in the experiments, the use of the joint space does not scale well with the task's complexity, since it does not exploit the Markovian nature of the environment transitions.

\subsection{\qrmax~Algorithm}

In this section, we describe \qrmax, a novel algorithm 
extending \rmax~to non-Markovian rewards and Markovian transitions, by exploiting the factorization described above.
As shown later on, {\qrmax} provides the same guarantees for near-optimal solutions as {\rmax}, but exhibits a significantly higher 
sample efficiency.
The main advantage of {\qrmax} is that the terms of 
Equations \ref{eq:P} and \ref{eq:R},
i.e., $P(s' | s,a)$, $P(q' | s',q)$, $R_E(s,a,s')$, and $R_{\A}(q,s',q')$, can be estimated in parallel by setting different thresholds,
$t_E$ and $t_Q$, to determine when they are \emph{known}.
For $n_E(s,a)$ and $n_Q(q,s')$ the number of visits to the environment's and  DFA's states, respectively, 
we use the following conditions to label states as known: $n_E(s,a) \geq t_E \implies \textit{Known}(s,a) $
and $n_Q(q,s') \geq t_Q \implies \textit{Known}(q,s')$. 

The experience for $n_E(s,a)$ is collected independently of the state of $\A$ and once $\textit{Known}(s,a)$ holds and the environment 
transition and reward functions are estimated, respectively as $\hat{P}(s' | s,a)$ and $\hat{R_E}(s,a,s')$, these values are 
used also in other unseen states of $\A$.
This decoupling yields a significant improvement in sample efficiency with a standard, monolithic formulation of the joint space.
Observe, in particular, that for deterministic $\A$ (as is the case here), $t_Q=1$ is sufficient to know all $\A$'s states, thus yielding a significant saving in time for learning $\A$. We keep the threshold $t_Q$ for modularity and generality, to make the algorithm applicable also to more general machines, such as \emph{stochastic} RMs~\cite{Corazza_Gavran_Neider_2022}.

\begin{algorithm}[t]
\caption{\qrmax}\label{alg:qrmax}
\begin{scriptsize}
\Input{$S, Q, A$ states and actions, $E$ environment, \\
 $\gamma$ discount factor, 
 $R_{\max}$ max expected reward, \\
 $t_E, t_Q$ thresholds, 
 $T$ value iteration horizon\\ 
 }
\Output{$QT$ optimal Q-table\\
~\\}
$QT(s,q,a) \gets R_{\max} / (1-\gamma)$, $\forall s \in S, q \in Q, a \in A$ \\
$\tau_E(s,a,s') \gets 0 $, $\rho_E(s,a,s') \gets 0 $,  $\forall s, s' \in S, a \in A$ \\
$n_E(s,a) \gets 0 $, $\forall s \in S, a \in A$ \\
$\tau_Q(q,s',q') \gets 0 $, $\rho_Q(q,s',q') \gets 0 $, $\forall s' \in S, q, q' \in Q$ \\
$n_Q(q,s') \gets 0 $, $\forall s' \in S, q \in Q$ \\
$s,q \gets E.reset()$ \\
\For {$t = 1, \ldots , N_{steps}$} {
    $\bar{a} \gets \mathrm{argmax}_a \, {QT(s,q,a)} $ \\
    $s', q', r_E, r_{\A}, done = E.step(\bar{a})$ \\
    doVI = False \\
    \If{$n_E(s,\bar{a})<t_E$} {
      $\tau_E(s,\bar{a},s') \; += 1$ \\
      $\rho_E(s,\bar{a},s') \; += r_E$ \\
      $n_E(s,\bar{a}) \; += 1$ \\
      \If{$n_E(s,\bar{a})=t_E$} {
        doVI = True  
      }    
    }
    
    \If{$n_Q(q,s')<t_Q$} {
      $\tau_Q(q,s',q') \; += 1$ \\
      $\rho_Q(q,s',q') \; += r_{\A}$ \\
      $n_Q(q,s') \; += 1$ \\
      \If{$n_Q(q,s')=t_Q$}{
        doVI = True
      }
    }
    \If{done}{
        $QT(s',q',a') \gets 0$, $\forall a' \in A$
    }
    \If{\emph{doVI}}{
        $QT \gets \textit{VI}(QT, \tau_E, \rho_E, n_E, \tau_Q, \rho_Q, n_Q, \gamma, T)$ \\
    }
    \eIf{done}{
        $s,q \gets E.reset() $ \\
    }{ $s,q \gets s',q'$ }
}
{\bf return} $QT$
\end{scriptsize}
\end{algorithm}

Following \rmax's approach, \qrmax~(Algorithm \ref{alg:qrmax}) 
initializes the Q-table $QT$ with a maximum value (line 1) for every state-action pair (notice states now consist of $s$ and $q$) and 
selects the next action to execute as the $\argmax$ of the current Q-table (line 8). 
In lines 11--26, environment transitions and rewards ($\tau_E, \rho_E, n_E$) and automaton transitions and rewards ($\tau_Q, \rho_Q, n_Q$) are accumulated separately, using different thresholds, $t_E, t_Q$.
If any pair $(s,a)$ or $(q,s')$ reaches the corresponding target threshold (lines 15-17 and 23-25), a value iteration step ($VI$) is run 
for $T$ iterations (line 31) to update the Q-table values.
When an episode terminates, the Q-table values for the final states are set to zero (line 28) and the environment is reset (line 34).
For value iteration, a standard algorithm is adopted, which uses Equations \ref{eq:P} and \ref{eq:R} 
to describe the process and estimate the model functions as follows, considering only \emph{known} pairs:
\begin{footnotesize}
\[
\begin{aligned}
\hat{P}(s'|s,a) = \frac{\tau_E(s,a,s')}{n_E(s,a)}, ~~~
\hat{P}(q'|s',q) = \frac{\tau_Q(q,s',q')}{n_Q(q,s')}, \\
\hat{R}_E(s,a,s') = \frac{\rho_E(s,a,s')}{n_E(s,a)}, ~~
\hat{R}_Q(q,s',q') = \frac{\rho_Q(q,s',q')}{n_Q(q,s')}.
\end{aligned}
\]
\end{footnotesize}
\paragraph{Bellman Backup Used in VI.}
For known components, VI:
\begin{footnotesize}
\begin{multline}\label{eq:vi-backup}
Q_{k+1}(s,q,a)
= \sum_{s'} \hat{R}_E(s,a,s')
+ \sum_{s'} \hat{P}(s'|s,a) \\
\mathopen{}\Bigl(
\sum_{q'} \hat{R}_Q(q,s',q')
+ \gamma \sum_{q'} \hat{P}(q'|s',q)\max_{a'} Q_k(s',q',a')
\Bigr).
\end{multline}
\end{footnotesize}

Here $\hat R_E(s,a,s')$ and $\hat R_Q(q,s',q')$ are \emph{reward-mass} estimates:
$\hat R_E(s,a,s')=\hat P(s'|s,a)\,\hat r_E(s,a,s')$ and
$\hat R_Q(q,s',q')=\hat P(q'|s',q)\,\hat r_Q(q,s',q')$.
Thus Eq.~\eqref{eq:vi-backup} is identical to the canonical expected-reward backup, but written in mass form to avoid conditioning on rare successor events.

During value iteration we update $QT(s,q,a)$ only when
$\textsc{Known}_E(s,a)$ holds and, for every successor state $s'$ with
$\tau_E(s,a,s')>0$, also $\textsc{Known}_Q(q,s')$ holds.
All other entries retain their initial optimistic value $R_{\max}/(1-\gamma)$; equivalently, any unobserved successor probability mass is assigned to an absorbing optimistic state with value $R_{\max}/(1-\gamma)$ (standard \textsc{R-Max}), underpinning the PAC--MDP analysis in Section~4.2.


In contrast to RM-aware methods that assume direct access to the automaton transition function $\eta$,
\qrmax\ does not require $\eta$ as an input: it only assumes access to the automaton-state signal, i.e.,
the agent observes the current automaton state $q$ and, after each action, the updated state $q'$ together with the reward,
as formalized above.
If $q$ is not observable and the RM is unknown, the problem becomes partially observable; we leave automaton-state inference to future work.

When the DFA/RM is available, we can additionally generate counterfactual experience across automaton states and apply the
same factorized updates to it, obtaining the structure-aware variant \qrmaxrm, which we also evaluate experimentally.

\subsection{Theoretical Guarantees of \textsc{QR-Max}}\label{subsec:qrmax_theory}

Let \(X=S\times Q\), \(V_{\max}=R_{\max}/(1-\gamma)\), and \(U=|S||A|m_E+|S||Q|m_Q\), where \(U\) is the number of factorized components that can be unknown before becoming known. We assume one-step rewards are bounded in \([0,R_{\max}]\). For the PAC analysis, we denote the knownness thresholds by \(m_E=t_E\) and \(m_Q=t_Q\). The thresholds \(m_E,m_Q\) are chosen by standard concentration bounds and a union bound over the events for \(R_E,P_E,R_{\A}\), and \(P_Q\), so that, with probability at least \(1-\delta\), every known component induces reward error at most \(\xi_R\), environment-transition error at most \(\xi_E\), and automaton-transition error at most \(\xi_Q\). They are chosen so that \(2(\xi_R+\gamma V_{\max}(\xi_E+\xi_Q))/(1-\gamma)+\eta_{\mathrm{VI}}\le\epsilon\), where \(\eta_{\mathrm{VI}}\) is the value-iteration truncation error. For deterministic reward machines with deterministic transition rewards, one observation suffices for each observed \((q,s')\), hence \(m_Q=1\) and \(\xi_Q=0\) on known automaton components.

\begin{lemma}[Factored Accuracy]
On the above event, every known product component satisfies
\(|\widehat R-R|\le\xi_R\) and
\(\|\widehat P(\cdot\mid s,q,a)-P(\cdot\mid s,q,a)\|_1\le\xi_E+\xi_Q\).
Indeed, from Eq.~\eqref{eq:P},
\(\|\widehat P-P\|_1
\le
\|\widehat P_E(\cdot\mid s,a)-P_E(\cdot\mid s,a)\|_1
+
\sum_{s'}\widehat P_E(s'\mid s,a)
\|\widehat P_Q(\cdot\mid q,s')-P_Q(\cdot\mid q,s')\|_1
\le \xi_E+\xi_Q\).
\end{lemma}

\begin{lemma}[Near-Optimality in the Known Model]
Conditioned on Lemma~1, the greedy policy computed in the optimistic model is \(\epsilon\)-optimal whenever no unknown component is reached within the effective horizon. This follows from the discounted simulation bound \(\|\widehat{\mathcal T}V-\mathcal TV\|_{\infty}\le\xi_R+\gamma V_{\max}(\xi_E+\xi_Q)\), valid for all \(\|V\|_{\infty}\le V_{\max}\), and from the above choice of \(\xi_R,\xi_E,\xi_Q\), and \(\eta_{\mathrm{VI}}\).
\end{lemma}

\begin{theorem}[PAC-MDP Bound]\label{thm:qrmax_pac}
With the above thresholds, \textsc{QR-Max} is PAC-MDP. With probability at least \(1-\delta\), the number of interaction steps in which its greedy policy is not \(\epsilon\)-optimal is bounded by the standard optimistic-model escape bound \(O((HU/p)\log(U/\delta))\), where \(H=\lceil(1-\gamma)^{-1}\ln(2R_{\max}/[\epsilon(1-\gamma)])\rceil\), \(p=\epsilon(1-\gamma)/(2R_{\max})\), and \(U=|S||A|m_E+|S||Q|m_Q\). Equivalently, the leading unknown-component dependence is factorized rather than product-based.
\end{theorem}

\emph{Proof Sketch.}
Condition on the event of Lemma~1. If the optimistic greedy policy is not \(\epsilon\)-optimal, Lemma~2 implies that, by the standard \textsc{R-Max} escape-event argument~\cite{Brafman03,SLL09}, it must reach an unknown component within horizon \(H\) with probability at least \(p\). Each escape is charged to the first unknown component reached. Since environment components can be charged at most \(|S||A|m_E\) times and automaton components at most \(|S||Q|m_Q\) times, the standard PAC-MDP argument yields the stated bound. Value iteration affects computation time but does not consume environment interactions.

\medskip\noindent\paragraph{Product-Space \textsc{R-Max} Comparison.}
Applied directly to the product MDP over \(S\times Q\), \textsc{R-Max} learns unknown components indexed by \((s,q,a)\), giving a product-space count of order \(|S||Q||A|m\). In contrast, \textsc{QR-Max} learns the environment component once for each \((s,a)\) and the automaton component once for each \((q,s')\), giving the factorized count \(U=|S||A|m_E+|S||Q|m_Q\). Thus the expensive environment-transition estimation is not repeated for every automaton state. In deterministic reward machines, where \(m_Q=1\), the dominant learned component is typically the shared environment model, which gives the main sample-efficiency advantage over product-space \textsc{R-Max}.

A complete proof is provided in Appendix~\ref{app:qrmax_proofs}.


\section{Bucket-QR-Max: Continuous State Spaces}
\label{sec:bucketqrmax}

\qrmax\ assumes that the environment state component $s\in S$ is \emph{discrete}. 
In many real-world problems the agent instead observes a continuous vector $s^{\mathrm{cont}}\!\in\!\mathbb{R}^d$, while actions remain finite and discrete. A common way to handle continuous states is to discretise them into a finite set of ``buckets'' and treat each bucket as a discrete state. We follow this idea and integrate a hash-based discretisation into \qrmax, yielding \bqrmax, a model-based extension that retains the structural advantages of \qrmax\ (separate counts for Markovian dynamics and non-Markovian reward, optimistic value iteration) while handling continuous observations \emph{without neural approximation}. Under a bounded-discretisation assumption on the induced buckets (formalised as Assumption~\ref{ass:bounded_disc} in Appendix~\ref{app:bucket_pac}), we derive a conditional PAC-MDP bound for \bqrmax\ in the underlying continuous MDP (Theorem~\ref{thm:bucket_pac}).

\subsubsection{Locality-Sensitive Hashing Discretisation}
Let $h:\mathbb{R}^{d}\!\to\!\mathcal{B}$ be a family of
$L$ independent SimHash functions~\cite{Charikar2002},
$h(\cdot)=(h_{1}(\cdot),\dots,h_{L}(\cdot))$, where each
$h_{\ell}$ projects $s^{\text{cont}}$ onto
$\{\!0,\dots,2^{d_{h}}\!-\!1\}$.
Given $s^{\text{cont}}_{t}$ we obtain a \emph{bucket}
$b_t=h(s^{\text{cont}}_{t})\in\mathcal{B}\subset\mathbb{N}^{L}$.
The mapping is \emph{data–dependent}: new buckets are added on-the-fly
and assigned the next integer index
$\iota:\mathcal{B}\to\{0,\dots,B_t\}$.
A joint observation is therefore represented as
$(b,q)\in\mathcal{B}\times Q$.
The automaton component remains exact and finite, while the continuous
part is covered by a dynamically growing \emph{countable} partition.

\subsubsection{Bucket-QR-Max Counters and Reward Sums}
It keeps $n_{E_T}(b,a)$ for environment transitions, $n_{E_R}(b,a)$ and $R_E(b,a)$ for environment rewards, $n_{Q_T}(q,b')$ for reward-machine transitions, and $n_{Q_R}(q,b')$ and $R_Q(q,b')$ for non-Markovian rewards. The corresponding knownness conditions are
{\footnotesize
\(
(n_{E_T}(b,a),n_{E_R}(b,a))\ge(t_{E_T},t_{E_R}),\;
(n_{Q_T}(q,b'),n_{Q_R}(q,b'))\ge(t_{Q_T},t_{Q_R})
\).
}
Only when \emph{all four} conditions are met is the triple $(b,a,q)$ marked \emph{known} and Value Iteration triggered. The thresholds \(t_{E_T},t_{E_R},t_{Q_T},t_{Q_R}\) are chosen by the same PAC-MDP concentration argument as in Sec.~\ref{subsec:qrmax_theory}, applied to the induced bucket MDP. The agent maintains an \emph{optimistic} Q-table \(Q[b,q,a]=R_{\max}/(1-\gamma)\) for every discovered bucket $b$ and runs synchronous Value Iteration over the finite abstraction on \(\iota(\mathcal{B})\times Q\) until convergence up to $\varepsilon_{\text{VI}}$. The tabular solver therefore keeps \qrmax's factorised counts and an optimistic value function over $\mathcal{O}(|\mathcal{B}_t||Q||A|)$ entries.

\subsubsection{Exploration and Optimism}
Because the hash granularity is fixed \emph{a priori}, each bucket $b$
plays the role of a discrete state $s$ in \qrmax: the same factorised
counters and model estimates $(P_E,P_Q,R_E,R_Q)$ are reused, and the
Q-table remains optimistic at $R_{\max}/(1-\gamma)$ until $(b,a,q)$
becomes known.  Thus \bqrmax\ inherits exactly the same
optimistic-exploration mechanism as the discrete algorithm.

\subsubsection{Design Intuition and Conditional PAC Guarantee}
For any fixed finite partition $\mathcal{B}$ of the state space,
\bqrmax\ reduces to running \qrmax\ on the abstract
MDP with states $(b,q)\in\mathcal{B}\times Q$.  In Appendix~\ref{app:bucket_pac}, we formalise a bounded-discretisation
assumption (small bucket diameter and Lipschitz-continuous dynamics and
rewards, see Assumption~\ref{ass:bounded_disc}) and show that, under this
assumption, the induced bucket MDP is a close approximation of the
underlying continuous MDP.  The resulting conditional
PAC-MDP bound (Theorem~\ref{thm:bucket_pac}) mirrors the discrete bound in
Theorem~\ref{thm:qrmax_pac}, with $|S|$ replaced by $|\mathcal{B}|$.
Intuitively, when nearby continuous states tend to fall into the same
bucket and the environment is sufficiently smooth, an $\varepsilon$-optimal
policy for the bucket MDP is also near-optimal in the original
continuous MDP.  In practice, SimHash does not deterministically bound bucket diameters; our PAC guarantee is conditional on the induced partition satisfying the bounded-diameter assumption, and hash collisions can be absorbed into an additional failure probability $\delta_{\mathrm{LSH}}$ (overall success at least $1-(\delta+\delta_{\mathrm{LSH}})$), see Appendix~\ref{app:bucket_pac} for details.

\subsubsection{Relation to Bucket-R-Max and Implementation.}
The \emph{Covering-RMAX} algorithm replaces exact states in \rmax\ with a balls-in-metric-space partition~\cite{bernstein2008adaptive}.
\bqrmax\ differs in two essential ways:
\begin{compactenum}
    \item \textbf{Hash vs.\ metric balls.}
          LSH only probabilistically groups nearby states: by increasing
          the number of hash functions and projections, the probability
          that distant states collide in the same bucket can be made
          small.  SimHash needs \emph{no} explicit metric and updates in
          $O(L)$ time, whereas searching the nearest-centre set grows
          with the cover size. In high-dimensional spaces, Covering-RMAX
          can suffer from the curse of dimensionality, as the number of
          metric balls required grows with the dimension, whereas for a
          fixed hash family SimHash maintains a fixed per-update
          computational cost.
    \item \textbf{Decoupled NM-reward.}
          We preserve the \qrmax\ factorization, yielding a sample
          complexity in
          $| \mathcal{B}|\,|A| + |\mathcal{B}|\,|Q|$
          instead of the cubic
          $|\mathcal{B}|\,|Q|\,|A|$ inherited by a naive continuous
          \rmax\ reduction.
\end{compactenum}

\bqrmax\ differs from discrete \qrmax\ in exactly three lines:
(i) \emph{LSH discretisation} $b\leftarrow h(s^{\text{cont}})$;
(ii) four separate counters $n_{E_T},n_{E_R},n_{Q_T},n_{Q_R}$, each with
its own threshold; only when all four exceed their thresholds is
$(b,a,q)$ marked known and VI invoked; (iii) all counters, model estimates, and Q-values are indexed by
buckets $b$ instead of discrete states $s$.
Full pseudocode is reported in Algorithm~\ref{alg:bucket_qrmax} in Appendix~\ref{app:bucket_pac}.

\section{Experiments}
\label{sec:experiments}

In this section, we compare \qrmax\ and \qrmaxrm\ against several state‑of‑the‑art algorithms, on different environments of increasing difficulty.\footnote{Code: \url{https://github.com/Alee08/qrmax}.} 
Our goal is to show that the factorization introduced in \qrmax\ enhances sample efficiency and enables optimal solution of complex tasks, confirming the theoretical results obtained in Sec.~\ref{sec:factorization}.

The compared methods include:
1) model-based methods for MDPs over the joint space $S \times Q$:
\textsc{UCBVI‑sB} (simplified Bernstein bonus), \textsc{UCBVI‑H} (Hoeffding bonus), and \textsc{UCBVI‑B} (Bernstein bonus) from \cite{azar17a}; \textsc{PSRL} \cite{osband2013}; \textsc{OPSRL} \cite{tiapkin2022optimistic}; \rmax~\cite{Brafman03};
2) the model-free method \textsc{QRM}~\cite{pmlr-v80-icarte18a} for NMRDP.
We do not compare directly  with~\cite{GaonB20}, which focuses on learning the RM, since when this is available, the approach corresponds to \rmax, which we evaluate. Also, as already mentioned,~\cite{ShaoK23} is indirectly compared, corresponding, on the problems considered, to~\textsc{QRM}.
For clarity, \qrmax\ uses only the observed automaton-state signal ($q$) and does not assume access to $\eta$, whereas \qrmaxrm\ assumes the RM is given (including $\eta$) and uses it for counterfactual updates across automaton states.

Unless stated otherwise, we use the authors' recommended hyper-parameters.
For finite-horizon baselines that require a horizon (\textsc{UCBVI, PSRL/OPSRL}), we set $H$ to the VI-optimal path length plus a small safety margin (Appendix~\ref{app:experiments}); this uses oracle knowledge to choose a favorable $H$ for these baselines, hence the comparison is conservative w.r.t.\ \qrmax. The benchmark we use for experimental comparison takes inspiration from, and extends, the Office gridworld environment~\cite{pmlr-v80-icarte18a} (full details are reported in Appendix~\ref{app:experiments}),
used to evaluate NMRDP solutions.
In all the experiments, during training, we use stochastic actions, all having a nominal outcome, i.e., one  with highest probability wrt all the others.

For each experimental configuration, we compute a VI baseline by running value iteration on the full MDP model (obtained from the environment). Periodically we compare the current learned policy (greedy w.r.t. the Q-
table) to the VI policy by running 10,000 evaluation episodes per policy in the stochastic environment and collecting discounted episode returns. We apply a two-sided Welch t-test to the two return samples.
With early stopping enabled, in stochastic environments training terminates when the Welch t-test fails to detect a significant difference at a 10\% significance level \((p\text{-value} \ge 0.1)\), treating the two policies as
statistically indistinguishable. In deterministic settings the t-test is skipped due to zero variance. We apply this rule uniformly; additional Appendix curves (e.g., fixed-budget success rates) confirm the same ordering.
For visualization purposes, we run the learned policies on the corresponding deterministic testing environments, where actions always produce their nominal outcome. This simplifies comparing the learned policies with VI solutions, in terms of required timesteps to reach the goal.
We report mean $\pm$ std over 10 seeds on three configurations.

\begin{figure}[t]
  \centering
  \includegraphics[width=\columnwidth]{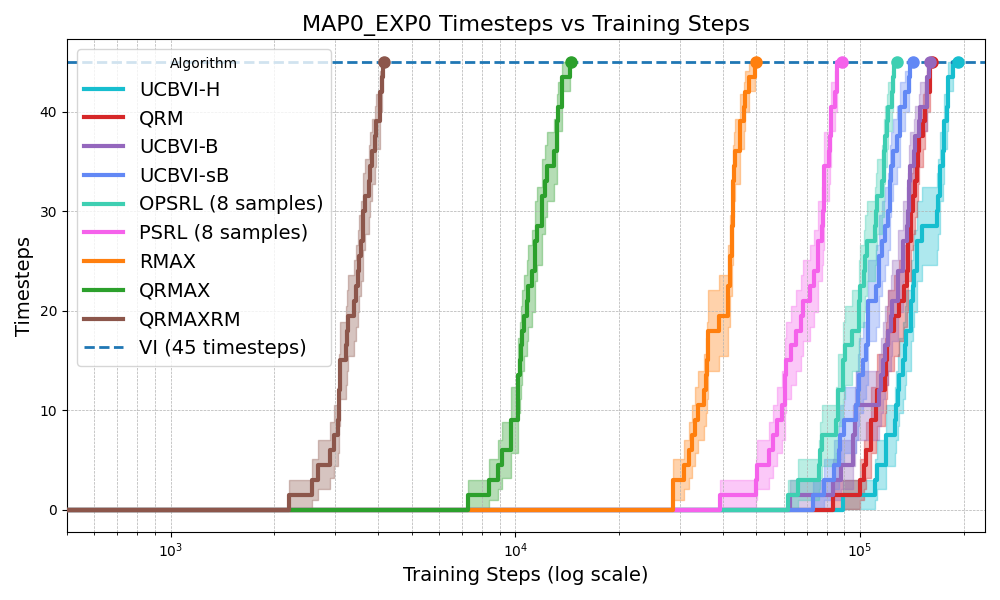}
  \caption{Configuration 1 (map0 exp0).}
  \label{fig:map0_exp0}
\end{figure}

\paragraph{Configuration 1 — \texttt{map0\_exp0} (\(10\times10\)).}
We use the simple grid world from prior work~\cite{tiapkin2022optimistic}, using \(H=50\) and eight Thompson samples, following the original papers.
The agent's non-Markovian task consists in starting from the top‑left cell,  collecting a letter in the bottom‑right cell, returning to the start position to pick up a coffee, and finally delivering both items to the office in the bottom‑left cell. 
The issued reward is \(1\) on task  completion, \(0\) otherwise.
Figure~\ref{fig:map0_exp0} shows the success rate: \qrmaxrm\ and \qrmax\ achieve significantly better performance in roughly one order of magnitude fewer training steps than \textsc{UCBVI}, \textsc{PSRL}, and \textsc{OPSRL}, confirming that the factorised dynamics improves sample efficiency.

\begin{figure}[t]
  \centering
  \includegraphics[width=\columnwidth]{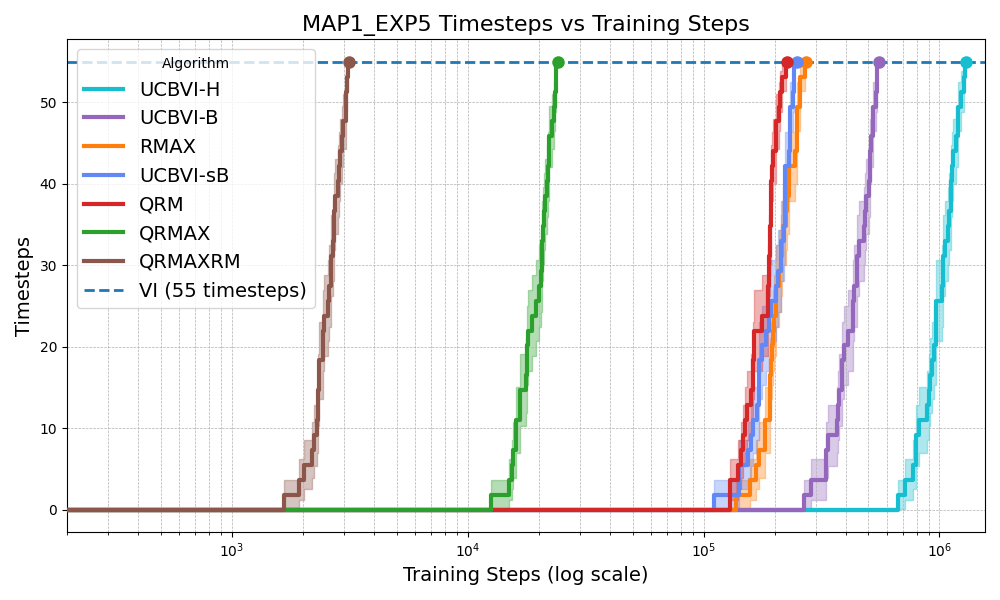}
  \caption{Configuration 2 (map1 exp5).}
  \label{fig:map1_exp5}
\end{figure}

\paragraph{Configuration 2 — \texttt{map1\_exp5} (Office World, \(12\times9\)).}
This map contains walls and ornamental objects that impose a penalty of \(-100\) upon contact.  
For \textsc{UCBVI} and its variants we set \(H = 150\).  
As illustrated in Figure~\ref{fig:map1_exp5}, as task complexity grows, \qrmax\ and \qrmaxrm\  maintain a gap of nearly two orders of magnitude over \textsc{UCBVI‑H} and \textsc{UCBVI‑B}, and about one order over \textsc{QRM} (which has access to the RM), \rmax, and \textsc{UCBVI‑sB}.  
\textsc{PSRL} and \textsc{OPSRL} exceed the predefined memory/CPU budget on this instance (see Appendix~\ref{app:experiments} for the resource limits).

\begin{figure}[t]
  \centering
  \includegraphics[width=\columnwidth]{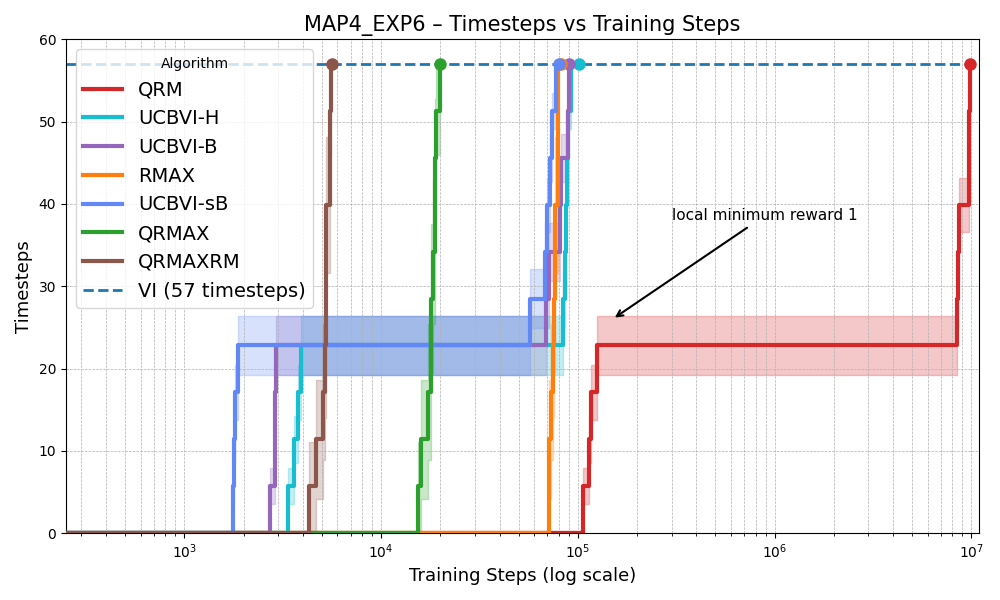}
  \caption{Configuration 3 (map4 exp6).}
  \label{fig:map4_exp7}
\end{figure}

\paragraph{Configuration 3 — \texttt{map4\_exp6} (\(15\times15\)).}
The third configuration is a maze-like space featuring two coffee machines: a \emph{good} one (reward \(1000\)) and a \emph{regular} one (reward \(1\)).  
The agent receives a reward only after picking up a coffee and delivering it to the office.  
Figure~\ref{fig:map4_exp7} shows that \qrmax\ and \qrmaxrm\ converge to the optimal policy, bringing the better coffee in \(57\) steps, whereas \textsc{QRM}, \textsc{UCBVI} and their variants remain trapped for many training steps in the local optimum associated with the nearer coffee machine.

\paragraph{Evaluation of \bqrmax.}
We evaluate \bqrmax on a continuous-state variant of
Frozen Lake and Office World (5 subtasks), using SimHash-based discretizations
and the same factorized counting scheme as in the discrete case
(see Appendix for details). As baselines, we train a DQN on the
same inputs $(s^{\mathrm{cont}}, q)$ and sparse non-Markovian reward,
with a standard two-layer MLP and replay buffer, and we also include
a controlled \textsc{R-Max} variant that uses the same SimHash buckets
but learns a monolithic model over joint states $(b,q)$ (see Appendix).
In this setting \bqrmax reaches near-optimal performance in
$50 \pm 10$ episodes, over an order of magnitude faster than DQN
(which requires $>500$ episodes) and with lower variance,
indicating that combining SimHash discretizations with separate
environment and automaton counters yields fast, stable learning
in continuous state spaces without manual gridding or neural
function approximation.

\paragraph{Summary of the Results.}
The experimental analysis demonstrates that, by exploiting model-based RL and the proposed factorization, \qrmax, \qrmaxrm, and \bqrmax consistently achieve better sample efficiency than both model-free baselines and model-based counterparts without factorization on all the scenarios we consider. 
Wall-clock times remained comparable to UCBVI within the same resource limits (see Appendix).
We also observe different impacts of hyper-parameter tuning on solution quality. The main hyper-parameters of \qrmax and variants are the thresholds $t_E$ and $t_Q$, which control the number of samples needed to reliably estimate transitions and rewards. We set \(t_E=39\) empirically and \(t_Q=1\) because the considered RMs are deterministic. In practice, $t_E$ is relatively easy to set because performance is weakly sensitive beyond the threshold, and training steps scale roughly linearly with $t_E$.
In contrast, the hyper-parameters for the other methods are significantly more difficult to tune and yield higher result sensitivity. For instance, model-free RL algorithms require setting the learning rate, exploration factors, and initial $Q$-table values, resulting in high impact on the solutions; moreover, such parameters cannot easily be related to the specific problem.
The other model-based RL algorithms considered in this paper include parameters such as the horizon $H$, whose setup requires both domain and task knowledge, and the Thompson-sampling count, which has a similar meaning of $t_E$ and $t_Q$, but a higher impact especially on the algorithm's memory requirements. 



\begin{table}[t]
\centering
\setlength\tabcolsep{4pt}
\resizebox{\columnwidth}{!}{%
\begin{tabular}{llrrrrrrrr}
\toprule
\textbf{Map} & \textbf{Exp} &
\textbf{QR‑MAX} & \textbf{QR‑MAXRM} & \textbf{R‑MAX} & \textbf{QRM} &
\textbf{UCBVI‑sB} & \textbf{UCBVI‑B} & \textbf{UCBVI‑H} &
\textbf{PSRL/OPSRL$^{\dagger}$} \\
\midrule
MAP0 & EXP0 &
\textbf{14\,483} & \textbf{4\,150} & 49\,759 & 162\,040 &
142\,015 & 159\,092 & 191\,983 & 88\,437 / 128\,188 \\[2pt]
MAP1 & EXP5 &
\textbf{24\,222} & \textbf{3\,125} & 272\,080 & 225\,140 &
250\,800 & 555\,977 & 1\,320\,070 & --- \\[2pt]
MAP4 & EXP6 &
\textbf{20\,150} & \textbf{5\,630} & 82\,000 & 9\,910\,340 &
80\,160 & 90\,030 & 101\,120 & --- \\
\bottomrule
\multicolumn{10}{p{0.96\linewidth}}{
$^{\dagger}$Eight posterior samples per episode, as in prior work~\cite{tiapkin2022optimistic}.}
\end{tabular}%
}
\caption{Steps to a VI-optimal policy (mean over 10 seeds).}
\label{tab:steps-map-exp}
\end{table}

\section{Conclusions}
In this work, we introduced \qrmax, the first model-based RL algorithm for discrete-action NMRDPs that exploits the decoupling between Markovian environment dynamics and non-Markovian rewards. We prove that \qrmax is PAC-MDP, i.e., all but a polynomial number of interaction steps are $\varepsilon$-optimal with high probability, and show on an extended Office grid-world benchmark that it substantially reduces the required interactions compared to both model-free and standard model-based approaches.
We also described \bqrmax, which leverages SimHash to handle continuous-state NMRDPs without manual gridding or neural approximators; under mild smoothness and discretization assumptions, it admits a conditional PAC-MDP guarantee (see Appendix).
Across our benchmarks, \qrmax and its variants exhibit higher sample efficiency and scalability, reduced computational cost, and easier hyper-parameter tuning than the evaluated baselines.
Future directions include extending the factorization framework to deep RL techniques, handling stochastic RMs~\cite{Corazza_Gavran_Neider_2022}, learning reward models automatically, and extending the approach to multi-agent scenarios, where reward machines have already been used for cooperative and concurrent multi-agent RL~\cite{Neary2021RewardMachines,Trapasso2025MAPRL}.

\FloatBarrier
\section*{Acknowledgments}
This research is supported by the PNRR MUR project PE0000013-
FAIR.


\bibliographystyle{named}
\bibliography{aaai25}

@inproceedings{BacchusBG96,
	author = {Fahiem Bacchus and Craig Boutilier and Adam J. Grove},
	title = {Rewarding Behaviors},
        area = {Planning (including under Uncertainty)},
	booktitle = {Proc. AAAI},
	year = {1996},
	pages = {1160-1167},
}

@inproceedings{BDGP2018, 
    title={{LTLf/LDLf} Non-{Markovian} Rewards}, 
    booktitle={Proc. AAAI}, 
    author={Brafman, Ronen and De Giacomo, Giuseppe and Patrizi, Fabio}, 
    year={2018},
    pages = {1771-1778},
}

@book{sutton, 
  title={Reinforcement learning: An introduction}, 
  author={Sutton, Richard S and Barto, Andrew G}, 
  year={2018}, 
  publisher={MIT press} 
}

@inproceedings{GaonB20,
  author       = {Maor Gaon and
                  Ronen I. Brafman},
  title        = {Reinforcement Learning with Non-{Markovian} Rewards},
  booktitle    = {Proc. AAAI},
  pages        = {3980--3987},
  year         = {2020},
}

@article{Brafman03,
author = {Brafman, Ronen I. and Tennenholtz, Moshe},
title = {R-max - a general polynomial time algorithm for near-optimal reinforcement learning},
year = {2003},
volume = {3},
journal = {J. Mach. Learn. Res.},
pages = {213–231},
}

@article{pmlr-v80-icarte18a,
  author       = {Rodrigo {Toro Icarte} and
                  Toryn Q. Klassen and
                  Richard Anthony Valenzano and
                  Sheila A. McIlraith},
  title        = {Reward Machines: Exploiting Reward Function Structure in Reinforcement
                  Learning},
  journal      = {J. Artif. Intell. Res.},
  volume       = {73},
  pages        = {173--208},
  year         = {2022},
}

@inproceedings{Gupta2021NonMarkovianRL,
  title={Non-{Markovian} Reinforcement Learning using Fractional Dynamics},
  author={Gaurav Gupta and Chenzhong Yin and Jyotirmoy V. Deshmukh and Paul Bogdan},
  booktitle={Proc. CDC},
  year={2021},
  pages={1542-1547},
}

@Article{watkins1992,
author="Watkins, Christopher and Dayan, Peter",
title="Q-learning",
journal="Mach. Learn.",
year="1992",
volume="8",
number="3",
pages="279--292",
}

@inproceedings{Corazza_Gavran_Neider_2022, 
    title={Reinforcement Learning with Stochastic Reward Machines}, 
    booktitle={Proc. AAAI}, 
    author={Corazza, Jan and Gavran, Ivan and Neider, Daniel}, 
    year={2022}, 
    pages={6429-6436} 
}

@inproceedings{de2020restraining,
  title = {Restraining bolts for Reinforcement Learning agents},
  author = {De Giacomo, Giuseppe and Iocchi, Luca and Favorito, Marco and Patrizi, Fabio},
  booktitle = {Proc, ICAPS},
  volume = {34},
  pages = {13659--13662},
  year = {2020},
}

@article{TGSPK06,
  author       = {Sylvie Thi{\'{e}}baux and
                  Charles Gretton and
                  John K. Slaney and
                  David Price and
                  Froduald Kabanza},
  title        = {Decision-Theoretic Planning with non-Markovian Rewards},
  journal      = {J. Artif. Intell. Res.},
  volume       = {25},
  pages        = {17--74},
  year         = {2006},
}

@article{SLL09,
  author       = {Alexander L. Strehl and
                  Lihong Li and
                  Michael L. Littman},
  title        = {Reinforcement Learning in Finite MDPs: {PAC} Analysis},
  journal      = {J. Mach. Learn. Res.},
  volume       = {10},
  pages        = {2413--2444},
  year         = {2009},
  url          = {https://dl.acm.org/doi/10.5555/1577069.1755867},
  doi          = {10.5555/1577069.1755867},
  bibsource    = {dblp computer science bibliography, https://dblp.org}
}

@inproceedings{Charikar2002,
author = {Charikar, Moses S.},
title = {Similarity estimation techniques from rounding algorithms},
year = {2002},
isbn = {1581134959},
publisher = {Association for Computing Machinery},
address = {New York, NY, USA},
url = {https://doi.org/10.1145/509907.509965},
doi = {10.1145/509907.509965},
booktitle = {Proceedings of the Thiry-Fourth Annual ACM Symposium on Theory of Computing},
pages = {380–388},
numpages = {9},
location = {Montreal, Quebec, Canada},
series = {STOC '02}
}

@inproceedings{bernstein2008adaptive,
  title={Adaptive Aggregation for Reinforcement Learning with Efficient Exploration: Deterministic Domains.},
  author={Bernstein, Andrey and Shimkin, Nahum},
  booktitle={COLT},
  pages={323--334},
  year={2008},
  organization={Citeseer}
}

@inproceedings{ShaoK23,
  author       = {Daqian Shao and
                  Marta Kwiatkowska},
  title        = {Sample Efficient Model-free Reinforcement Learning from {LTL} Specifications
                  with Optimality Guarantees},
  booktitle    = {Proceedings of the Thirty-Second International Joint Conference on
                  Artificial Intelligence, {IJCAI} 2023, 19th-25th August 2023, Macao,
                  SAR, China},
  pages        = {4180--4189},
  publisher    = {ijcai.org},
  year         = {2023},
  url          = {https://doi.org/10.24963/ijcai.2023/465},
  doi          = {10.24963/IJCAI.2023/465},
  bibsource    = {dblp computer science bibliography, https://dblp.org}
}

@InProceedings{azar17a,
  title = 	 {Minimax Regret Bounds for Reinforcement Learning},
  author =       {Mohammad Gheshlaghi Azar and Ian Osband and R{\'e}mi Munos},
  booktitle = 	 {Proceedings of the 34th International Conference on Machine Learning},
  pages = 	 {263--272},
  year = 	 {2017},
  editor = 	 {Precup, Doina and Teh, Yee Whye},
  volume = 	 {70},
  series = 	 {Proceedings of Machine Learning Research},
  month = 	 {06--11 Aug},
  publisher =    {PMLR},
  url = 	 {https://proceedings.mlr.press/v70/azar17a.html}
}

@misc{osband2013,
      title={(More) Efficient Reinforcement Learning via Posterior Sampling}, 
      author={Ian Osband and Daniel Russo and Benjamin Van Roy},
      year={2013},
      eprint={1306.0940},
      archivePrefix={arXiv},
      primaryClass={stat.ML},
      url={https://arxiv.org/abs/1306.0940}, 
}

@inproceedings{
tiapkin2022optimistic,
title={Optimistic Posterior Sampling for Reinforcement Learning with Few Samples and Tight Guarantees},
author={Daniil Tiapkin and Denis Belomestny and Daniele Calandriello and Eric Moulines and Remi Munos and Alexey Naumov 
and Mark Rowland and Michal Valko and Pierre Menard},
booktitle={Advances in Neural Information Processing Systems},
editor={Alice H. Oh and Alekh Agarwal and Danielle Belgrave and Kyunghyun Cho},
year={2022},
url={https://openreview.net/forum?id=gvwDosudtyA}
}

@inproceedings{Strens2000,
author = {Strens, Malcolm J. A.},
title = {A Bayesian Framework for Reinforcement Learning},
year = {2000},
isbn = {1558607072},
publisher = {Morgan Kaufmann Publishers Inc.},
address = {San Francisco, CA, USA},
booktitle = {Proceedings of the Seventeenth International Conference on Machine Learning},
pages = {943–950},
numpages = {8},
series = {ICML '00}
}

@article{Guestrin_2003,
   title={Efficient Solution Algorithms for Factored MDPs},
   volume={19},
   ISSN={1076-9757},
   url={http://dx.doi.org/10.1613/jair.1000},
   DOI={10.1613/jair.1000},
   journal={Journal of Artificial Intelligence Research},
   publisher={AI Access Foundation},
   author={Guestrin, C. and Koller, D. and Parr, R. and Venkataraman, S.},
   year={2003},
   month=oct, pages={399–468} }

@misc{osband2014nearoptimalreinforcementlearningfactored,
      title={Near-optimal Reinforcement Learning in Factored MDPs}, 
      author={Ian Osband and Benjamin Van Roy},
      year={2014},
      eprint={1403.3741},
      archivePrefix={arXiv},
      primaryClass={stat.ML},
      url={https://arxiv.org/abs/1403.3741}, 
}

@inproceedings{Strehl2007FactoredRMax,
  author    = {Strehl, Alexander L.},
  title     = {Model-Based Reinforcement Learning in Factored-State MDPs},
  booktitle = {Proceedings of the 2007 IEEE International Symposium on Approximate Dynamic Programming and Reinforcement Learning},
  pages     = {103--110},
  year      = {2007},
  publisher = {IEEE},
  doi       = {10.1109/ADPRL.2007.368176}
}

@inproceedings{10.5555/1624312.1624325,
author = {Kearns, Michael and Koller, Daphne},
title = {Efficient reinforcement learning in factored MDPs},
year = {1999},
publisher = {Morgan Kaufmann Publishers Inc.},
address = {San Francisco, CA, USA},
booktitle = {Proceedings of the 16th International Joint Conference on Artificial Intelligence - Volume 2},
pages = {740–747},
numpages = {8},
location = {Stockholm, Sweden},
series = {IJCAI'99}
}

@article{Lin_Zhang_2025, title={Efficient Reinforcement Learning in Probabilistic Reward Machines}, volume={39}, url={https://ojs.aaai.org/index.php/AAAI/article/view/34061}, DOI={10.1609/aaai.v39i18.34061}, abstractNote={In this paper, we study reinforcement learning in Markov Decision Processes with Probabilistic Reward Machines (PRMs), a form of non-Markovian reward commonly found in robotics tasks. We design an algorithm for PRMs that achieves a regret bound of Õ((HOAT)^(1/2) + H²O²A^(3/2) + H(T)^(1/2)), where H is the time horizon, O is the number of observations, A is the number of actions, and T is the number of time steps. This result improves over the best-known bound, Õ(H(OAT)^(1/2)), for MDPs with Deterministic Reward Machines (DRMs), a special case of PRMs. When T ≥ H³O³A² and OA ≥ H, our regret bound leads to a regret of Õ((HOAT)^(1/2)), which matches the established lower bound of Ω((HOAT)^(1/2)) for MDPs with DRMs up to a logarithmic factor. To the best of our knowledge, this is the first efficient algorithm for PRMs. Additionally, we present a new simulation lemma for non-Markovian rewards, which enables reward-free exploration for any non-Markovian reward given access to an approximate planner.
Complementing our theoretical findings, we show through extensive experimental evaluations that our algorithm indeed outperforms prior methods in various PRM environments.}, number={18}, journal={Proceedings of the AAAI Conference on Artificial Intelligence}, author={Lin, Xiaofeng and Zhang, Xuezhou}, year={2025}, month={Apr.}, pages={18728-18736} }

@article{DBLP:journals/corr/abs-2412-19194,
  author       = {Hippolyte Bourel and
                  Anders Jonsson and
                  Odalric{-}Ambrym Maillard and
                  Chenxiao Ma and
                  Mohammad Sadegh Talebi},
  title        = {Provably Efficient Exploration in Reward Machines with Low Regret},
  journal      = {CoRR},
  volume       = {abs/2412.19194},
  year         = {2024},
  url          = {https://doi.org/10.48550/arXiv.2412.19194},
  doi          = {10.48550/ARXIV.2412.19194},
  eprinttype    = {arXiv},
  eprint       = {2412.19194},
  bibsource    = {dblp computer science bibliography, https://dblp.org}
}

@inproceedings{Bourel2023,
 author = {Hippolyte Bourel and Anders Jonsson and Odalric-Ambrym Maillard and Mohammad Sadegh Talebi},
 booktitle = {International Conference on Artificial Intelligence and Statistics (AISTATS)},
 pages = {4114–4146},
 title = {{Exploration in Reward Machines with Low Regret}},
 url_paper = {https://proceedings.mlr.press/v206/bourel23a/bourel23a.pdf},
 year = {2023}
}

@inproceedings{Trapasso2025MAPRL,
  author    = {Trapasso, Alessandro and Jonsson, Anders},
  title     = {Concurrent Multiagent Reinforcement Learning with Reward Machines},
  booktitle = {{ECAI} 2025 - 28th European Conference on Artificial Intelligence},
  series    = {Frontiers in Artificial Intelligence and Applications},
  volume    = {413},
  pages     = {3735--3742},
  publisher = {IOS Press},
  year      = {2025},
  doi       = {10.3233/FAIA251253},
}

@inproceedings{Neary2021RewardMachines,
  author    = {Cyrus Neary and Zhe Xu and Bo Wu and Ufuk Topcu},
  title     = {Reward Machines for Cooperative Multi-Agent Reinforcement Learning},
  booktitle = {{AAMAS} '21: 20th International Conference on Autonomous Agents and Multiagent Systems,
               Virtual Event, United Kingdom, May 3--7, 2021},
  pages     = {934--942},
  publisher = {{ACM}},
  year      = {2021},
  doi       = {10.5555/3463952.3464063},
  url       = {https://www.ifaamas.org/Proceedings/aamas2021/pdfs/p934.pdf}
}

@article{cipollone_ras_2025,
	title = {Exploiting robot abstractions in episodic {RL} via reward shaping and heuristics},
	volume = {193},
	issn = {09218890},
	doi = {10.1016/j.robot.2025.105116},
	journal = {Robotics and Autonomous Systems},
	author = {Cipollone, Roberto and Favorito, Marco and Maiorana, Flavio and De Giacomo, Giuseppe and Iocchi, Luca and Patrizi, Fabio},
	month = nov,
	year = {2025},
	pages = {105116},
}
\clearpage

\appendix

\makeatletter
\@addtoreset{theorem}{section}
\@addtoreset{equation}{section}
\makeatother

\renewcommand{\thetheorem}{\thesection.\arabic{theorem}}
\renewcommand{\theequation}{\thesection.\arabic{equation}}

\section{Detailed Proofs for \textsc{QR-Max}}
\label{app:qrmax_proofs}
We give the full proofs of Lemmas A.1–A.2 and Theorem A.3 that were stated without proof in Section 4.2 of the main paper.

\subsection{Notation and Preliminaries}
Let $N_E=|S|\,|A|$ and $N_Q=|S|\,|Q|$.  For every environment state-action pair $(s,a)$ and next state $s'$, define:
\begin{align*}
\hat P_E(s'\mid s,a)&=\frac{\tau_E(s,a,s')}{n_E(s,a)}, &\hat R_E(s,a,s')&=\frac{\rho_E(s,a,s')}{n_E(s,a)},
\\
\hat P_Q(q'\mid q,s')&=\frac{\tau_Q(q,s',q')}{n_Q(q,s')}, &\hat R_Q(q,s',q')&=\frac{\rho_Q(q,s',q')}{n_Q(q,s')}.
\end{align*}
All other notation follows Algorithm~1 in the main text.

\subsection{Proof of Lemma~A.1 (Accurate Estimates)}
$\beta = \epsilon(1-\gamma)/4$
and define the thresholds $m_E$ and $m_Q$ accordingly.
\begin{lemma}[Factored Accuracy]
\label{lem:factored_accuracy_full}
With probability at least $1-\delta$, whenever QR-MAX marks a pair known (i.e., $(s,a)$ after $m_E$ visits or $(q,s')$ after $m_Q$ visits), the following simultaneously hold for all $(s,a)$ and $(q,s')$:
\begin{equation}\label{eq:bounds}
\begin{aligned}
  |\hat R_E(s,a) - \bar R_E(s,a)| 
  &\le \beta,\\
  \|\hat P_E(\cdot\mid s,a) - P_E(\cdot\mid s,a)\|_1
  &\le \frac{\beta}{R_{\max}},\\
  |\hat R_Q(q,s') - \bar R_Q(q,s')|
  &\le \beta,\\
  \|\hat P_Q(\cdot\mid q,s') - P_Q(\cdot\mid q,s')\|_1
  &\le \frac{\beta}{R_{\max}}.
\end{aligned}
\end{equation}

\end{lemma}
\begin{proof}
For reward estimation in $[0,R_{\max}]$, Hoeffding's inequality implies that for $m_E$ i.i.d. samples:
\[
\Pr\Bigl[\bigl|\tfrac1{m_E}\sum_i r_i - \mathbb E[r]\bigr|>\beta\Bigr]
\le 2\exp\bigl(-2m_E\beta^2/R_{\max}^2\bigr).
\]
Setting $m_E=\lceil8R_{\max}^2\beta^{-2}\ln(2N_E/\delta)\rceil$ bounds this by $\delta/(2N_E)$.  A union bound over all $N_E$ pairs ensures all reward estimates are within $\beta$ with probability at least $1-\delta/2$.

For transition estimation in $\ell_1$, Weissman's inequality states that for a distribution over $|S|$ outcomes and $m_E$ samples:
\[
\Pr\bigl[\|\hat P_E - P_E\|_1>\epsilon_P\bigr]
\le (2^{|S|}-2)\exp\bigl(-m_E\epsilon_P^2/2\bigr).
\]
Choosing $\epsilon_P=\beta/R_{\max}$ and requiring $(2^{|S|})\exp(-m_E\beta^2/(2R_{\max}^2))\le\delta/(2N_E)$ yields a similar threshold on $m_E$.  The same argument applies to the $(q,s')$ counts with $m_Q,N_Q$.  A final union bound gives overall failure probability at most $\delta$.
\end{proof}

\subsection{Proof of Lemma~A.2 (Near-Optimal Optimistic Policy)}
\begin{lemma}[Near-Optimality in the Known Model]
\label{lem:known_model_full}
Conditioned on Lemma~\ref{lem:factored_accuracy_full}, the greedy policy
computed in the optimistic model is \(\epsilon\)-optimal whenever no
unknown component is reached within the effective horizon \(H\).
\end{lemma}
\begin{proof}
Condition on the event of Lemma~\ref{lem:factored_accuracy_full}.
Consider an execution of the greedy policy in which no unknown
environment or automaton component is reached within the effective
horizon \(H\). Along this event, all Bellman backups involve only known
components. Hence the empirical factorised model and the true model
differ only by the reward and transition errors controlled in
Lemma~\ref{lem:factored_accuracy_full}.

For any value function \(V\) with \(\|V\|_\infty\le V_{\max}\), the
one-step simulation bound gives
\[
\|\widehat T V - T V\|_\infty
\le
\xi_R+\gamma V_{\max}(\xi_E+\xi_Q),
\]
where \(\xi_R,\xi_E,\xi_Q\) are the reward, environment-transition, and
automaton-transition errors. By the choice of the thresholds and of the
value-iteration accuracy \(\eta_{\mathrm{VI}}\), the accumulated
discounted error over the effective horizon is at most \(\epsilon\).
Therefore, conditional on not reaching an unknown component within
\(H\), the greedy policy computed in the optimistic known model is
\(\epsilon\)-optimal in the true model.
\end{proof}

\subsection{Proof of Theorem~\ref{thm:qrmax_pac}}

\begin{theorem}[PAC-MDP Bound for \textsc{QR-Max}]
\label{thm:qrmax_pac_full}
Let
\[
U = |S||A|m_E + |S||Q|m_Q.
\]
With the thresholds chosen as in Lemma~\ref{lem:factored_accuracy_full},
\textsc{QR-Max} is PAC-MDP. With probability at least \(1-\delta\), the
number of interaction steps in which its greedy policy is not
\(\epsilon\)-optimal is bounded by
\[
O\!\left(\frac{H U}{p}\log\frac{U}{\delta}\right),
\]
where
\[
H=\left\lceil \frac{1}{1-\gamma}
\ln\frac{2R_{\max}}{\epsilon(1-\gamma)}\right\rceil,
\qquad
p=\frac{\epsilon(1-\gamma)}{2R_{\max}}.
\]
\end{theorem}

\begin{proof}
Condition on the concentration event of Lemma~\ref{lem:factored_accuracy_full}.
If the greedy policy in the optimistic model is not \(\epsilon\)-optimal,
then Lemma~\ref{lem:known_model_full} implies that, within the effective
horizon \(H\), the policy must reach an unknown component with probability
at least \(p\). Each such escape is charged to the first unknown component
reached.

Environment components can be charged at most \(|S||A|m_E\) times and
automaton components at most \(|S||Q|m_Q\) times. Hence the total number
of escapes is at most
\[
U = |S||A|m_E + |S||Q|m_Q.
\]
The standard R-Max escape-event argument then gives
\[
O\!\left(\frac{H U}{p}\log\frac{U}{\delta}\right)
\]
non-\(\epsilon\)-optimal interaction steps with probability at least
\(1-\delta\). Value iteration affects computation time but does not
consume environment interactions.
\end{proof}

\section{Bucket-QR-Max: Pseudocode and Conditional PAC-MDP Bound}
\label{app:bucket_pac}

We now give the full pseudocode for \bqrmax\ and state a PAC--MDP
guarantee in the continuous-state setting, under a standard
bounded-discretisation assumption. The proof proceeds by viewing
\bqrmax\ as \qrmax\ on the abstract MDP induced by the buckets, and
then controlling the approximation error due to discretisation.

\begin{algorithm}[htbp]
\scriptsize
\caption{\bqrmax}
\label{alg:bucket_qrmax}
\DontPrintSemicolon

\Input{Hash map $h:\mathbb{R}^d\to\mathcal B$, action set $A$,
automaton states $Q$, environment $E$, discount factor $\gamma$,
maximum reward $R_{\max}$, thresholds
$t_{E_T},t_{E_R},t_{Q_T},t_{Q_R}$, VI horizon $T$, number of steps
$N_{\mathrm{steps}}$.}

\Output{Optimistic action-value table $Q_B$.}

$V_{\max}\leftarrow R_{\max}/(1-\gamma)$\;
$\mathcal B_{\mathrm{disc}}\leftarrow\emptyset$\tcp*{Discovered buckets}

Initialize all counters lazily to zero:
$\tau_E(b,a,b')$, $n_{E_T}(b,a)$, $\rho_E(b,a)$, $n_{E_R}(b,a)$,
$\tau_Q(q,b',q')$, $n_{Q_T}(q,b')$, $\rho_Q(q,b')$, $n_{Q_R}(q,b')$\;

\BlankLine
\SetKwFunction{AddBucket}{AddBucket}
\SetKwProg{Fn}{Function}{:}{}
\Fn{\AddBucket{$b$}}{
  \If{$b\notin\mathcal B_{\mathrm{disc}}$}{
    $\mathcal B_{\mathrm{disc}}\leftarrow
    \mathcal B_{\mathrm{disc}}\cup\{b\}$\;
    $Q_B(b,q,a)\leftarrow V_{\max}$ for all $q\in Q$, $a\in A$\;
    Initialize all counters involving $b$ to zero\;
  }
}

\BlankLine
$s^{\mathrm{cont}},q\leftarrow E.\mathrm{reset}()$\;
$b\leftarrow h(s^{\mathrm{cont}})$\;
\AddBucket{$b$}\;

\For{$t\leftarrow 1$ \KwTo $N_{\mathrm{steps}}$}{
  $\bar a\leftarrow \arg\max_{a\in A} Q_B(b,q,a)$\;
  $s^{\mathrm{cont}\prime},q',r_E,r_A,\mathrm{done}
  \leftarrow E.\mathrm{step}(\bar a)$\;
  $b'\leftarrow h(s^{\mathrm{cont}\prime})$\;
  \AddBucket{$b'$}\;
  $\mathrm{doVI}\leftarrow \mathrm{False}$\;

  \BlankLine
  \tcp{Environment transition component}
  \If{$n_{E_T}(b,\bar a)<t_{E_T}$}{
    $\tau_E(b,\bar a,b')\leftarrow \tau_E(b,\bar a,b')+1$\;
    $n_{E_T}(b,\bar a)\leftarrow n_{E_T}(b,\bar a)+1$\;
    \If{$n_{E_T}(b,\bar a)=t_{E_T}$}{
      $\mathrm{doVI}\leftarrow \mathrm{True}$\;
    }
  }

  \tcp{Environment reward component}
  \If{$n_{E_R}(b,\bar a)<t_{E_R}$}{
    $\rho_E(b,\bar a)\leftarrow \rho_E(b,\bar a)+r_E$\;
    $n_{E_R}(b,\bar a)\leftarrow n_{E_R}(b,\bar a)+1$\;
    \If{$n_{E_R}(b,\bar a)=t_{E_R}$}{
      $\mathrm{doVI}\leftarrow \mathrm{True}$\;
    }
  }

  \BlankLine
  \tcp{Automaton transition component}
  \If{$n_{Q_T}(q,b')<t_{Q_T}$}{
    $\tau_Q(q,b',q')\leftarrow \tau_Q(q,b',q')+1$\;
    $n_{Q_T}(q,b')\leftarrow n_{Q_T}(q,b')+1$\;
    \If{$n_{Q_T}(q,b')=t_{Q_T}$}{
      $\mathrm{doVI}\leftarrow \mathrm{True}$\;
    }
  }

  \tcp{Non-Markovian reward component}
  \If{$n_{Q_R}(q,b')<t_{Q_R}$}{
    $\rho_Q(q,b')\leftarrow \rho_Q(q,b')+r_A$\;
    $n_{Q_R}(q,b')\leftarrow n_{Q_R}(q,b')+1$\;
    \If{$n_{Q_R}(q,b')=t_{Q_R}$}{
      $\mathrm{doVI}\leftarrow \mathrm{True}$\;
    }
  }

  \BlankLine
  \If{$\mathrm{done}$}{
    $Q_B(b',q',a')\leftarrow 0$ for all $a'\in A$\;
  }

  \If{$\mathrm{doVI}$}{
    $Q_B\leftarrow
    \mathrm{BucketVI}(Q_B,\tau_E,\rho_E,n_{E_T},n_{E_R},
    \tau_Q,\rho_Q,n_{Q_T},n_{Q_R},\gamma,T)$\;
  }

  \If{$\mathrm{done}$}{
    $s^{\mathrm{cont}},q\leftarrow E.\mathrm{reset}()$\;
    $b\leftarrow h(s^{\mathrm{cont}})$\;
    \AddBucket{$b$}\;
  }
  \Else{
    $s^{\mathrm{cont}},q,b\leftarrow s^{\mathrm{cont}\prime},q',b'$\;
  }
}

\Return{$Q_B$}\;
\end{algorithm}

The routine $\mathrm{BucketVI}$ performs optimistic value iteration on
the finite abstraction $\mathcal B_{\mathrm{disc}}\times Q$. The empirical
estimates used in the backup are
\[
\widehat P_E(b'\mid b,a)
=
\frac{\tau_E(b,a,b')}{n_{E_T}(b,a)},
\qquad
\widehat r_E(b,a)
=
\frac{\rho_E(b,a)}{n_{E_R}(b,a)},
\]
and
\[
\widehat P_Q(q'\mid q,b')
=
\frac{\tau_Q(q,b',q')}{n_{Q_T}(q,b')},
\qquad
\widehat r_Q(q,b')
=
\frac{\rho_Q(q,b')}{n_{Q_R}(q,b')}.
\]
A triple $(b,q,a)$ is treated as known only when
$n_{E_T}(b,a)\ge t_{E_T}$, $n_{E_R}(b,a)\ge t_{E_R}$,
and, for every observed successor bucket $b'$ with
$\tau_E(b,a,b')>0$,
$n_{Q_T}(q,b')\ge t_{Q_T}$ and $n_{Q_R}(q,b')\ge t_{Q_R}$.

For compactness, let
\[
V_k(b,q):=\max_{a\in A}Q_k(b,q,a),
\]
and define
\[
G_k(q,b')
:=
\widehat r_Q(q,b')
+
\gamma
\sum_{q'\in Q}
\widehat P_Q(q'\mid q,b')V_k(b',q').
\]
For known triples, $\mathrm{BucketVI}$ applies the factorised Bellman backup
\begin{equation}
\label{eq:bucket_vi_backup}
Q_{k+1}(b,q,a)
=
\widehat r_E(b,a)
+
\sum_{b'\in\mathcal B_{\mathrm{disc}}}
\widehat P_E(b'\mid b,a)G_k(q,b').
\end{equation}
Unknown triples retain their optimistic value
$V_{\max}=R_{\max}/(1-\gamma)$.

\subsection{Bounded-Discretisation Assumption}

We formalise the regularity condition under which the bucket abstraction
approximates the underlying continuous MDP closely enough.

\begin{assumption}[Bounded discretisation error]\label{ass:bounded_disc}
Let $M$ be the true MDP with continuous state space $S\subseteq\mathbb{R}^d$
and let $h:S\to\mathcal{B}$ be the SimHash-based discretisation used
by \textsc{Bucket-QRMAX}, inducing a partition
$\mathcal{B}=\{b_1,\dots,b_{|\mathcal{B}|}\}$ of $S$.
We assume that there exists a diameter parameter $\phi>0$ such that,
for every bucket $b\in\mathcal{B}$ and any pair of states
$s,s'\in S$ with $h(s)=h(s')=b$, we have
\[
\|s-s'\|\;\le\;\phi.
\]
Moreover, we assume that the reward and transition dynamics are
Lipschitz-continuous in the state, i.e., there exists $L>0$ such that
for all $s,s'\in S$ and all $a\in A$,
\[
|R_E(x,a,z)-R_E(y,a,z)| \le L\|x-y\| \qquad \forall x,y,z\in S
\]
and the transition kernels satisfy a corresponding Lipschitz condition
(e.g., in total-variation or Wasserstein distance) with constant $L$.
\end{assumption}

Assumption~\ref{ass:bounded_disc} captures the intuition that each
bucket groups together nearby states and that the MDP is sufficiently
smooth in the state.  In particular, if $\phi$ is small, then the error
incurred by replacing $s$ with its bucket $b=h(s)$ is also small.

\subsection{Conditional PAC-MDP Bound}

We now show that, under Assumption~\ref{ass:bounded_disc}, the
discretisation error can be absorbed into the standard PAC--MDP
analysis, yielding a conditional PAC guarantee for \textsc{Bucket-QRMAX}.

\begin{theorem}[Conditional PAC--MDP Bound for \bqrmax]
\label{thm:bucket_pac}
Assume~\ref{ass:bounded_disc} holds.
Let \(\varepsilon>0\) and \(\delta\in(0,1)\), and set
\[
\beta=\frac{\varepsilon(1-\gamma)}{8}.
\]
Choose thresholds
\[
m_{E_T},\,m_{E_R}
=
\left\lceil
\frac{8R_{\max}^2}{\beta^2}
\ln\frac{2|\mathcal B||A|}{\delta}
\right\rceil,
\qquad
\]
\[
m_{Q_T},\,m_{Q_R}
=
\left\lceil
\frac{8R_{\max}^2}{\beta^2}
\ln\frac{2|\mathcal B||Q|}{\delta}
\right\rceil.
\]
Let
\[
U_{\mathcal B}
=
|\mathcal B||A|(m_{E_T}+m_{E_R})
+
|\mathcal B||Q|(m_{Q_T}+m_{Q_R}).
\]
Moreover, let
\[
H=
\left\lceil
\frac{1}{1-\gamma}
\ln\frac{2R_{\max}}{\varepsilon(1-\gamma)}
\right\rceil,
\qquad
p=
\frac{\varepsilon(1-\gamma)}{2R_{\max}}.
\]
Suppose that the bucket diameter satisfies
\(\phi\le c\beta/L\) for a suitable universal constant \(c>0\).
Then, with probability at least \(1-\delta\), \bqrmax\ executes at most
\[
O\!\left(
\frac{H U_{\mathcal B}}{p}
\log\frac{U_{\mathcal B}}{\delta}
\right)
\]
interactions in which its policy is not \(\varepsilon\)-optimal in the
true continuous MDP \(M\), up to the discretisation error controlled by
Assumption~\ref{ass:bounded_disc}.
\end{theorem}

\smallskip
\noindent\textbf{1. Discretisation Error and Abstract MDP.}
Let \(M\) be the true continuous-state MDP and let \(\bar M\) be the
abstract MDP whose states are the bucket--automaton pairs
\((b,q)\in\mathcal B\times Q\), with transitions and rewards given by
the bucket-averaged quantities. Under Assumption~\ref{ass:bounded_disc},
any two states \(s,s'\) mapped to the same bucket \(b\) satisfy
\(\|s-s'\|\le \phi\). Lipschitz continuity then implies that the reward
and transition probabilities of \(M\) at \(s\) and \(s'\) differ by at
most \(O(L\phi)\).

By a standard simulation-lemma argument, there exists a universal
constant \(c>0\) such that, if \(\phi\le c\beta/L\), then \(\bar M\)
is a \(\beta\)-approximation of \(M\): for any policy \(\pi\) and any
state \(s\) with bucket \(b=h(s)\),
\[
\bigl|V^{\pi}_{M}(s) - V^{\pi}_{\bar M}(b,q)\bigr|
\le \frac{\beta}{2}.
\]

\smallskip
\noindent\textbf{2. Estimation Error in the Bucket MDP.}
Within the abstract MDP \(\bar M\) on \((b,q)\), \bqrmax\ behaves as
\qrmax\ on a discrete MDP with state space
\(\mathcal B\times Q\). The thresholds
\(m_{E_T},m_{E_R},m_{Q_T},m_{Q_R}\) are chosen so that, once a triple
\((b,a,q)\) is marked known, the empirical estimates are
\(\beta\)-accurate with high probability. More precisely, a concentration
argument and a union bound over all \((b,a)\) and \((q,b')\) yield
\[
\|\widehat P-P\|_1 \le \frac{\beta}{R_{\max}},
\qquad
|\widehat R-R|\le \beta
\]
for all known components, with failure probability at most \(\delta\).

Conditioned on this event, applying
Theorem~\ref{thm:qrmax_pac_full} to the abstract bucket MDP gives the
same optimistic escape bound with
\[
U_{\mathcal B}
=
|\mathcal B||A|(m_{E_T}+m_{E_R})
+
|\mathcal B||Q|(m_{Q_T}+m_{Q_R}).
\]
Thus the greedy policy is \(\beta\)-optimal in \(\bar M\), except for
\[
O\!\left(
\frac{H U_{\mathcal B}}{p}
\log\frac{U_{\mathcal B}}{\delta}
\right)
\]
interaction steps.

\smallskip
\noindent\textbf{3. Combining the Two Sources of Error.}
For any state \(s\) with bucket \(b=h(s)\), decompose the value gap as
\[
\begin{aligned}
\bigl|V^{\pi_t}_{M}(s)-V^{*}_{M}(s)\bigr|
\le{}&
\bigl|V^{\pi_t}_{M}(s)-V^{\pi_t}_{\bar M}(b,q)\bigr| \\
&+
\bigl|V^{\pi_t}_{\bar M}(b,q)-V^{*}_{\bar M}(b,q)\bigr| \\
&+
\bigl|V^{*}_{\bar M}(b,q)-V^{*}_{M}(s)\bigr|.
\end{aligned}
\]
The two discretisation terms are controlled by Step~1, while the
estimation term is controlled by Step~2 on all non-escape steps. Choosing
the constants in the thresholds so that these three contributions sum to
at most \(\varepsilon\) yields the stated guarantee. Therefore the same
optimistic escape-event argument gives
\[
O\!\left(
\frac{H U_{\mathcal B}}{p}
\log\frac{U_{\mathcal B}}{\delta}
\right)
\]
non-\(\varepsilon\)-optimal interactions.

\begin{remark}[On the role of SimHash]\label{rem:lsh}
Assumption~\ref{ass:bounded_disc} is a property of the partition
$\mathcal{B}$ induced by the discretisation.  In our implementation
the partition is generated by a SimHash-based locality-sensitive hash.
SimHash does not deterministically guarantee a given diameter~$\phi$,
but increasing the number of hash functions and projections reduces
the probability that distant states collide in the same bucket.
Thus,  \bqrmax can be viewed as satisfying a PAC-type
property with probability at least $1-(\delta+\delta_{\mathrm{LSH}})$,
where $\delta_{\mathrm{LSH}}$ denotes the probability that the induced
partition violates Assumption~\ref{ass:bounded_disc}.  In the main
text we focus on the PAC bound for the induced bucket MDP, and leave
a detailed analysis of $\delta_{\mathrm{LSH}}$ to future work.
\end{remark}

\section{Additional Experimental Results} \label{app:experiments}
The experiments were performed on a Linux x86-64 workstation with a single-socket AMD Ryzen 9 7950X processor (Zen~4, 16~cores/32~threads, boost clock $\sim$5.9\,GHz). The system is equipped with 128~GB of DDR5 RAM ($\approx$124~GiB available at idle) and an 8~GB swap partition. Hardware acceleration is provided by two NVIDIA GeForce RTX~4090 (AD102) GPUs. This configuration underlies all the results presented in our work.

\subsection{Office gridworld Benchmark}
The benchmark used for experimental comparison in this paper takes inspiration from and extends the Office gridworld environment \cite{pmlr-v80-icarte18a}
already used to evaluate solutions for NMRDPs.
We extend the original environment with the following features: (1) stochastic outcomes of actions: the nominal effect of an action is achieved with probability $h$, and two adjacent actions with probability $(1-h)/2$; (2) three different sizes of the map $9 \times 12$, $12 \times 12$, and $15 \times 15$ (the latter is shown in Figure \ref{fig:map3}); additional locations to be used as goals; the definition of 5 different tasks of increasing complexity. The benchmark can be used to test algorithms in 15 environments (5 tasks in 3 maps) with increasing complexity.

In the first task, the agent must deliver coffee to the office. In the second task, the goal is to deliver mail to the office. The third task requires the agent to deliver both coffee and mail to the office in any order.
The fourth task requires the agent to visit locations A, B, C, and D in the correct sequence. Finally, in the fifth task, the agent must first visit locations A, B, C, and D, then collect both mail and coffee in any order, and ultimately reach the office (the DFA for the latter task is illustrated in Figure \ref{fig:RM_exp5}). 

The agent observation space is given by its own position extended with the state of the DFA, i.e., $(s,q) \in S \times Q$. The action space is formed by the four cardinal actions. 
The reward signal is 1 for completing the task (i.e., for goal states in ${\cal A}$) and 0 otherwise. Walls make the agent remain in the same cell, while hitting decorations terminates the episode with reward \(-100\). 

%

\subsection{Environment \texttt{map0}.}
We build on the grid--world \texttt{map0} introduced by \cite{tiapkin2022optimistic}, a \(10\times10\) grid.
For UCBVI, PSRL, and OPSRL the planning horizon is fixed at \(H = 50\); the two Bayesian methods use eight Thompson samples per episode, following the authors’ default.

\medskip\noindent\textbf{Non-Markovian extension.}
The original one-shot objective is replaced by a four-stage mission that forces the agent to remember past events.  
The episode begins in the top-left cell; the agent must first travel to the bottom-right cell to pick up a letter, then return to the start cell to collect a coffee, and finally deliver both items to the office located in the bottom-left corner.  
A reward of \(1\) is granted only when this entire sequence is completed; every other transition yields zero reward.

\begin{figure}[t]
	\centering
	\includegraphics[width=0.9
	\linewidth]{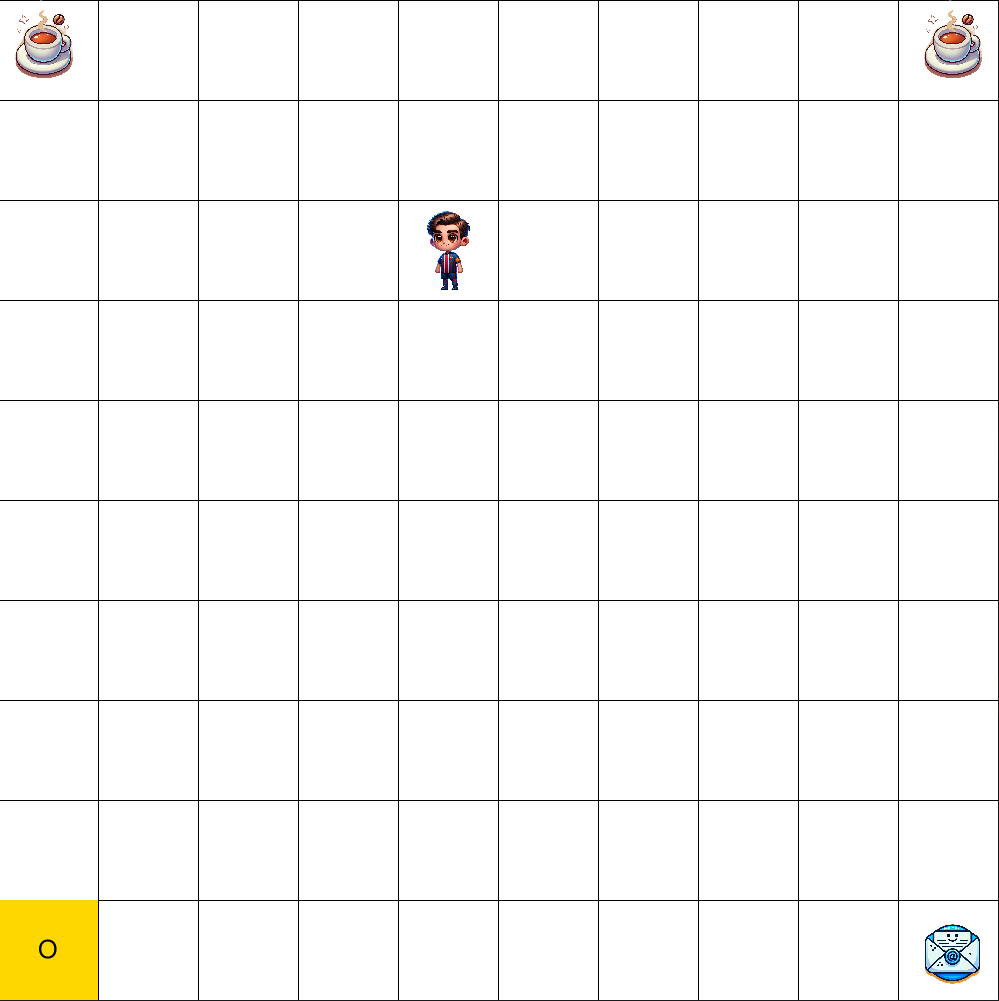}
	\caption{Map0 - task 0 (10x10)}
	\label{fig:map0}
\end{figure}


\begin{figure}[t]
	\centering
	\includegraphics[width=0.9
	\linewidth]{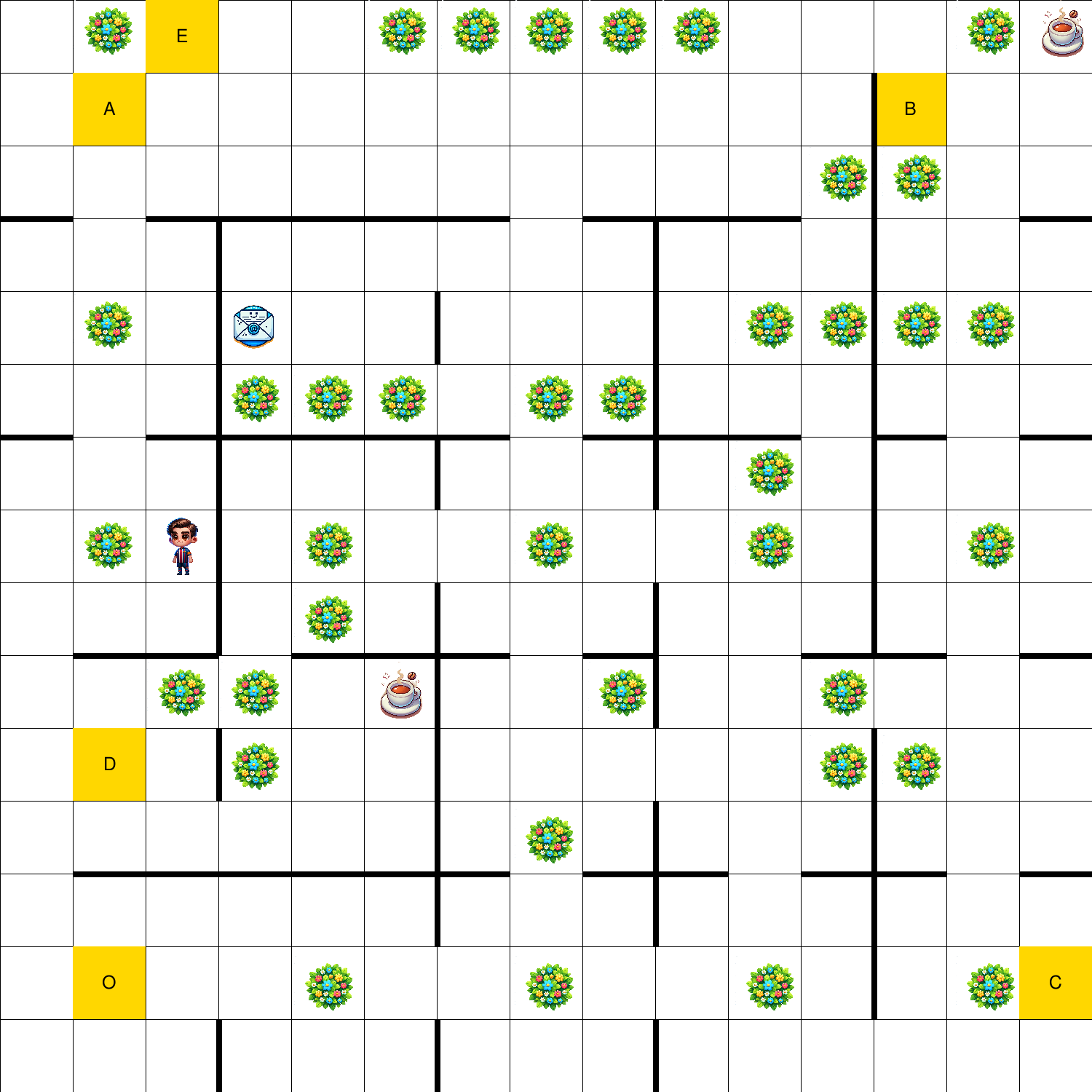}
	\caption{Map4 - task 6 (15x15) - Office World}
	\label{fig:map4}
\end{figure}


\subsection{Comparison algorithms}
\label{sec:baselines}

Our evaluation includes both optimistic planners and posterior–sampling methods defined over the joint space \(S \times Q\).
We adopt the three variants of \textsc{UCBVI} proposed by \cite{azar17a}: \textsc{UCBVI-sB} uses the simplified Bernstein bonus, \textsc{UCBVI-H} the Hoeffding bonus, and \textsc{UCBVI-B} the full Bernstein bonus; in each case planning is performed with tabular value iteration and the same optimism constant \(c\).
To represent the Bayesian family we run \textsc{PSRL} \cite{osband2013}, which samples one MDP from a Dirichlet–Gaussian posterior at the start of every episode, and \textsc{OPSRL} \cite{tiapkin2022optimistic}, which augments each posterior sample with an explicit exploration bonus.
As a classical optimistic baseline we include \rmax \cite{Brafman03}.
For the non-Markovian setting we also test the model-free algorithm \textsc{QRM} \cite{pmlr-v80-icarte18a}, which learns a Q-function directly over \(S \times Q\).

The algorithm introduced in this paper, \qrmax, exploits the factorisation of the environment and automaton dynamics, while \rmaxrm and \qrmaxrm extend \rmax and \qrmax, respectively, by propagating knowledge to unseen states as in \textsc{QRM}. 

The algorithms QL, \rmax, \qrmax, \rmaxrm and \qrmaxrm use the discount factor \(\gamma = 0.9\).
All the variants of UCBVI and PSRL/OPSRL use the discount factor \(\gamma = 1\).
For \rmax, \qrmax, \rmaxrm and \qrmaxrm, a transition is considered \emph{known} after 39 visits.
\textsc{QRM} and \qlearn\ employ an exploration rate \(\epsilon = 0.1\) and a learning rate \(\alpha = 0.1\).
Following the original papers, \textsc{PSRL} and \textsc{OPSRL} draw eight Thompson samples per episode, and the planning horizon is set to \(H=50\) for \texttt{map0} and \(H=150\) for \texttt{map1}.


\paragraph{Horizon and reward scheme.}
All finite–horizon exploration methods—\textsc{UCBVI} (every variant), \textsc{PSRL} and \textsc{OPSRL}—require a hand-chosen horizon \(H\).  To give these baselines favourable yet realistic conditions we set \(H\) to the \emph{average optimal-path length} computed by Value Iteration, adding a small safety margin for stochastic slips: \(H=50\) in \textsc{MAP0\_EXP0}, \(H=150\) in \textsc{MAP1\_EXP5} and \textsc{MAP4\_EXP6}, and \(H=350\) in \textsc{MAP2\_EXP5} (Table~\ref{tab:steps-by-map-exp-algo-bestpath-sr}).  Choosing \(H\) too low truncates the policy, while over-estimating it inflates exploration bonuses and slows learning at a rate of \(\widetilde{\mathcal O}(H^{2\text{--}4})\); the calibration therefore embodies non-trivial prior knowledge that horizon-free algorithms do not need.

A uniform, deliberately sparse reward is used throughout: the agent receives \(+1\) only when the Reward-Machine task is fully completed and \(-100\) upon hitting a decoration; all other transitions yield zero.  The strong terminal penalty restarts the episode immediately, ensuring that no method—irrespective of its dependence on \(H\)—wastes steps in an absorbing state.  This design isolates exploration efficiency as the sole differentiator between algorithms and prevents any hidden advantage from reward shaping.

\begin{table}[ht]
\centering
\scriptsize
\setlength{\tabcolsep}{6pt}
\begin{tabular}{@{}lccccc@{}}
\toprule
\textbf{Algorithm} & \textbf{Type} &
\multicolumn{1}{p{2cm}}{\centering \textbf{Dynamics\\factorized}} &
\multicolumn{1}{p{1.4cm}}{\centering \textbf{DFA\\known}} \\ \midrule
\textsc{Q‑Learning}  & model‑free  & \xmark & \xmark \\
\textsc{R‑Max}       & model‑based & \xmark & \xmark \\
\textsc{QR‑Max}      & model‑based & \checkmark & \xmark \\
\textsc{QRM}         & model‑free  & \xmark & \checkmark \\
\textsc{R‑MaxRM}     & model‑based & \xmark & \checkmark \\
\textsc{QR‑MaxRM}    & model‑based & \checkmark & \checkmark \\ \bottomrule
\end{tabular}
\caption{Compared algorithms.}
\label{tab:algorithms}
\end{table}


\begin{figure*}[ht]
	\centering
	\includegraphics[width=0.9
	\linewidth]{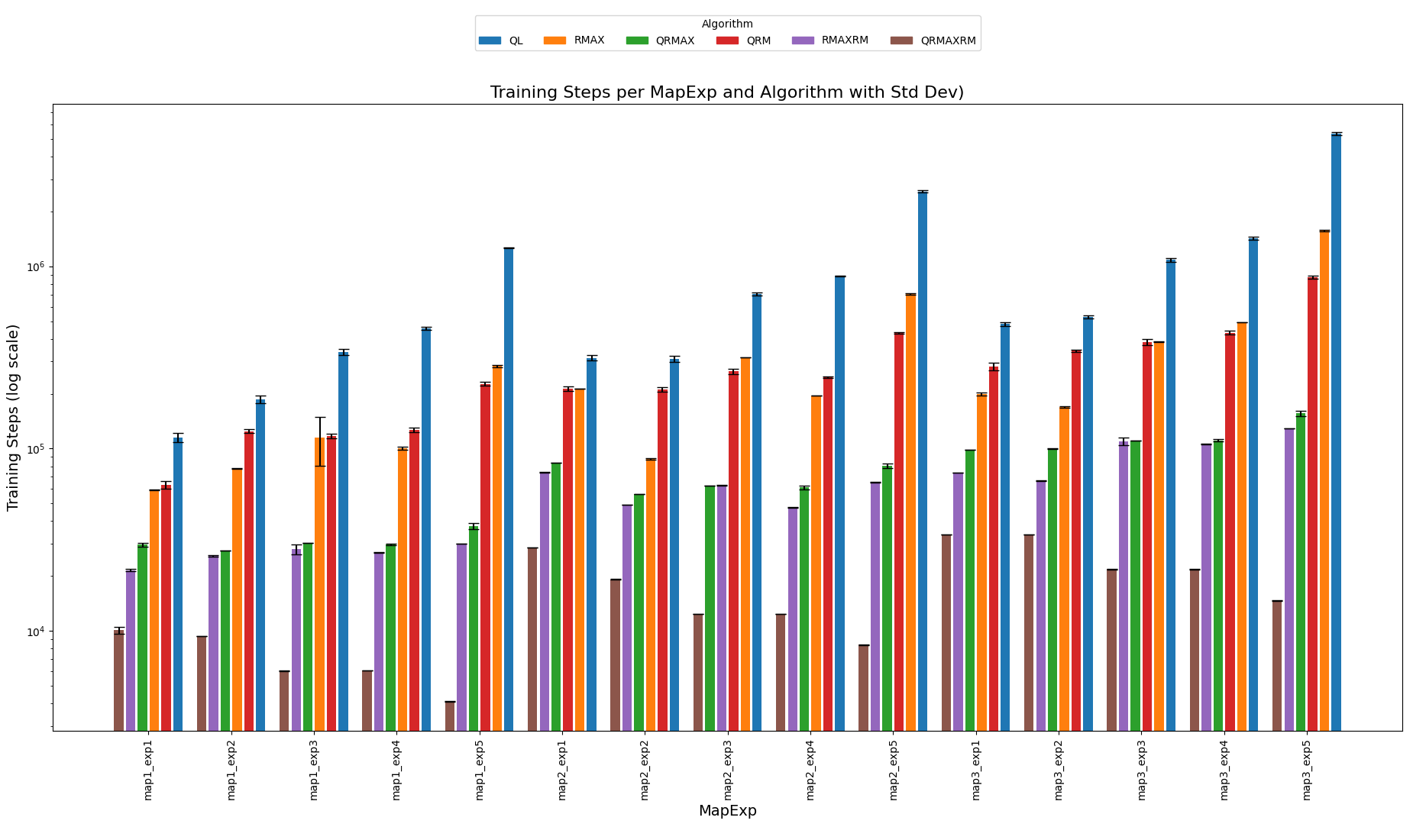}
	\caption{Scalability with respect to task complexity. Training steps needed to achieve the goal in 15 different tasks.}
	\label{fig:all_exp}
\end{figure*}
\subsection{Overall results on 15 configurations}

Figure~\ref{fig:all_exp} and Table~\ref{tab:steps-by-map-exp-algo-bestpath-sr} reports, for each \emph{map--task} instance, the \emph{training steps} (log scale, mean~$\pm$~std.\ dev.) required by each algorithm to learn a policy that is optimal in the corresponding deterministic MDP.  Results are averaged over thirty runs with 30 different random seeds, and at regular intervals the current policy is evaluated over 100 deterministic episodes; training stops as soon as its mean return is statistically indistinguishable from the optimal one ($p>0.1$, two‑sided Welch’s $t$‑test).
\begin{compactitem}
\item \textbf{Model‑based vs.\ model‑free:} both \rmax and \qrmax consistently require far fewer steps than their model‑free counterparts (\qlearn, \textsc{QRM}), regardless of DFA access.  
\item \textbf{Benefit of knowing the DFA:} supplying the reward‑machine structure (suffix \textit{RM}) yields an additional $10$–$50\times$ reduction in sample complexity (\rmaxrm, \qrmaxrm \ vs.\ \rmax, \qrmax).  
\item \textbf{Impact of the factorisation:} even with DFA knowledge, \qrmaxrm is about $1.5$–$2\times$ faster than \rmaxrm, and without DFA \qrmax is roughly an order of magnitude faster than \rmax, showing that sharing the learned environment model across automaton states remains advantageous.  \item \textbf{Overall winner:} \qrmaxrm achieves the lowest step count on every instance; even without the DFA, \qrmax outperforms \textsc{QRM}, confirming that the factorisation compensates for the lack of reward‑machine knowledge.  
\end{compactitem}

\begin{figure}[t]
	\centering
	\includegraphics[width=0.8
	\linewidth]{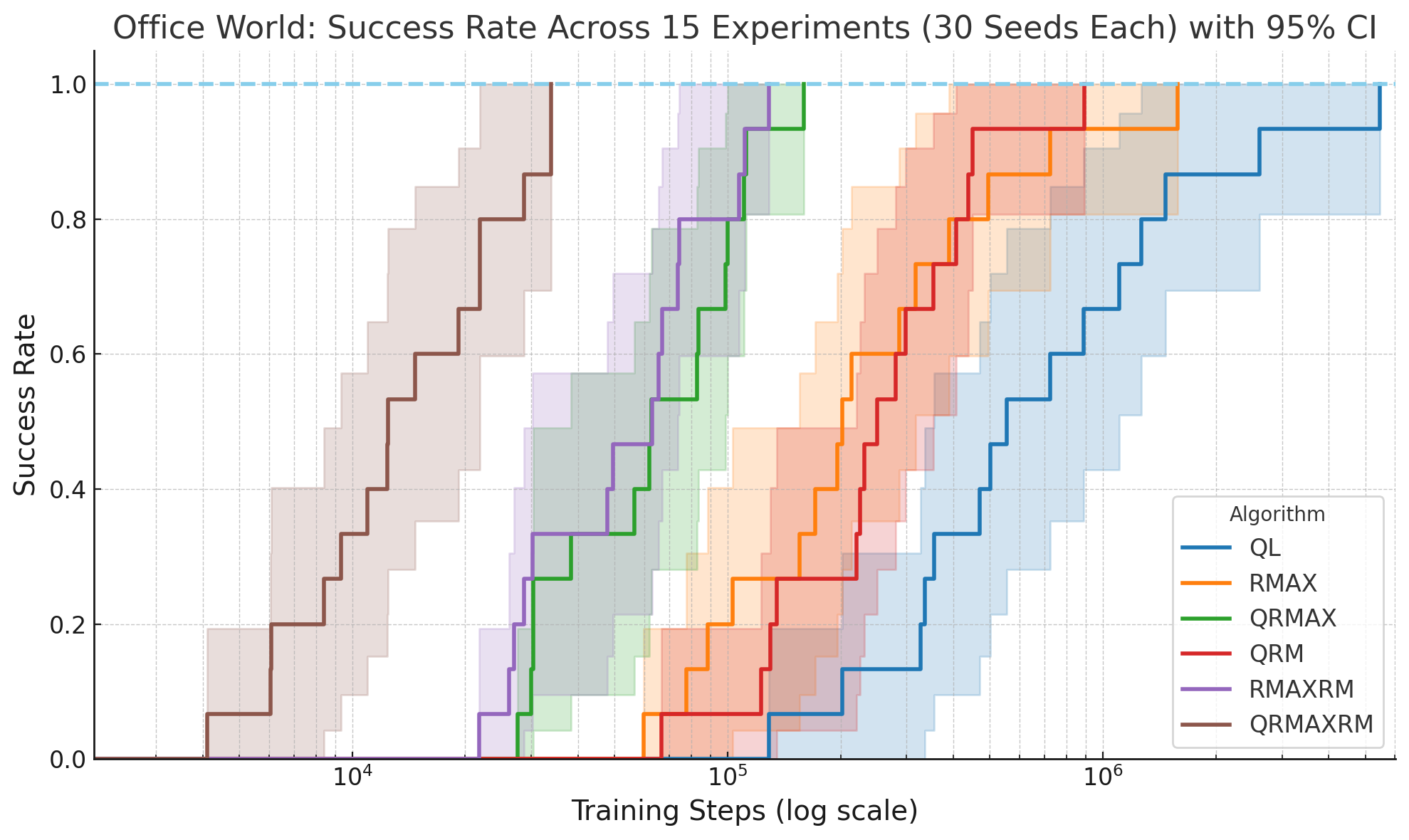}
	\caption{Office World - 15 tasks with 95 \% confidence intervals.}
	\label{fig:all_exp_success}
\end{figure}

In summary, factorising environment and reward dynamics delivers substantial and orthogonal gains over both classic \rmax and reward‑machine baselines, yielding the most sample‑efficient algorithm across all Office‑World configurations (Figure~\ref{fig:all_exp_success}).


\subsection{Detailed results on 5 configurations}

\begin{figure}[t]
	\centering
	\includegraphics[width=0.99
	\linewidth]{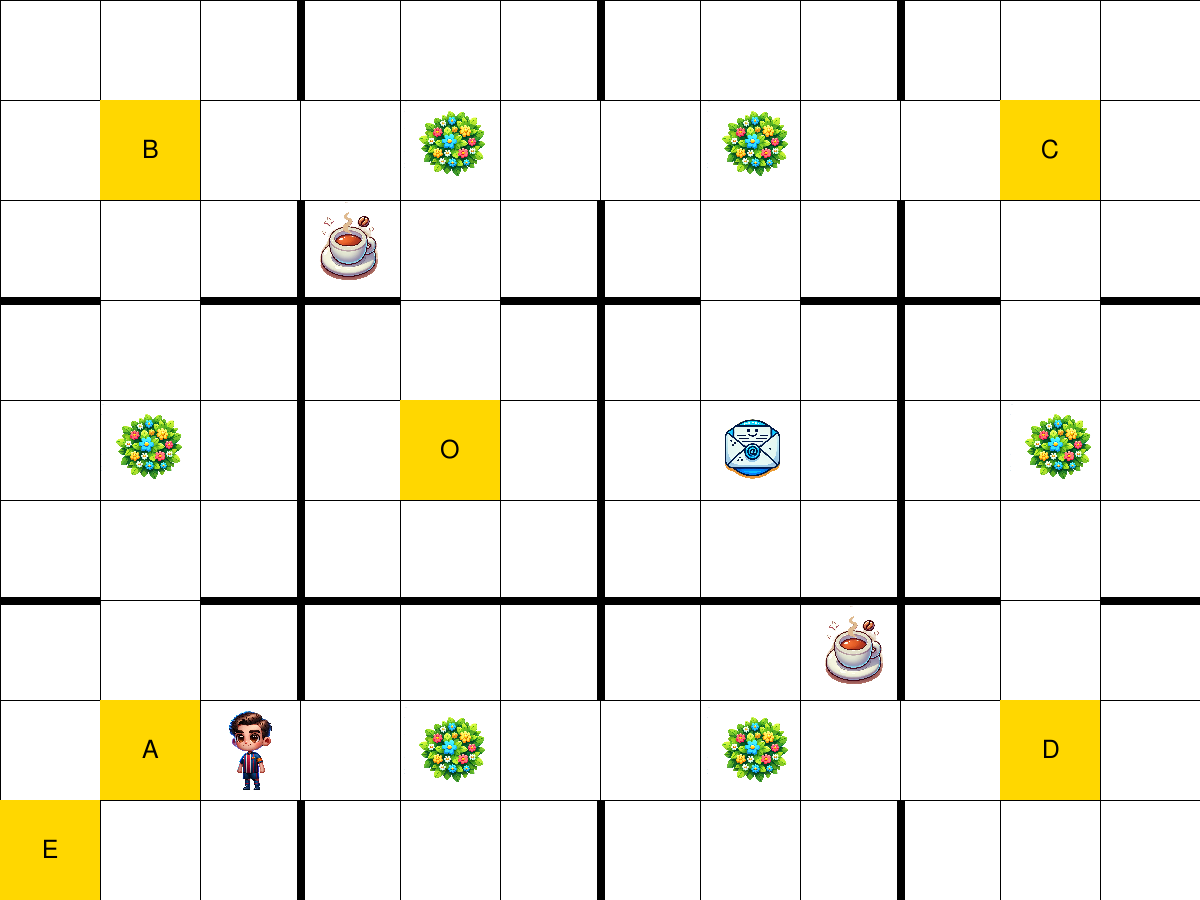}
	\caption{Map1 (12x9) - Office World}
	\label{fig:map1}
\end{figure}

\begin{figure}[t]
	\centering
	\includegraphics[width=0.99
	\linewidth]{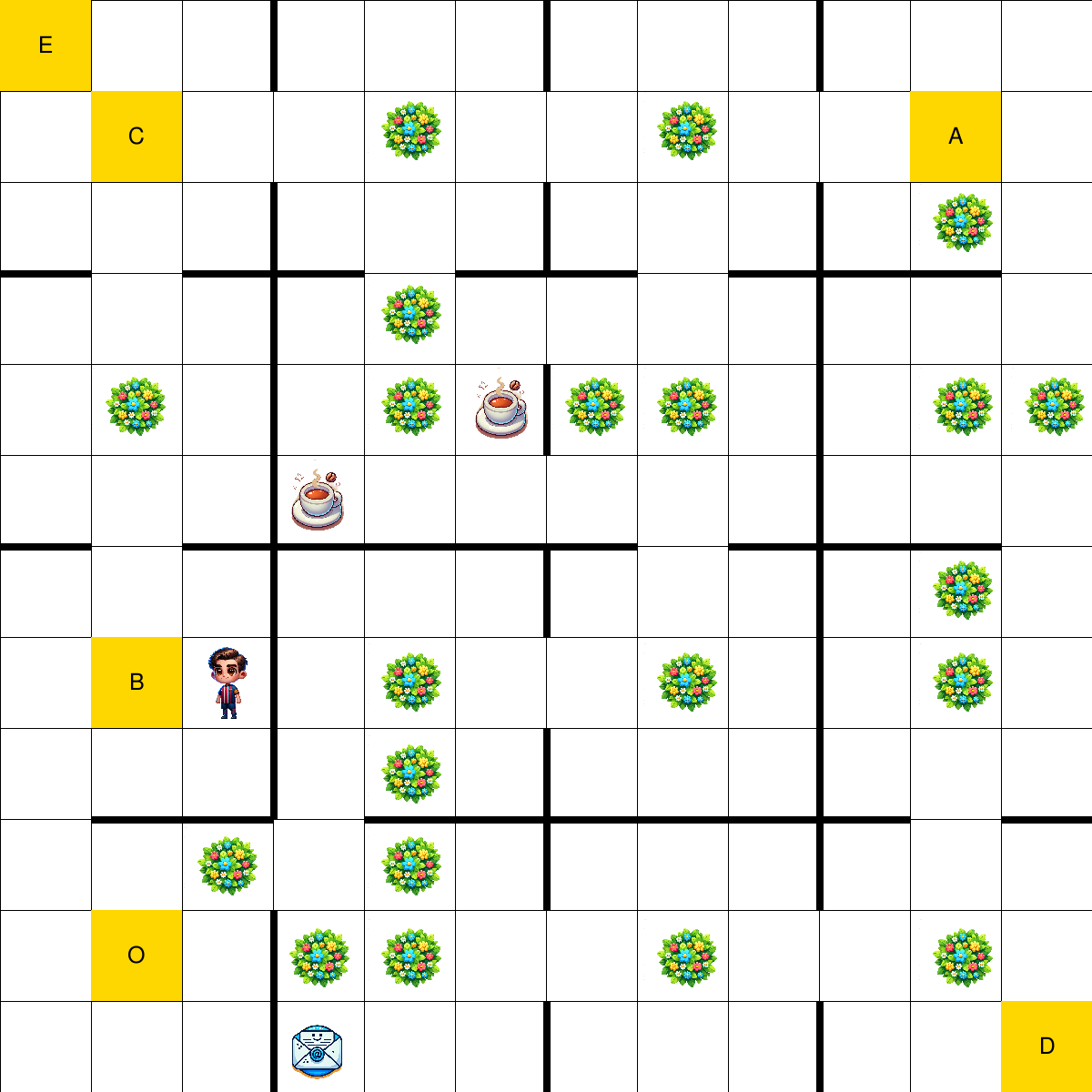}
	\caption{Map2 (12x12) - Office World}
	\label{fig:map2}
\end{figure}

\begin{figure}[t]
	\centering
	\includegraphics[width=0.99
	\linewidth]{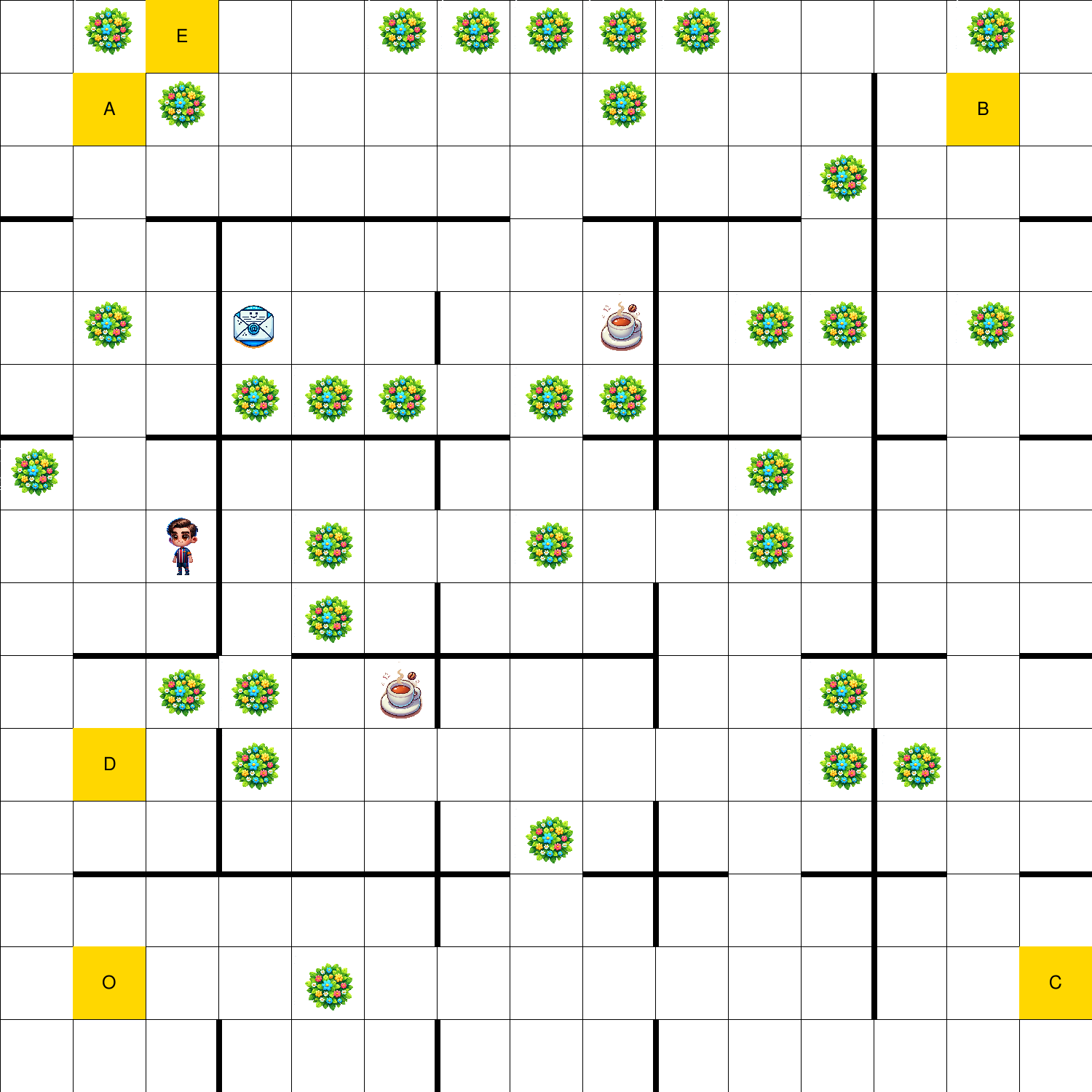}
	\caption{Map3 (15x15) - Office World}
	\label{fig:map3}
\end{figure}

In this section, we analyze in detail the behavior of the different algorithms on 5 relevant configurations of the Office gridworld environment.
In particular, we used three distinct maps and five different Reward Machines, each progressively more complex. Map 1 in figure \ref{fig:map1} is a 12x9 grid, the same used by Icarte et al. (2022). Map 2 in figure \ref{fig:map2} is a 12x12 grid, featuring more rooms and decorations. Map 3 in figure \ref{fig:map3} is the most complex, a 15x15 maze-like grid with goals arranged to challenge the agent.
All environments share the same discrete action set—\emph{up}, \emph{down}, \emph{left}, \emph{right}—and a stochastic transition noise of~\(0.2\).  
In \texttt{map0\_exp0} (a \(10\times10\) empty grid), each action has a \(20\%\) chance of being replaced by any other action.  
In \texttt{map1\_exp5} and \texttt{map4\_exp6} the noise deflects the agent towards a perpendicular action, mirroring the \emph{slip} mechanism of \textsc{Frozen Lake}.

The agent's objective is to reach the goals in a specific order without stepping on any decorations, which would terminate the episode.

\begin{figure}[t]
	\centering
	\includegraphics[width=0.99
	\linewidth]{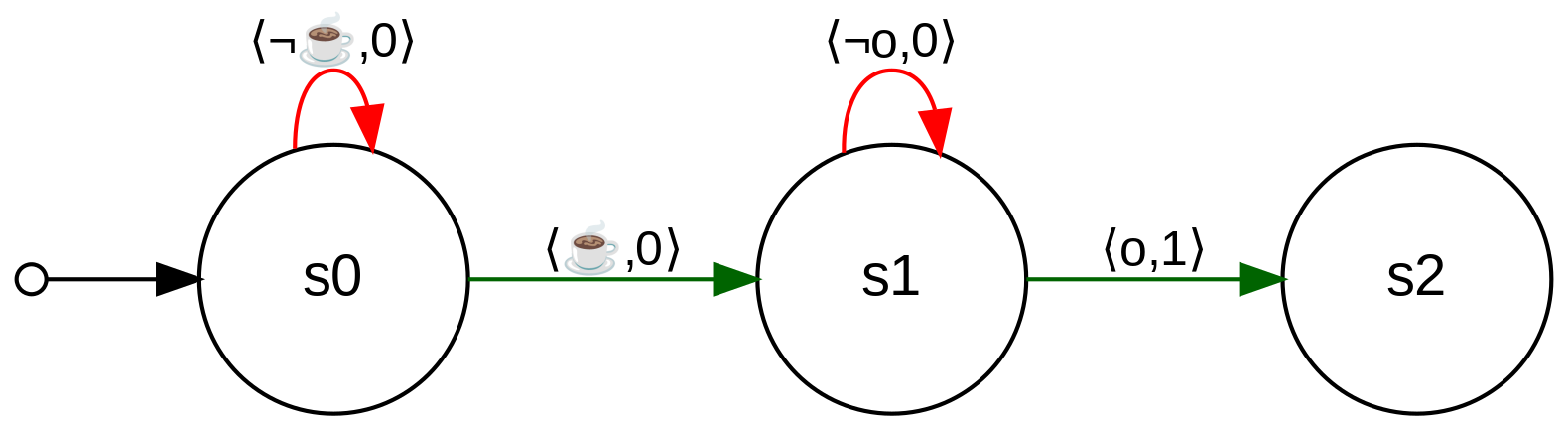}
	\caption{Reward Machine of Experiment 1 - Office World}
	\label{fig:RM_exp1}
\end{figure}

\begin{figure}[t]
	\centering
	\includegraphics[width=0.99
	\linewidth]{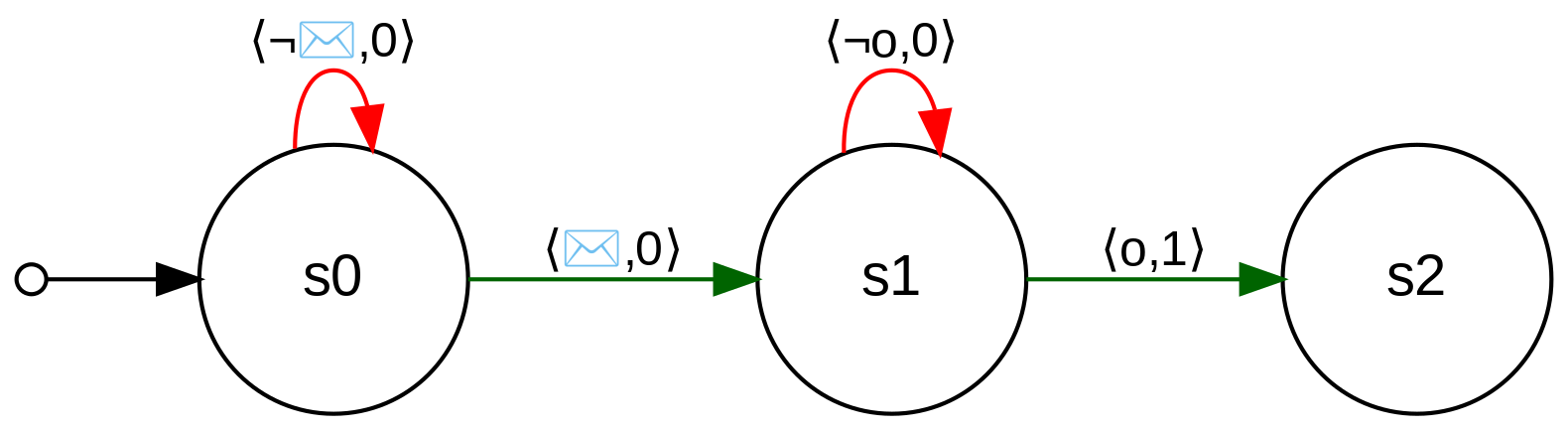}
	\caption{Reward Machine of Experiment 2 - Office World}
	\label{fig:RM_exp2}
\end{figure}

\begin{figure}[t]
	\centering
	\includegraphics[width=0.99
	\linewidth]{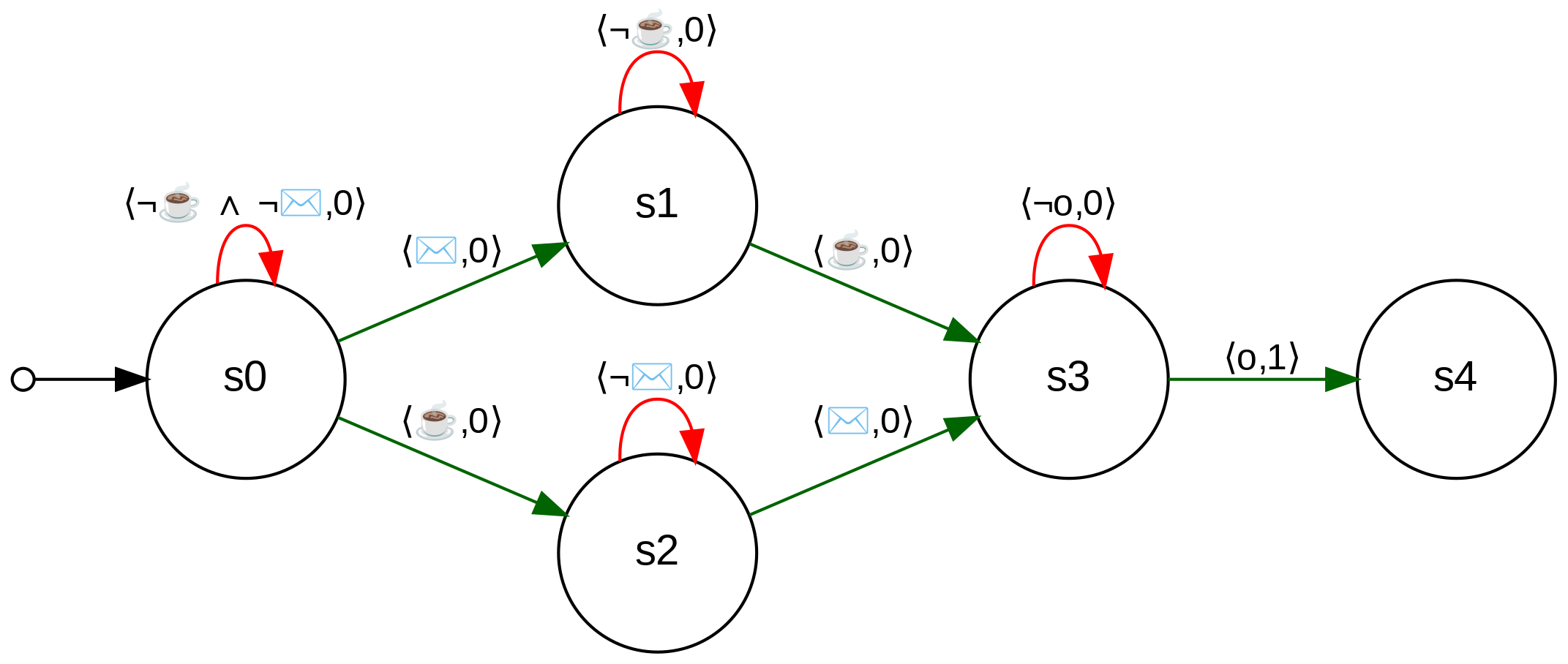}
	\caption{Reward Machine of Experiment 3 - Office World}
	\label{fig:RM_exp3}
\end{figure}

\begin{figure}[t]
	\centering
	\includegraphics[width=0.99
	\linewidth]{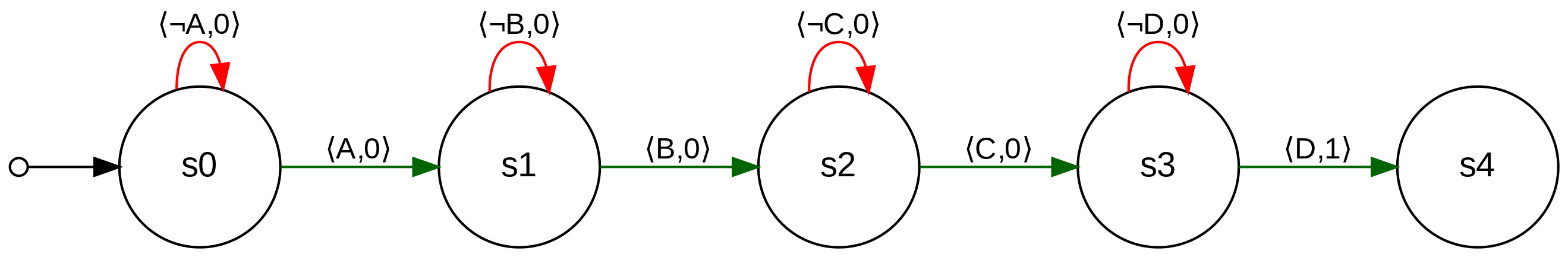}
	\caption{Reward Machine of Experiment 4 - Office World}
	\label{fig:RM_exp4}
\end{figure}

\begin{figure}[t]
	\centering
	\includegraphics[width=0.99
	\linewidth]{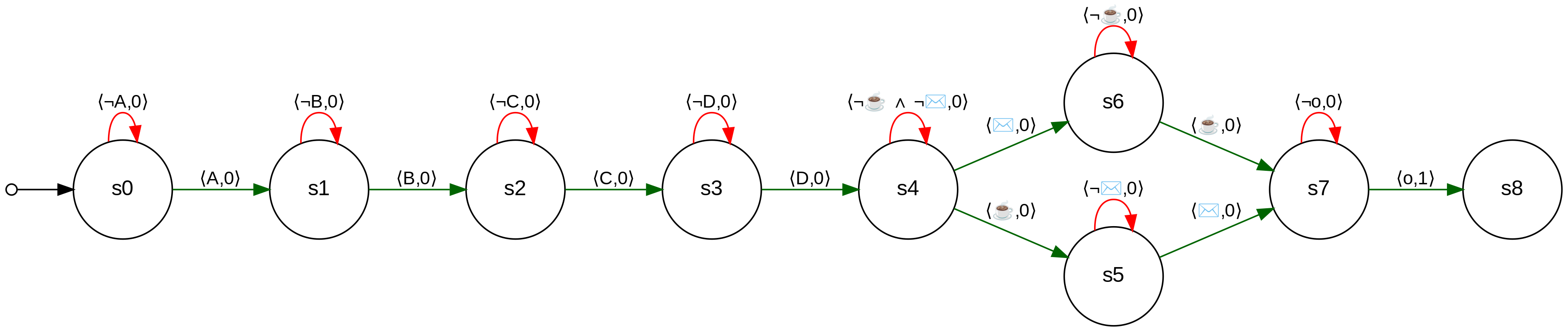}
	\caption{Reward Machine of Experiment 5 - Office World}
	\label{fig:RM_exp5}
\end{figure}

\begin{figure}[t]
	\centering
	\includegraphics[width=0.99
	\linewidth]{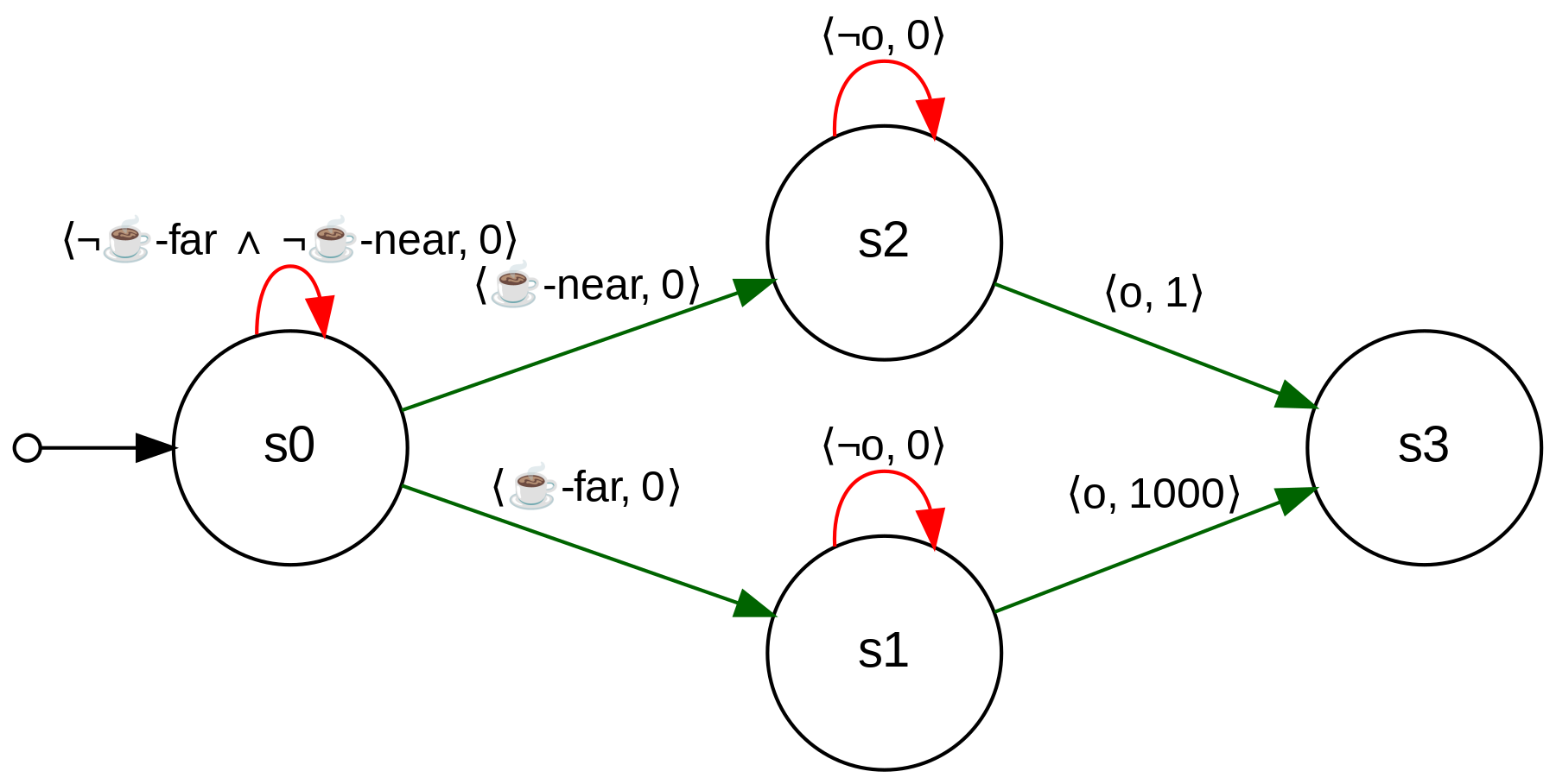}
	\caption{Reward Machine of Experiment 6 - Office World}
	\label{fig:RM_exp6}
\end{figure}

The five Reward Machines correspond to five experiments with increasing difficulty:

\begin{compactitem}
    \item \textbf{Task 1}: Deliver coffee to the office without breaking any decoration (figure \ref{fig:RM_exp1}).
    \item \textbf{Task 2}: Deliver mail to the office without breaking any decoration (figure \ref{fig:RM_exp2}).
    \item \textbf{Task 3}: Deliver both coffee and mail to the office, in any order, without breaking any decoration (figure \ref{fig:RM_exp3}).
    \item \textbf{Task 4}: Patrol locations A, B, C, and D in the correct order, without breaking any decoration (figure \ref{fig:RM_exp4}).
    \item \textbf{Task 5}: Patrol locations A, B, C, D, then collect email and coffee in any order, and finally go to the office (figure \ref{fig:RM_exp5}).
\end{compactitem}
In addition, we consider a sixth task (Task 6), defined only on the largest map (Map4), where the environment is a maze-like space featuring two coffee machines: a \emph{good} one (reward \(1000\)) and a \emph{regular} one (reward \(1\)); see Figure~\ref{fig:RM_exp6}.

Table~\ref{tab:steps-stats-all-maps} reports the mean and standard deviation of the number of training steps required by each algorithm to reach 100\% success, averaged over the five experiments (exp1–exp5) on each map.  As the table shows, the model-free baseline Q-Learning incurs extremely high sample complexity (on the order of $5\times10^{5}$ to $1.8\times10^{6}$ steps) with very large variance, particularly on the larger maps.  Introducing a model (\rmax) reduces the mean by a factor of roughly 3–6, while \qrmax further cuts it by another factor of $\approx4$, achieving means of $3.1\times10^{4}$, $6.9\times10^{4}$, and $1.16\times10^{5}$ on Maps 1–3 respectively.  Equipping the solvers with the reward-machine structure (the “RM” variants) yields additional speedups: \rmaxrm requires on average only $2.7\times10^{4}$, $6.0\times10^{4}$, and $9.8\times10^{4}$ steps, while \qrmaxrm plunges to $7.3\times10^{3}$, $1.6\times10^{4}$, and $2.5\times10^{4}$ steps on Maps 1–3, with relatively low standard deviations (see Table~\ref{tab:steps-stats-all-maps}).

Table~\ref{tab:steps-by-map-exp-algo-bestpath-sr} presents the full breakdown by map and experiment, together with the optimal path length and the success rate of value iteration (VI).  For example, on Map1 exp1 the optimal makespan is 21.08 steps (VI success 66.76\%), and QR-MAXRM converges in 10,964 steps—an order of magnitude fewer than R-MAXRM (21,777) and two orders fewer than Q-Learning (128,973).  As task complexity grows (e.g.\ exp5 on each map), the optimal path length increases substantially (up to 156.46 steps on Map3 exp5) and the VI success rate falls below 10\%, reflecting harder coordination challenges.  Nevertheless, QR-MAXRM still completes training in $1.47\times10^{4}$ steps on Map3 exp5, compared to $1.29\times10^{5}$ for R-MAXRM and $5.47\times10^{6}$ for Q-Learning.

Taken together, these results demonstrate that factorising environment and reward dynamics, combined with reward-machine knowledge, drives dramatic reductions in sample complexity across all environments and difficulty levels, while the per-experiment metrics in Table~\ref{tab:steps-by-map-exp-algo-bestpath-sr} highlight the scalability and robustness of QR-MAXRM even as the optimal plans lengthen and the domain becomes more challenging.


\begin{table}[ht]
\centering
\caption{Mean $\pm$ Std.\ dev.\ of training steps to reach 100\% success, aggregated over five experiments (exp1–exp5) per map}
\label{tab:steps-stats-all-maps}
\resizebox{\columnwidth}{!}{%
  \begin{tabular}{lccc}
  \toprule
  \textbf{Algorithm} & \textbf{Map 1} & \textbf{Map 2} & \textbf{Map 3} \\
  \midrule
  Q-Learning &  484,528 $\pm$ 457,245 &  977,552 $\pm$ 946,899 & 1,819,687 $\pm$ 2,080,127 \\
  R-MAX      &  136,622 $\pm$  91,486 &  307,805 $\pm$ 246,020 &   567,680 $\pm$   582,237 \\
  QR-MAX     &   31,316 $\pm$   4,073 &   69,494 $\pm$  12,854 &   116,237 $\pm$    25,001 \\
  QRM        &  137,263 $\pm$  59,580 &  282,943 $\pm$  90,035 &   479,702 $\pm$   236,923 \\
  R-MAXRM    &   26,776 $\pm$   3,199 &   60,058 $\pm$  11,165 &    97,639 $\pm$    26,300 \\
  QR-MAXRM   &    7,312 $\pm$   2,774 &   16,240 $\pm$   7,990 &    25,222 $\pm$     8,385 \\
  \bottomrule
  \end{tabular}%
}
\end{table}

\begin{table*}[ht]
\centering
\scriptsize
\caption{Training steps to reach the optimal policy compared with Value Iteration (VI). The last two columns show the performance of the optimal solutions computed through Value Iteration, highlighting the increasing difficulty of the task.}
\label{tab:steps-by-map-exp-algo-bestpath-sr}
\resizebox{\textwidth}{!}{%
\begin{tabular}{llrrrrrrrr}
\toprule
\textbf{Map} & \textbf{Exp} & \textbf{Q-Learning} & \textbf{R-MAX} & \textbf{QR-MAX}
              & \textbf{QRM}       & \textbf{R-MAXRM}  & \textbf{QR-MAXRM}
              & \textbf{Average length (VI)} & \textbf{Success rate (VI) (\%)} \\
\midrule
Map1 & exp1 &   128,973 &  59,691 &  \textbf{30,376} &  66,693 &  21,777 &  \textbf{10,964} &  21.08 &  66.76 \\
Map1 & exp2 &   201,835 &  77,797 &  \textbf{27,619} & 129,804 &  26,232 &   \textbf{9,339} &  39.31 &  39.40 \\
Map1 & exp3 &   355,070 & 155,347 &  \textbf{30,325} & 122,827 &  28,682 &   \textbf{6,087} &  40.94 &  38.68 \\
Map1 & exp4 &   469,313 & 103,211 &  \textbf{29,950} & 135,119 &  26,953 &   \textbf{6,062} &  42.23 &  34.37 \\
Map1 & exp5 & 1,267,449 & 287,066 &  \textbf{38,312} & 231,870 &  30,238 &   \textbf{4,107} &  77.86 &  15.47 \\
\addlinespace
Map2 & exp1 &   334,894 & 213,942 &  \textbf{83,593} & 225,491 &  74,250 &  \textbf{28,729} &  52.46 &  54.69 \\
Map2 & exp2 &   326,306 &  88,636 &  \textbf{56,445} & 220,334 &  49,577 &  \textbf{19,214} &  53.10 &  37.92 \\
Map2 & exp3 &   723,051 & 316,672 &  \textbf{62,583} & 280,763 &  63,066 &  \textbf{12,415} &  63.04 &  37.25 \\
Map2 & exp4 &   889,842 & 196,334 &  \textbf{61,775} & 249,912 &  47,873 &  \textbf{12,445} &  73.80 &  28.18 \\
Map2 & exp5 & 2,613,669 & 723,441 &  \textbf{83,076} & 438,213 &  65,523 &   \textbf{8,396} & 123.39 &  14.15 \\
\addlinespace
Map3 & exp1 &   501,559 & 201,731 &  \textbf{98,663} & 297,975 &  73,832 &  \textbf{33,815} &  55.83 &  33.32 \\
Map3 & exp2 &   553,517 & 170,926 &  \textbf{99,952} & 353,658 &  66,949 &  \textbf{33,853} &  63.76 &  32.57 \\
Map3 & exp3 & 1,105,417 & 388,562 & \textbf{110,808} & 406,122 & 111,349 &  \textbf{21,856} &  68.69 &  32.42 \\
Map3 & exp4 & 1,466,896 & 495,878 & \textbf{112,165} & 449,594 & 107,115 &  \textbf{21,869} &  81.91 &  21.27 \\
Map3 & exp5 & 5,471,046 &1,581,301 & \textbf{159,597} & 891,160 & 128,950 &  \textbf{14,717} & 156.46 &   6.55 \\
\bottomrule
\end{tabular}%
}
\end{table*}

\begin{figure}[t]
	\centering
	\includegraphics[width=0.9
	\linewidth]{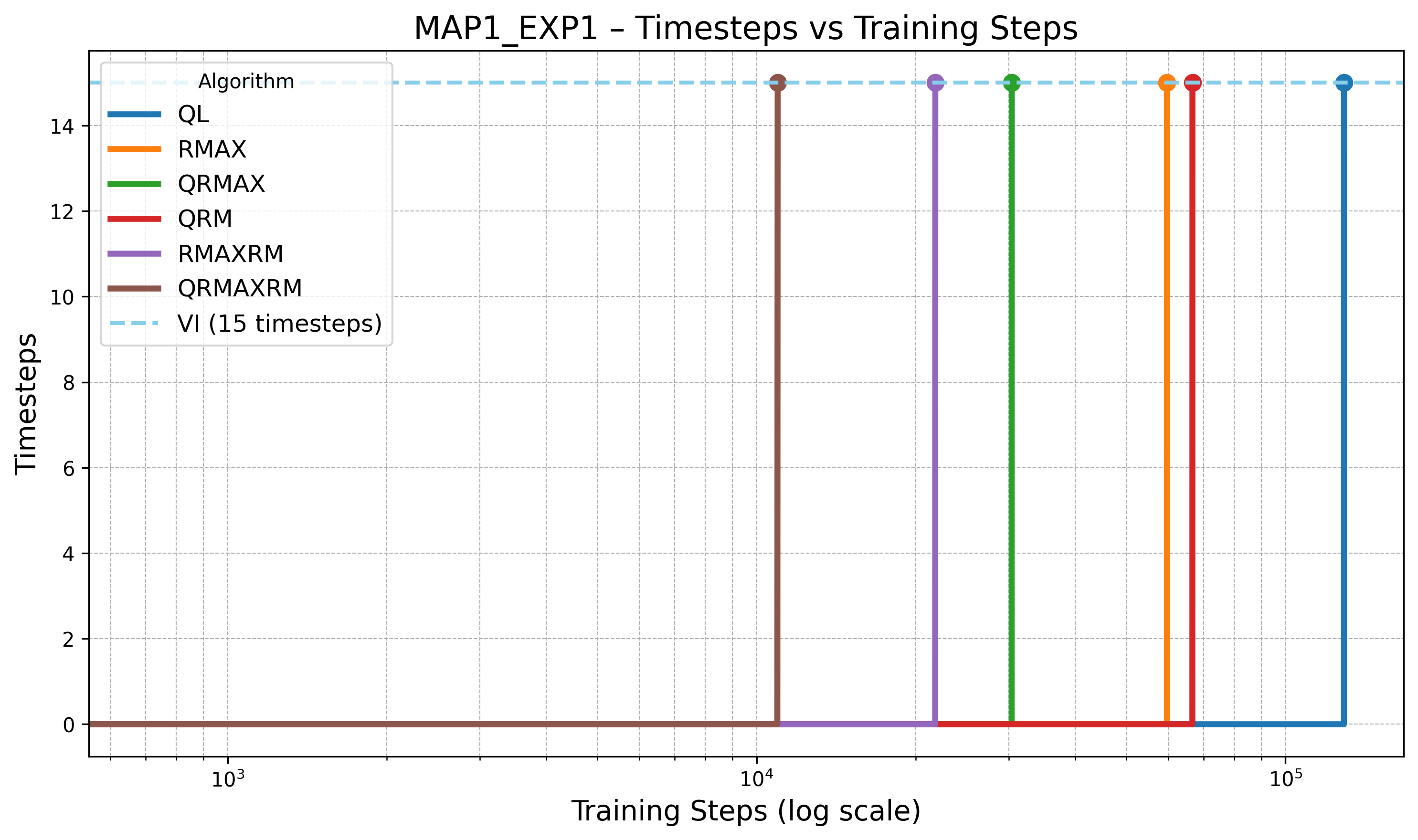}
	\caption{Map1 Exp1 - Office World}
	\label{fig:map1_exp1}
\end{figure}

\begin{figure}[t]
	\centering
	\includegraphics[width=0.9
	\linewidth]{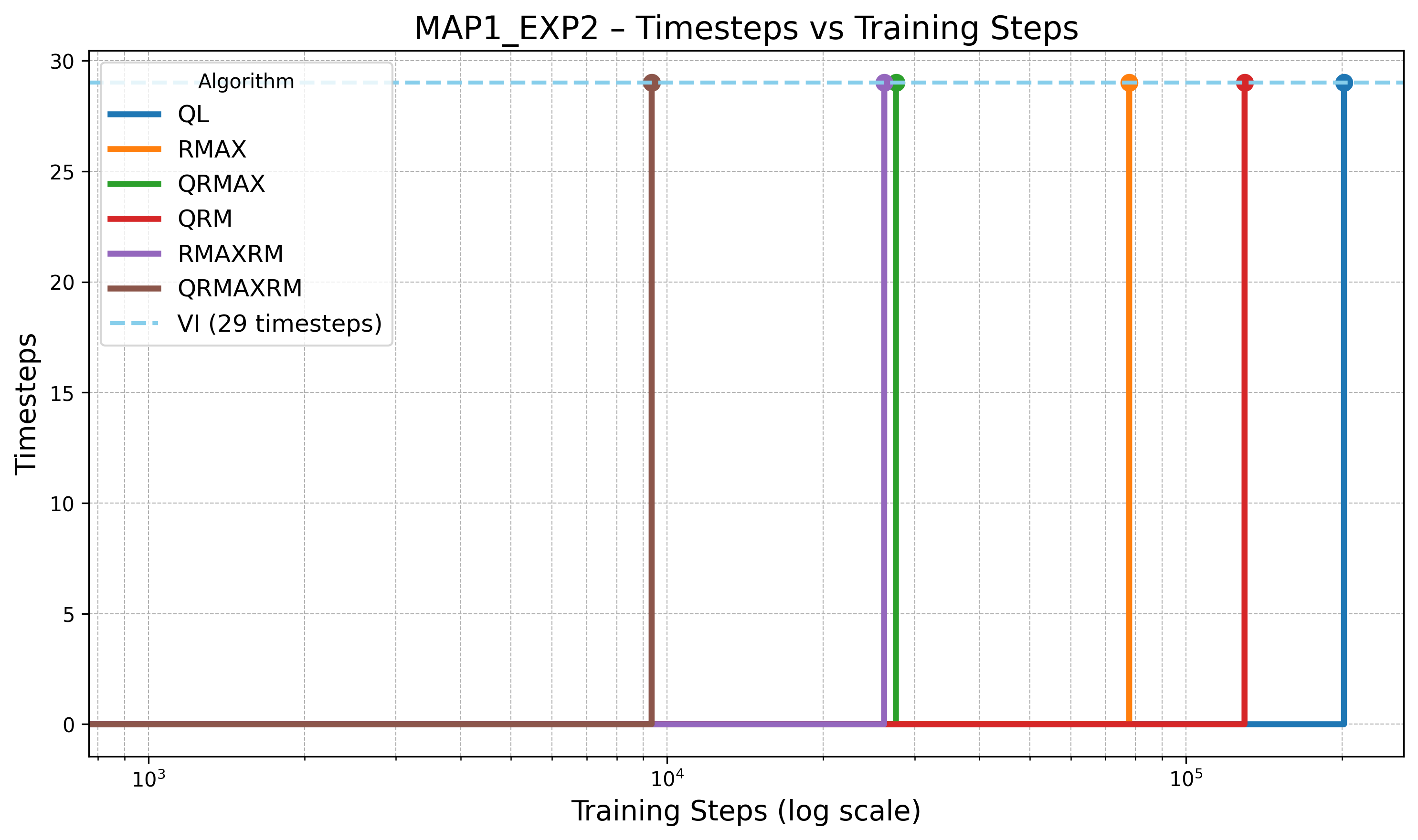}
	\caption{Map1 Exp2 - Office World}
	\label{fig:map1_exp2}
\end{figure}

\begin{figure}[t]
	\centering
	\includegraphics[width=0.9
	\linewidth]{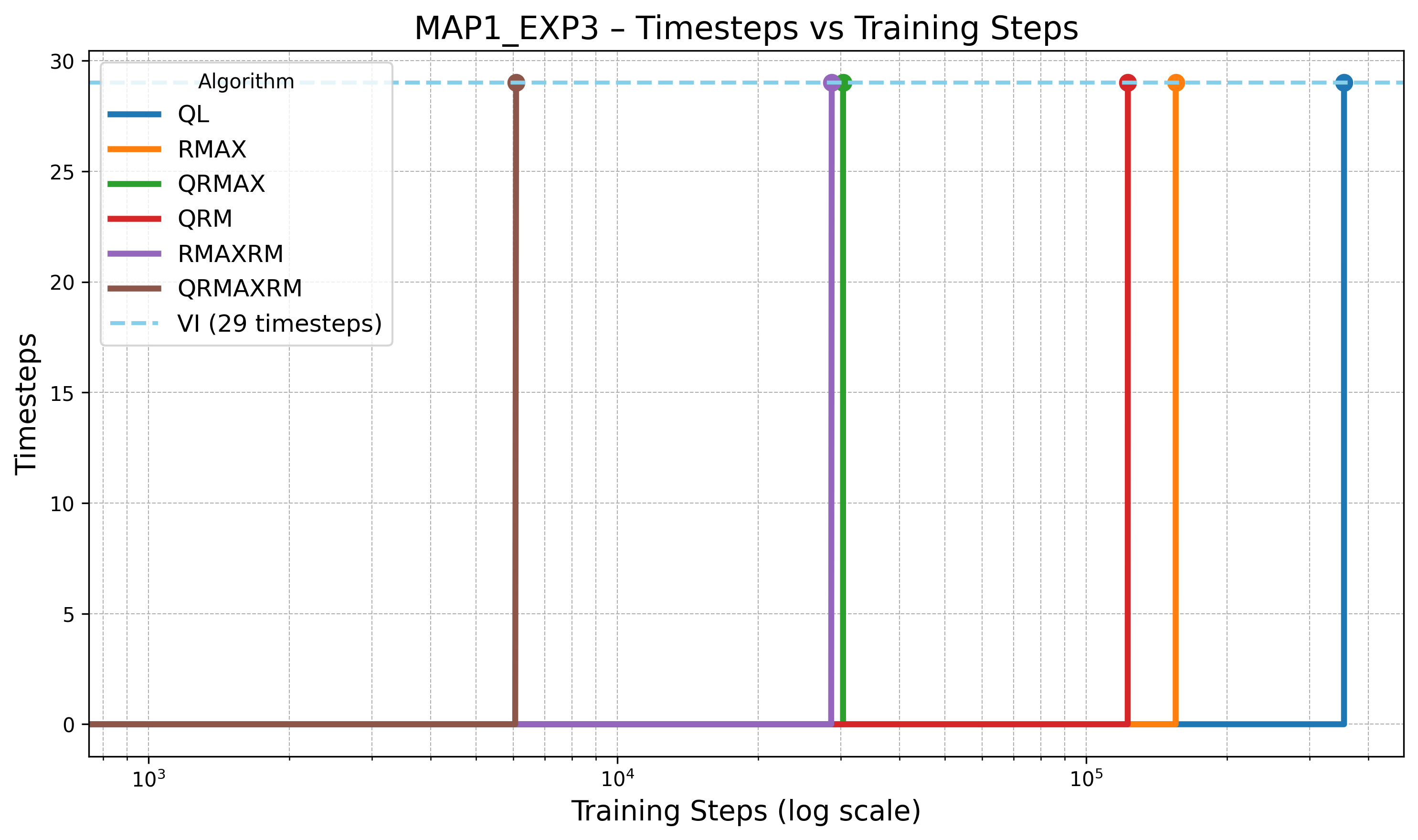}
	\caption{Map1 Exp3 - Office World}
	\label{fig:map1_exp3}
\end{figure}

\begin{figure}[t]
	\centering
	\includegraphics[width=0.9
	\linewidth]{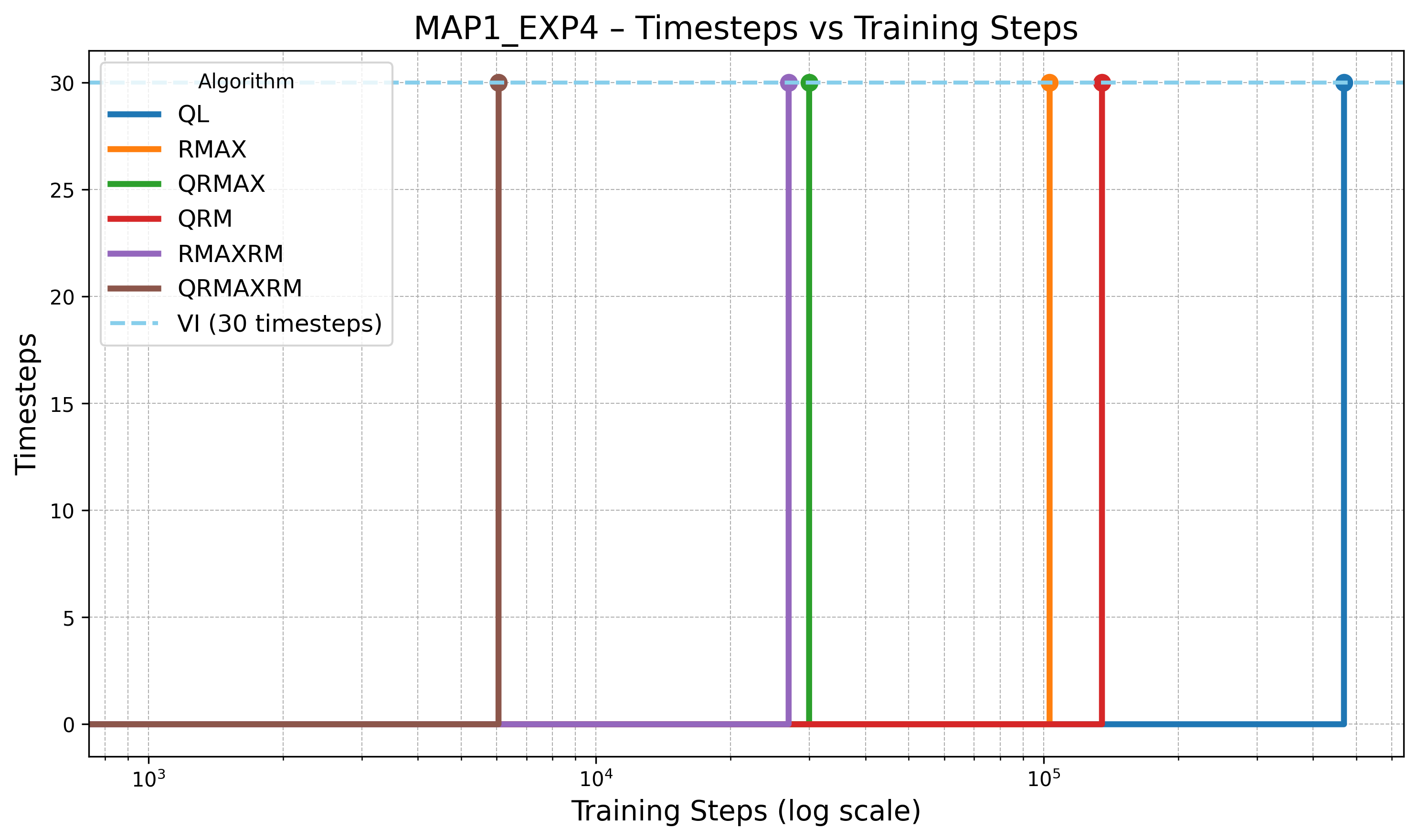}
	\caption{Map1 Exp4 - Office World}
	\label{fig:map1_exp4}
\end{figure}

\begin{figure}[t]
	\centering
	\includegraphics[width=0.9
	\linewidth]{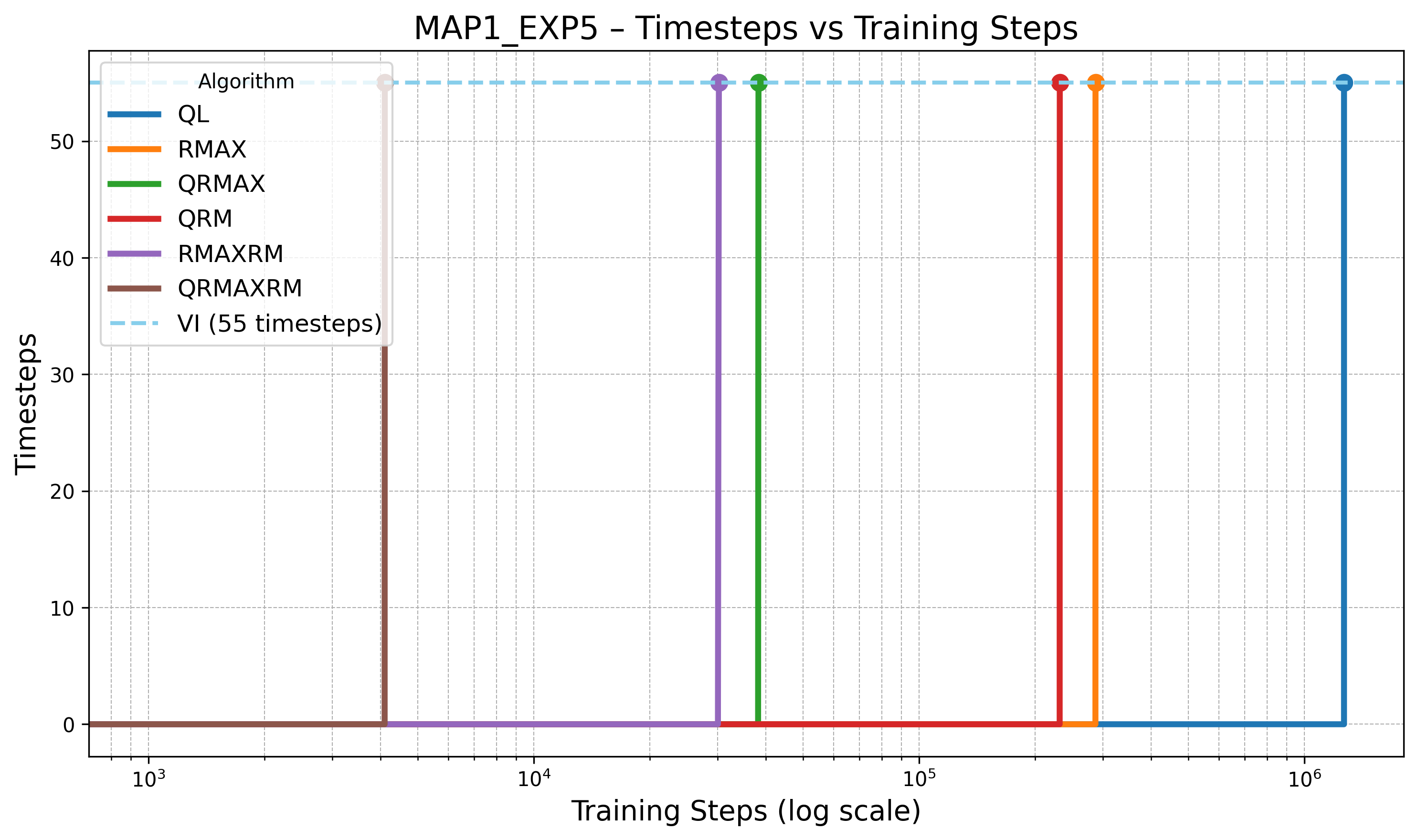}
	\caption{Map1 Exp5 - Office World}
	\label{fig:map1_exp5_}
\end{figure}


\begin{figure}[t]
	\centering
	\includegraphics[width=0.9
	\linewidth]{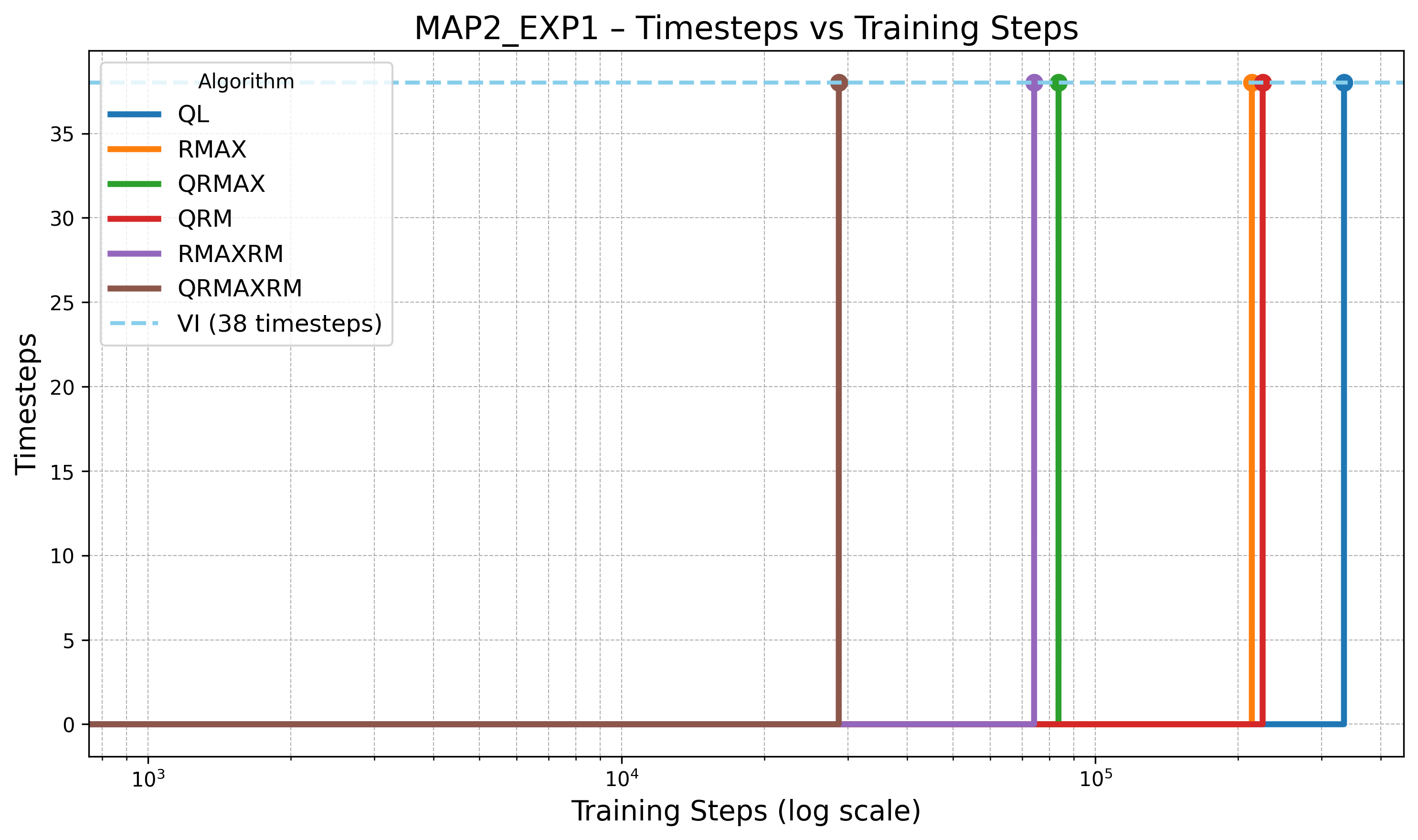}
	\caption{Map2 Exp1 - Office World}
	\label{fig:map2_exp1}
\end{figure}

\begin{figure}[t]
	\centering
	\includegraphics[width=0.9
	\linewidth]{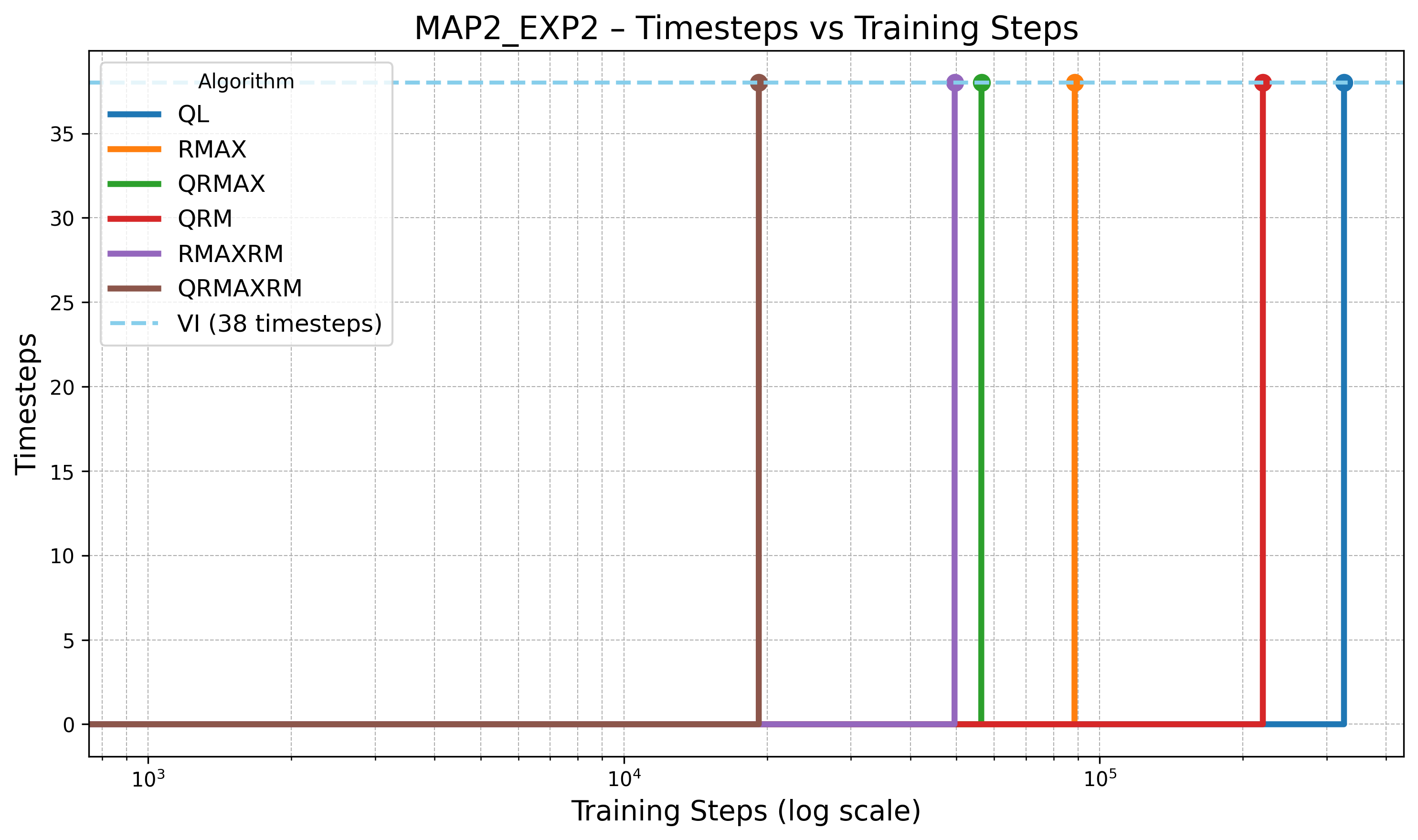}
	\caption{Map2 Exp2 - Office World}
	\label{fig:map2_exp2}
\end{figure}

\begin{figure}[t]
	\centering
	\includegraphics[width=0.9
	\linewidth]{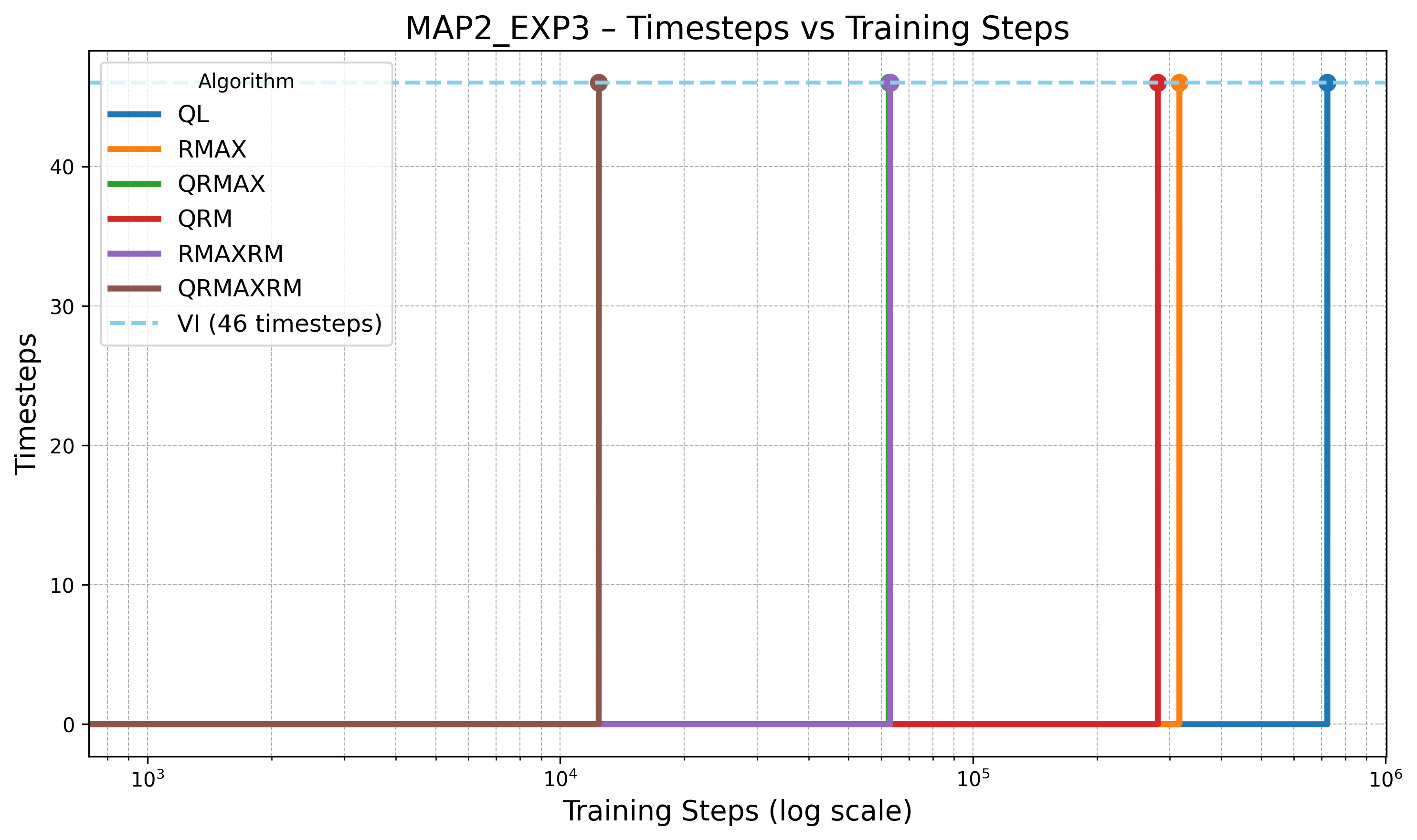}
	\caption{Map2 Exp3 - Office World}
	\label{fig:map2_exp3}
\end{figure}

\begin{figure}[t]
	\centering
	\includegraphics[width=0.9
	\linewidth]{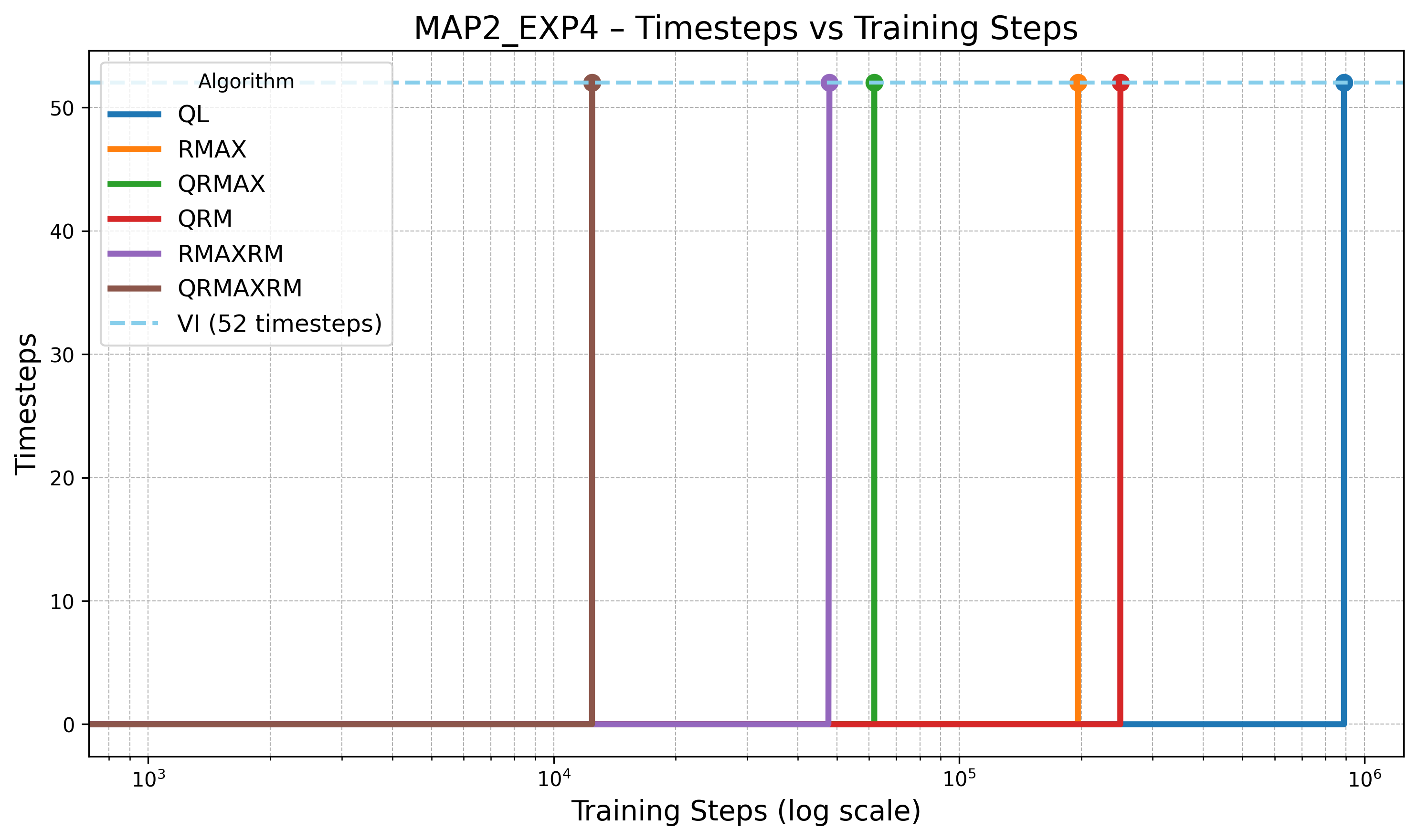}
	\caption{Map2 Exp4 - Office World}
	\label{fig:map2_exp4}
\end{figure}

\begin{figure}[t]
	\centering
	\includegraphics[width=0.9
	\linewidth]{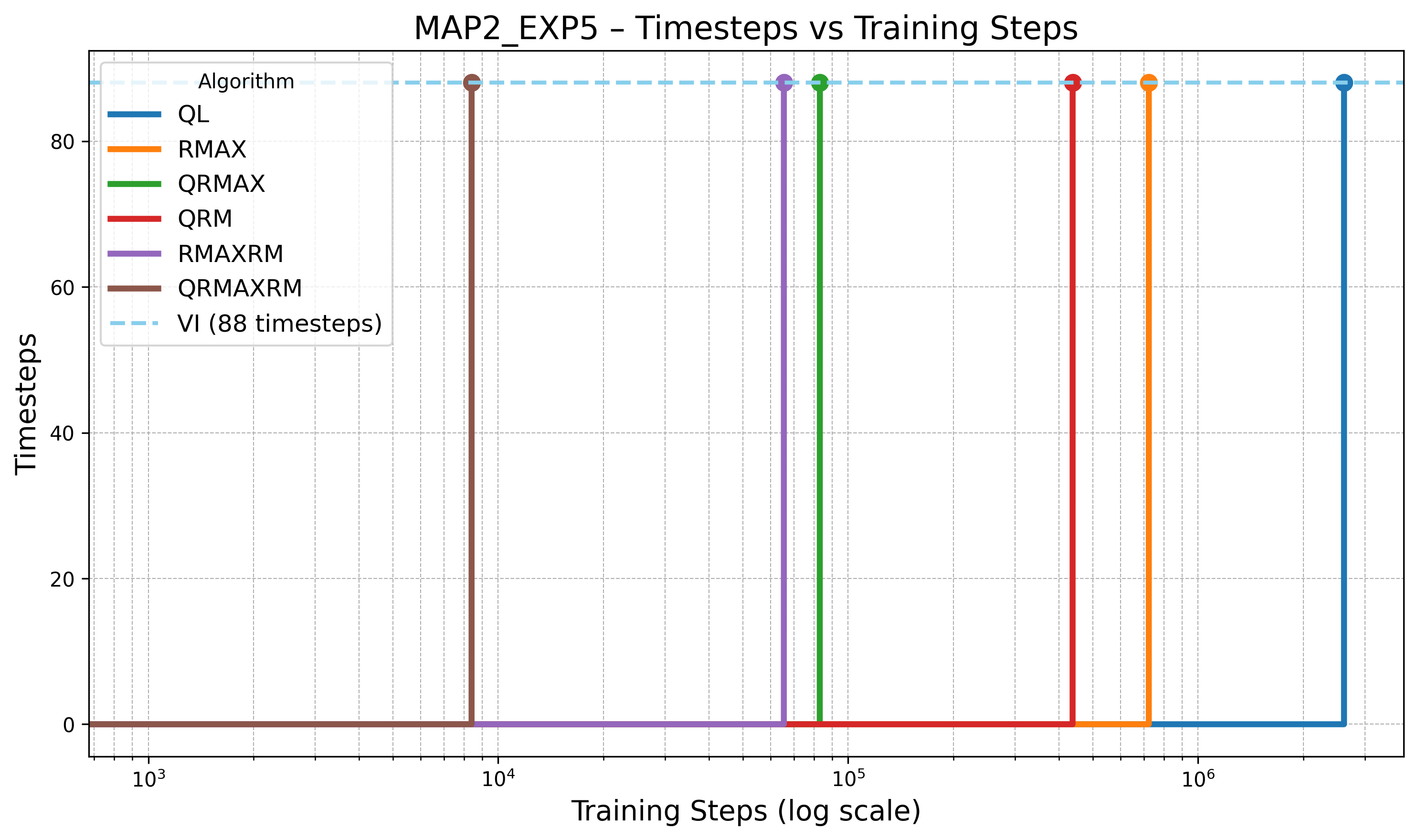}
	\caption{Map2 Exp5 - Office World}
	\label{fig:map2_exp5}
\end{figure}


\begin{figure}[t]
	\centering
	\includegraphics[width=0.9
	\linewidth]{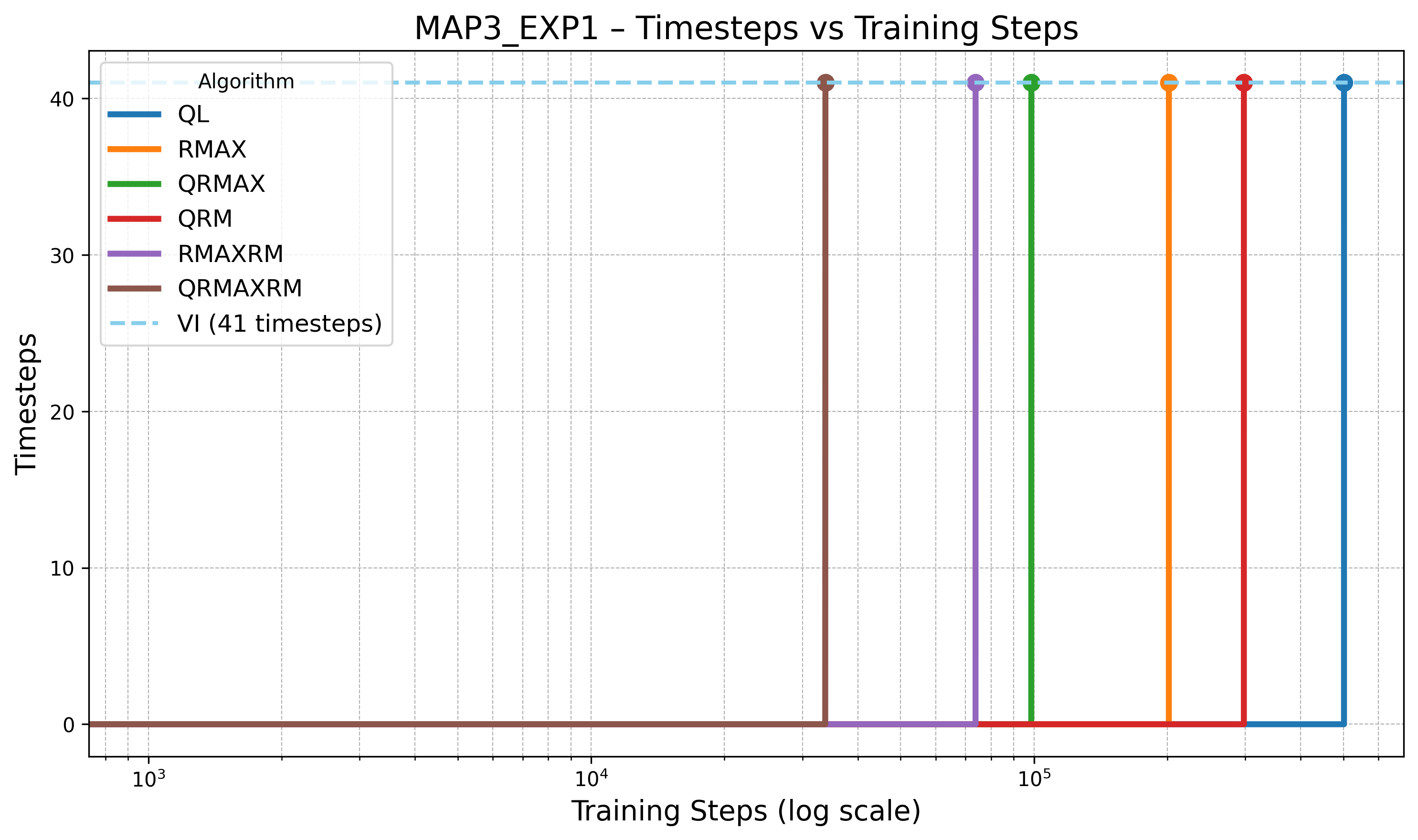}
	\caption{Map3 Exp1 - Office World}
	\label{fig:map3_exp1}
\end{figure}

\begin{figure}[t]
	\centering
	\includegraphics[width=0.9
	\linewidth]{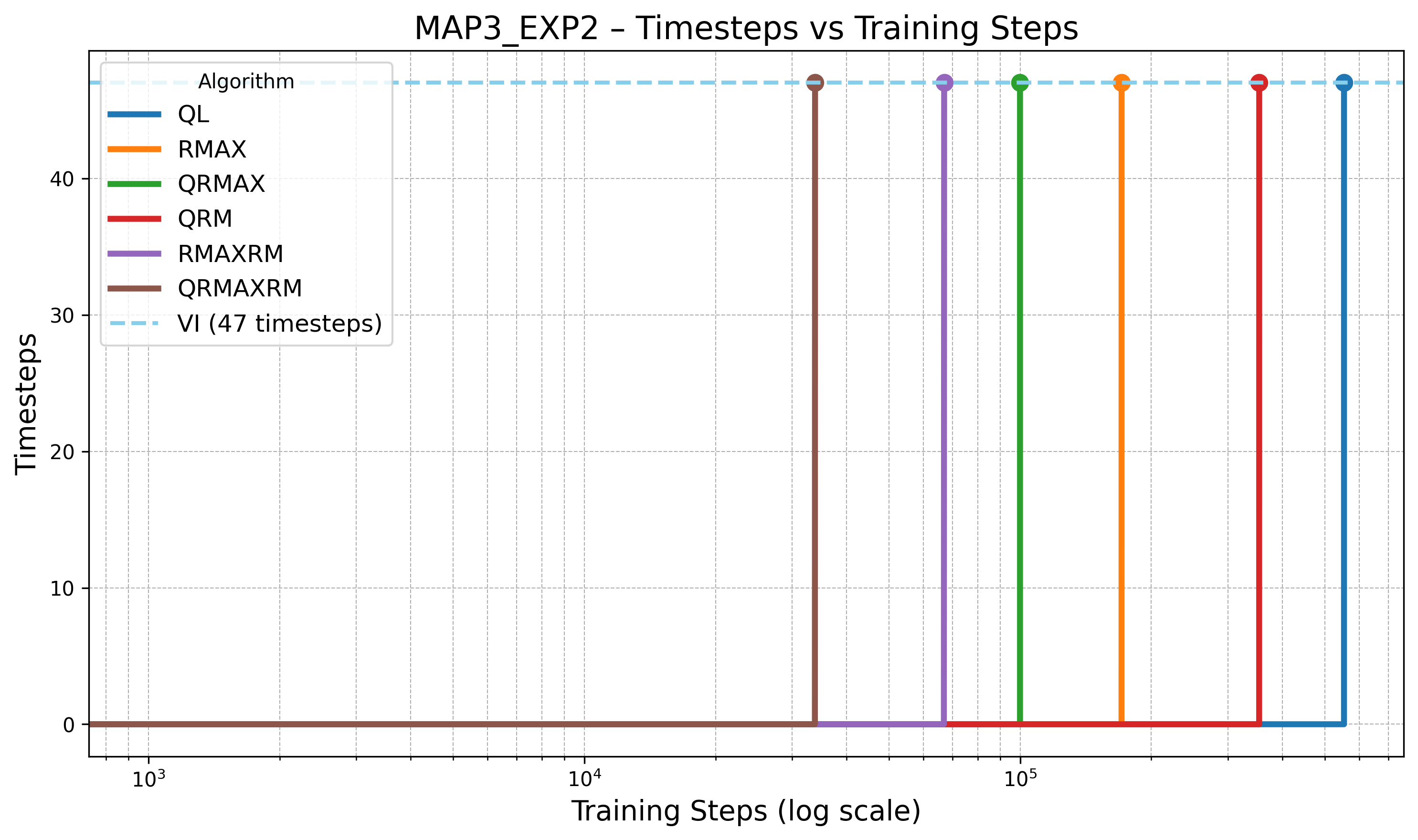}
	\caption{Map3 Exp2 - Office World}
	\label{fig:map3_exp2}
\end{figure}

\begin{figure}[t]
	\centering
	\includegraphics[width=0.9
	\linewidth]{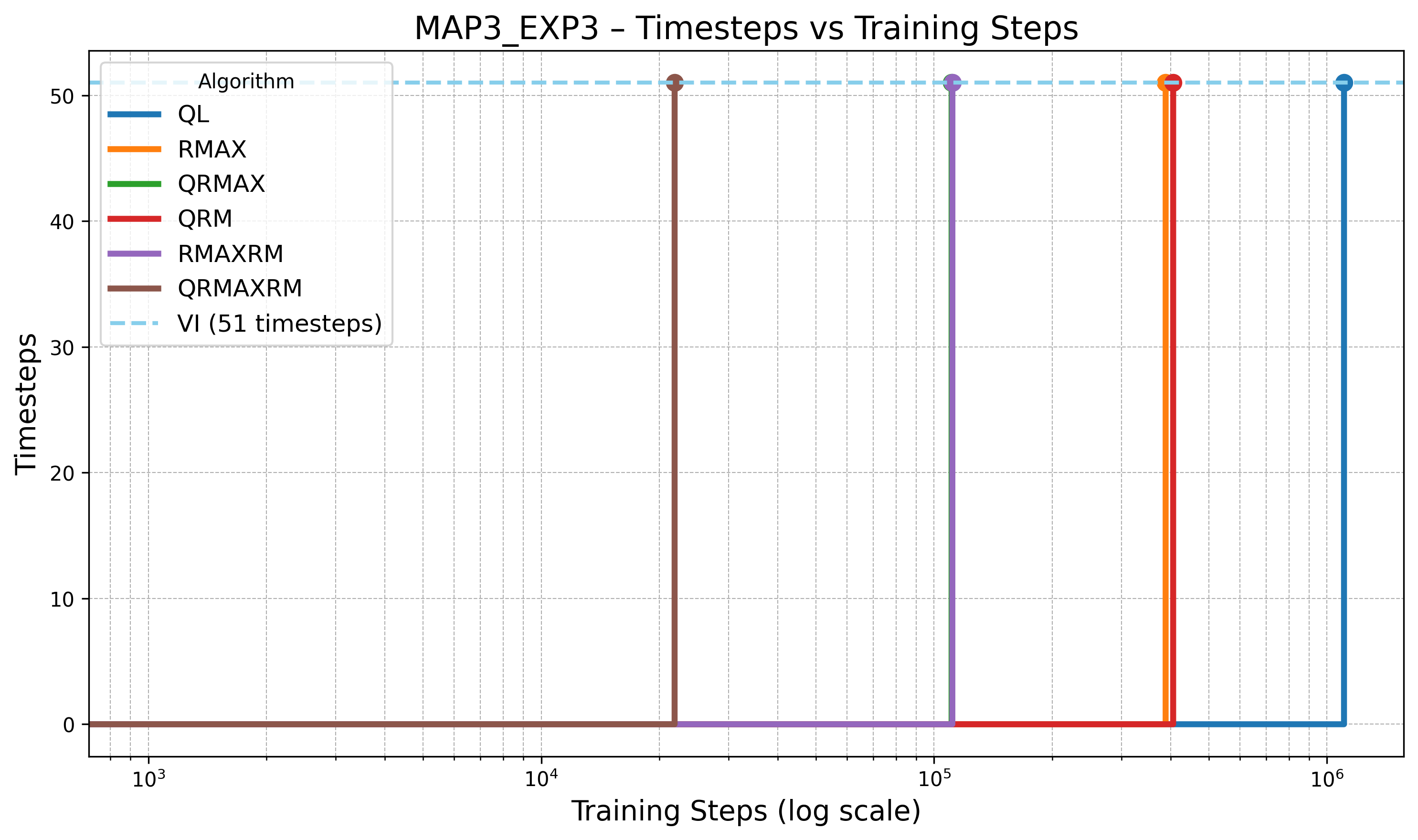}
	\caption{Map3 Exp3 - Office World}
	\label{fig:map3_exp3}
\end{figure}

\begin{figure}[t]
	\centering
	\includegraphics[width=0.9
	\linewidth]{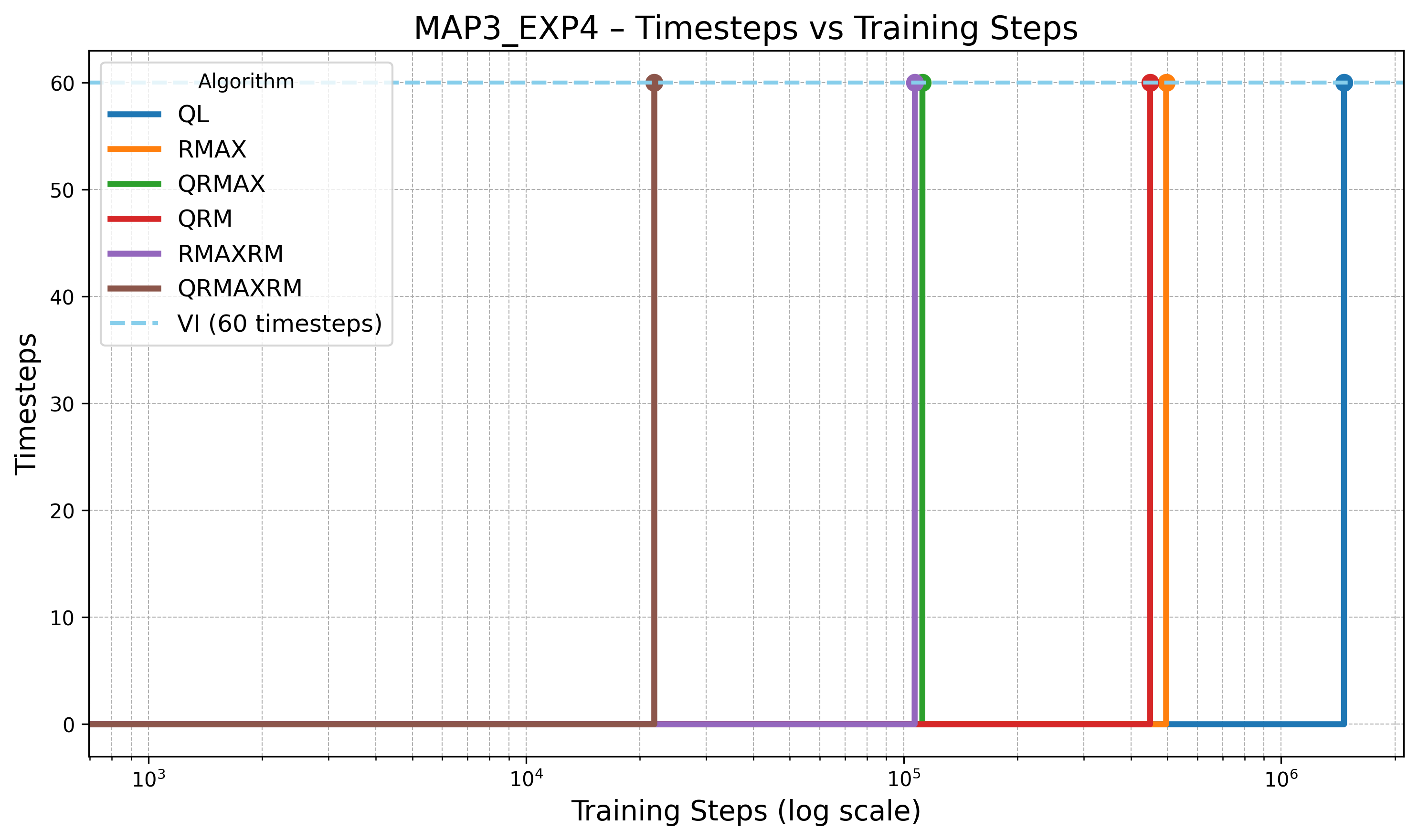}
	\caption{Map3 Exp4 - Office World}
	\label{fig:map3_exp4}
\end{figure}

\begin{figure}[t]
	\centering
	\includegraphics[width=0.9
	\linewidth]{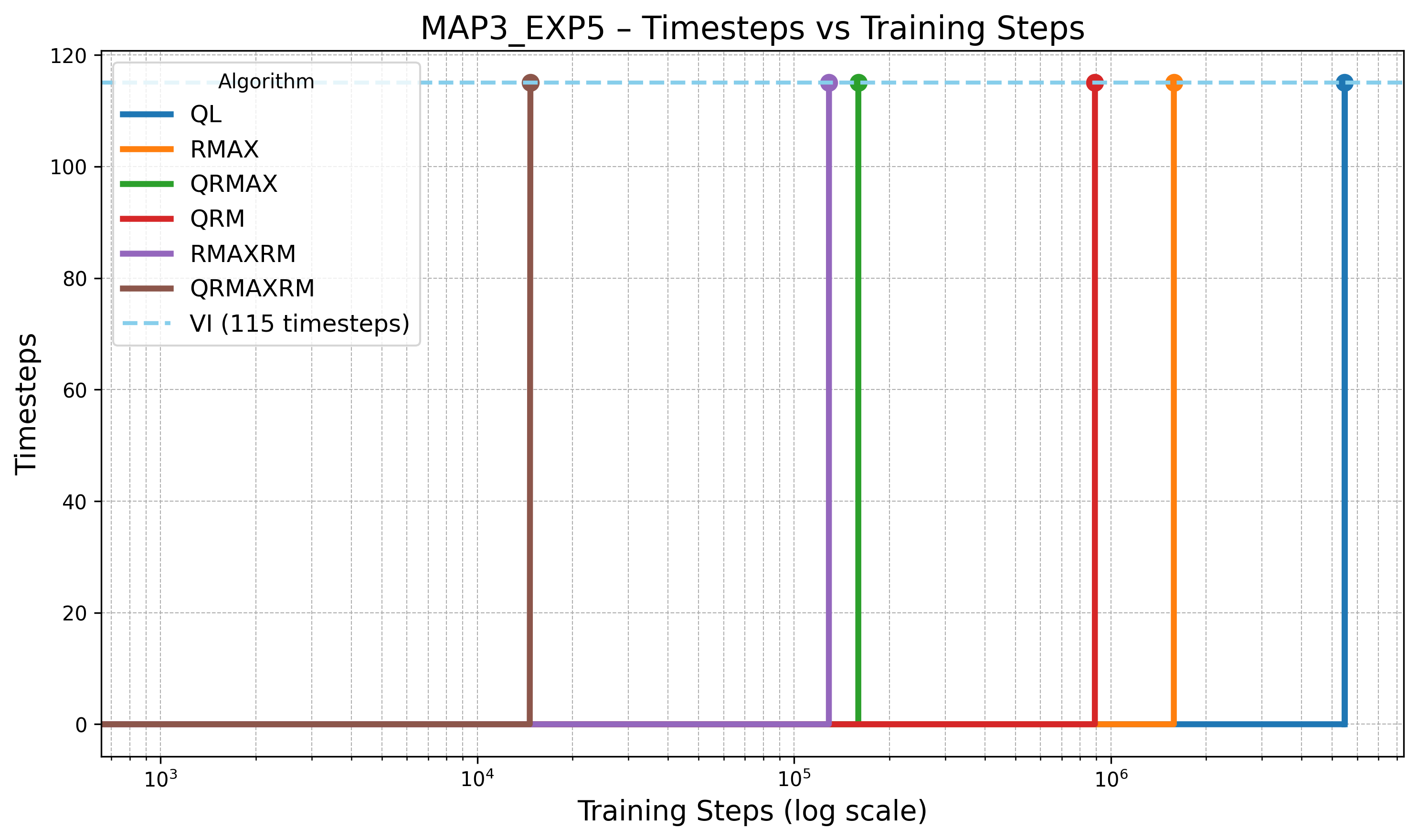}
	\caption{Map3 Exp5 - Office World}
	\label{fig:map3_exp5}
\end{figure}


\begin{table*}[t]
  \centering
  \scriptsize
  \setlength{\tabcolsep}{8.5pt}
  \begin{tabular}{lcccl}
    \toprule
    \textbf{Method} &
    \textbf{Memory} &
    \textbf{$H$-dep.?} &
    \textbf{Sample complexity$^{*}$} &
    \textbf{Key drivers}\\
    \midrule
    \textsc{QR-MAX} (ours) &
      \textbf{$\tilde{\mathcal O}\!\bigl(|S|^{2}|A| + |S||Q|\bigr)$} &
      \textbf{No} &
      \textbf{$\tilde{\mathcal O}\!\bigl(|S||A| + |S||Q|\bigr)$} &
      $(s,a,s')$ counts + automaton counts $(q,s',q')$\\
    \textsc{UCBVI} &
      $\tilde{\mathcal O}\!\bigl(H\,|S|^{2}|A|\bigr)$ &
      Yes &
      $\tilde{\mathcal O}\!\bigl(|S||A|\,H^{3}\bigr)$\,$\!^{\dagger}$ &
      Empirical counts and UCB bonuses for every time--step $(h,s,a)$\\
    \textsc{OPSRL} / \textsc{PSRL} &
      $\tilde{\mathcal O}\!\bigl(H\,|S|^{2}|A|\bigr)$ &
      Yes &
      — (high-probability regret only) &
      Posterior parameters (Dirichlet / Gaussian) for every $(h,s,a)$\\
    \textsc{R-MAX}$^{\ddagger}$ &
      $\tilde{\mathcal O}\!\bigl(|S|^{2}|A||Q|^{2}\bigr)$ &
      No &
      $\tilde{\mathcal O}\!\bigl(|S||A||Q|\bigr)$ &
      Counts on the full product MDP $S{\times}Q$\\
    \textsc{QRM} &
      $\tilde{\mathcal O}\!\bigl(|S||A||Q|\bigr)$ &
      No &
      — (no PAC) &
      One $Q$-value table per $(s,q,a)$\\
    \bottomrule
  \end{tabular}
  \caption{Asymptotic memory and (when available) PAC-style sample complexity.
  $\tilde{\mathcal O}$ hides poly‐log factors; $H$ is the episode horizon,
  $Q$ the number of reward-machine states.
  $^{\dagger}$Based on Azar et al.\ (2017), Thm.~2, combined with the standard conversion from regret to PAC-style bounds (omitting the dependence on $\varepsilon$).  
  $^{\ddagger}$Brafman \& Tennenholtz (2002).
  \newline
  $^{*}$Sample-complexity entries are given in terms of $|S|$, $|A|$, $|Q|$ and $H$, omitting the explicit dependence on $\varepsilon$ and $(1-\gamma)$ for clarity.
  \newline
  \textbf{Reference scenario} — 15 × 15 grid: $|S|{=}225$, $|A|{=}4$,
  $Q{=}8$, $H{=}250$.
  \newline
  • QR-MAX memory: $225^{2}\!\times4 + 225\!\times8 = 0.204$ M entries.  
  • OPSRL/PSRL memory: $250\!\times225^{2}\!\times4 = 50.6$ M.  
  $\Rightarrow$ \textbf{$\approx$ 250× less memory}.  
  \newline
  • QR-MAX PAC bound (in $|S|,|Q|$): $225\!\times4 + 225\!\times8 = 2.7$ k.  
  • UCBVI PAC-style bound (in $|S|,H$): $225\!\times4\!\times250^{3} \approx 3.5\times10^{12}$.  
  $\Rightarrow$ \textbf{$\approx$ 10\textsuperscript{9} × fewer samples}.}
  \label{tab:mem_sample_complexity}
\end{table*}

\begin{table*}[t]
\centering
\small
\caption{Average training steps to find a policy that produces results indistinguishable from VI optimal solution (mean over 10 random seeds).}%
\label{tab:steps-map-exp_}
\setlength\tabcolsep{4pt}
\resizebox{\textwidth}{!}{%
\begin{tabular}{llrrrrrrrr}
\toprule
\textbf{Map} & \textbf{Exp} &
\textbf{QR‑MAX} & \textbf{QR‑MAXRM} & \textbf{R‑MAX} & \textbf{QRM} &
\textbf{UCBVI‑sB} & \textbf{UCBVI‑B} & \textbf{UCBVI‑H} &
\textbf{PSRL/OPSRL$^{\dagger}$} \\
\midrule
MAP0 & EXP0 &
\textbf{14\,483} & \textbf{4\,150} & 49\,759 & 162\,040 &
142\,015 & 159\,092 & 191\,983 & 88\,437 / 128\,188 \\[2pt]
MAP1 & EXP5 &
\textbf{24\,222} & \textbf{3\,125} & 272\,080 & 225\,140 &
250\,800 & 555\,977 & 1\,320\,070 & --- \\[2pt]
MAP4 & EXP6 &
\textbf{20\,150} & \textbf{5\,630} & 82\,000 & 9\,910\,340 &
80\,160 & 90\,030 & 101\,120 & --- \\ [2pt]
MAP2 & EXP5 &
\textbf{83\,076} & \textbf{3\,767} & 723\,441 & 438\,213 &
1\,150\,000 & 1\,400\,000 & 1\,890\,000 & --- \\[2pt]
MAP3 & EXP5 &
\textbf{159\,597} & \textbf{5\,806} & 1\,581\,301 & 891\,160 &
1\,410\,000 & 1\,620\,000 & 2\,030\,000 & --- \\ [2pt]
\bottomrule
\multicolumn{10}{p{0.96\linewidth}}{\footnotesize
$^{\dagger}$Eight posterior samples per episode, as in the original papers.}
\end{tabular}%
} 
\end{table*}

\paragraph{Discussion of the results.}
As summarised in Table~\ref{tab:steps-map-exp_}, the factorised agent \qrmax consistently attains a policy statistically indistinguishable from the value--iteration optimum after roughly one order of magnitude fewer interactions than the strongest optimistic baselines. On the easiest instance (\texttt{MAP0--EXP0}) \qrmax converges in about \(1.4 \times 10^{4}\) transitions, whereas the most sample–efficient UCBVI variant requires over \(1.4 \times 10^{5}\); on the hardest instance (\texttt{MAP3--EXP5}) the respective figures are \(1.6 \times 10^{5}\) versus at least \(1.4 \times 10^{6}\).

When the Reward--Machine structure is supplied, \qrmaxrm delivers an even steeper reduction. Across the four largest benchmarks it converges in \(3.8 \times 10^{3}\) to \(5.8 \times 10^{3}\) interactions, while the model-free \textsc{QRM} requires from half a million to ten million and the non-factorised \rmaxrm more than \(1 \times 10^{5}\). Hence, coupling factorisation with explicit knowledge of the logical task cuts sample complexity by two, and in one case three, orders of magnitude relative to the best published alternatives. These results position \qrmax as the best when the reward structure is unknown, and establish \qrmaxrm as the most sample-efficient algorithm presently available when the Reward Machine is given.


\end{document}